%% file: neurips_2026.tex
\documentclass{article}

\usepackage[main, final]{neurips_2026}

\usepackage[utf8]{inputenc} 
\usepackage[T1]{fontenc}    
\usepackage{hyperref}       
\usepackage{url}            
\usepackage{booktabs}       
\usepackage{amsfonts}       
\usepackage{nicefrac}       
\usepackage{microtype}      
\usepackage{xcolor}         
\usepackage{subcaption}
\usepackage{multirow}
\usepackage{graphicx}
\usepackage{amsmath}
\usepackage{amssymb}
\usepackage{amsthm}
\usepackage{cleveref} 
\usepackage{adjustbox}
\usepackage{enumitem}

\DeclareMathOperator{\action}{\mathbf{a}}
\DeclareMathOperator{\state}{\mathbf{s}}

\newtheorem{theorem}{Theorem}

\newtheorem{proposition}[theorem]{Proposition}

\newcommand{\restatedpropositionname}{}
\newtheorem*{restatedpropositioninner}{\restatedpropositionname}
\newenvironment{restatedproposition}[1]{%
  \renewcommand{\restatedpropositionname}{Proposition~\ref{#1}}%
  \begin{restatedpropositioninner}%
}{\end{restatedpropositioninner}}

\newcommand{\Leaf}{\mathcal{L}}

\newcommand{\Node}{\mathcal{N}}

\title{Split-RL: Local Conflict Resolution in Reinforcement Learning}

\author{%
 Benjamin Fuhrer \\
  NVIDIA \\
  \texttt{bfuhrer@nvidia.com} \\
  \And 
 Chen Tessler \\
  NVIDIA Research \\
  \texttt{ctessler@nvidia.com} \\
\And 
 Gal Dalal \\
  NVIDIA Research \\
  \texttt{gdalal@nvidia.com} \\
}

\begin{document}
\maketitle
\begin{abstract}
Reinforcement learning (RL) policies are often trained by mixing contradictory feedback that is localized in space and time. For example, a robot may prioritize speed in open areas versus precision in tight gaps.
We introduce Split-RL, a Gradient Boosting Trees (GBT) algorithm that uses guidance labels to locally isolate these conflicting signals.
Standard neural networks (NNs) lack the structural mechanism to do this, as their parameters share feedback from all sources. While GBT provides the inductive bias to explicitly partition the state-space per objective, standard tree-fitting procedures aggregate contradictory gradient signals before they reach the leaves. Split-RL leverages this bias to integrate guidance labels into tree construction, routing these updates into disjoint state-space regions and preventing signal interference. Such labels can arise from heuristics, sensor-derived events, or existing rule-based systems, allowing Split-RL to incorporate rule-derived structure while retaining a learned policy. We theoretically show how early gradient aggregation loses objective-specific information and characterize when the Split-RL score isolates conflicting updates. We focus on constrained RL and offline RL with mixed datasets, each balancing two objectives that conflict locally. Across domains with spatial, rare-event, and temporal conflicts, Split-RL consistently outperforms existing methods by meeting the episodic cost budget while maintaining competitive rewards. In offline settings, Split-RL localizes the tradeoff between imitation and policy improvement, outperforming other methods and achieving near-expert performance across varying dataset sizes.
\end{abstract}

\section{Introduction}
\label{sec:introduction}
In many reinforcement learning (RL) settings, a single policy must be trained from multiple signals whose importance varies across state or time. For instance, an autonomous vehicle may optimize speed on highways but prioritize conservative behavior in school zones. Similarly, a controller may balance drug dosage effectiveness while retaining safe levels across time. In these settings, the challenge is not just that objectives conflict, but that the conflict is \emph{localized}: most states support compatible behavior, while specific, coherent regions require a different objective to dominate.

This setting exposes a mismatch with the inductive bias of standard parametric policies, such as neural networks (NNs). Because neural policies share parameters across states, an update driven by one region can also change the policy in another. When signals induce incompatible updates, such as a high-reward task update clashing with a critical safety-cost update, the averaged update can degrade individual objectives through gradient conflict~\citep{pcgrad,liu2021conflict,sener2018multi,navon2022multi}. We refer to this failure mode as \textbf{premature gradient aggregation}: combining objective-specific signals before the learner has isolated the states in which they should control the update. This premature coupling results in two distinct issues: \textbf{signal dilution}, where the majority distribution washes out rare but critical regions, and \textbf{gradient interference}, where conflicting updates of similar magnitude cancel into a suboptimal compromise.

These conflict regions are often easy to identify. A human operator, sensor-derived event, or simple rule-based heuristic can mark the preferred objective to optimize. This makes localized conflict a natural setting for combining rule-derived structure with learning: the rule specifies where an objective should dominate, while the learner determines the policy update within that region. Although these guidance labels provide a clear signal for localizing objectives, standard parametric models such as NNs lack a mechanism to exploit this information because their global updates aggregate signals regardless of local context. Gradient Boosting Trees (GBT) provides the structural machinery to explicitly partition the state-space, yet standard tree-fitting procedures fit only aggregated gradients, combining contradictory updates before they reach the leaves and preventing standard trees from isolating conflicting signals into disjoint regions.

\begin{figure}[htb!]
    \vspace{-0.5em}
    \centering
    \includegraphics[width=\linewidth]{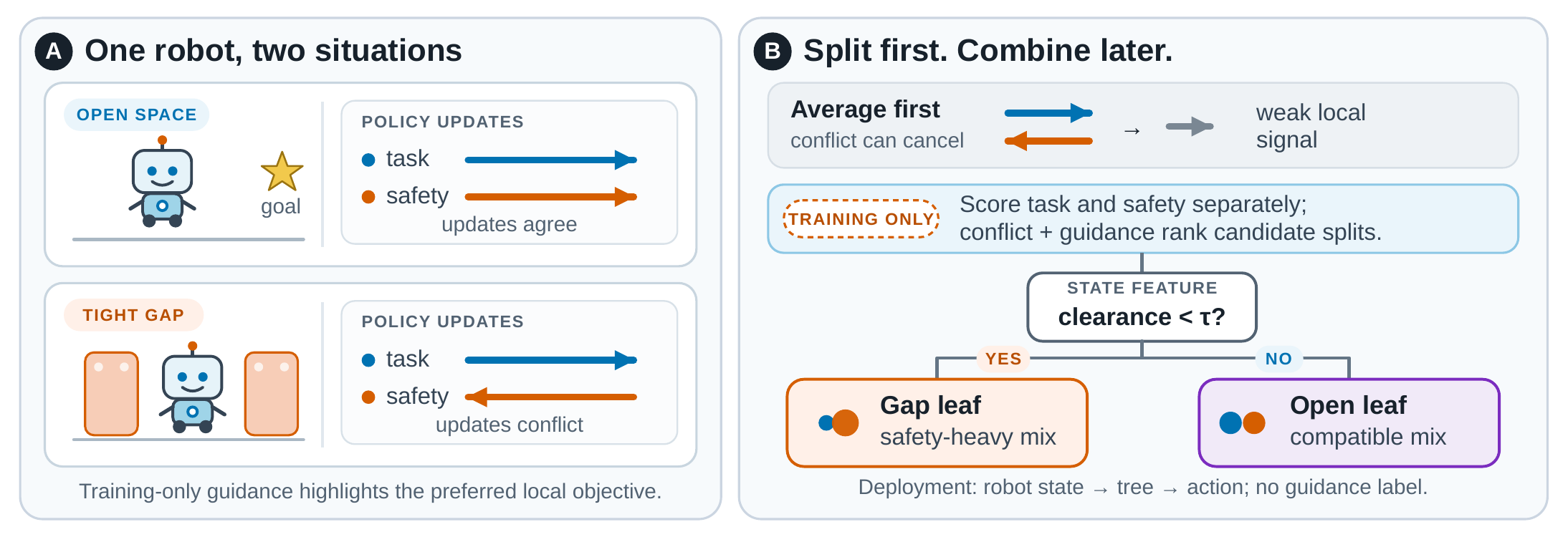}
\caption{\textbf{Split-RL -- Guided signal routing.}
\textbf{(A)} A robot's task and safety updates align in open space but conflict in a tight gap.
\textbf{(B)} Split-RL scores objectives separately and uses guidance during training to select tree splits before combining updates within leaves. The gap leaf emphasizes safety, while the open-space leaf retains a mixture of aligned updates.}
    \label{fig:split_rl_conflict}
        \vspace{-0.5em}
\end{figure}

To address these limitations, we build on recent advances in GBT-based RL~\citep{fuhrer2025gradient} to introduce \textbf{Split-RL}, a framework that uses guided supervision to control update sharing under localized objective conflict. Split-RL integrates guidance labels into the tree-split selection process to enforce structural isolation between conflicting objectives. By prioritizing partitions that route contradictory gradients into separate leaves, the algorithm incorporates rule-based logic into the learning process and shapes which samples share an update. This structural separation prevents the signal interference inherent in globally coupled models, ensuring that each objective dominates only within a localized region, as we illustrate in Figure~\ref{fig:split_rl_conflict}.

This form of conflict appears across RL. In constrained RL, safety violations are often localized, but standard methods~\citep{ppo_lag,cpo} control expected cost at the policy level rather than deciding which states should receive cost-sensitive updates. Similarly, offline RL requires balancing policy improvement against extrapolation error~\citep{kumar2020conservativeqlearningofflinereinforcement, kostrikov2021iql}. Because standard global regularization applies a uniform level of conservatism, it often washes out expert signals while failing to suppress suboptimal behaviors. Split-RL addresses these failures by allowing the preferred objective to be determined locally. This enables the model to satisfy stringent constraints and recover expert behaviors that are typically lost to the averaging effect of global regularization.

Our contributions are as follows:
\begin{enumerate}[leftmargin=*, nosep]
\item \textbf{Combining rule-based logic with RL.} We formalize a guidance-based framework where rules identify the preferred objective. This lets the learner incorporate rule-based logic while retaining a learned policy.
\item \textbf{Split-RL.} We propose Split-RL\footnote{The Split-RL code is available at \url{https://github.com/NVlabs/split_rl}.}
, a GBT-based algorithm that routes contradictory updates into specialized leaves to prevent premature aggregation. We show how early aggregation loses objective-specific information and when Split-RL isolates conflicting updates.
\item \textbf{State-of-the-art results in two distinct problem families.} Across three domains, only Split-RL meets the mean episodic cost budget in all 15 constrained scenarios. It achieves the highest reward within budget in 11. In the offline setting, Split-RL achieves the best return at every dataset size and reaches near-expert performance. Exposing the same labels to the baselines reproduces neither result.
\end{enumerate}

\section{Related Work}

We position Split-RL relative to constrained RL, multi-objective learning and gradient conflict, GBT methods in RL, and offline learning from mixed-quality data.
\paragraph{Constrained RL.} Constrained RL methods optimize rewards while controlling expected costs, typically falling into broad classes such as Lagrangian-based penalty updates \citep{ppo_lag, tessler2018reward}, trust-region or interior-point strategies \citep{cpo, liu2020ipo, yang2022constrained}, and feasibility-maintaining methods through projection, shielding, action masking, or barrier functions \citep{yang2020projection, alshiekh2018shielding, huang2020invalid, cheng2019end}. Broader research examines safety-preserving transformations, reachability analysis, and extreme value theory for rare violations \citep{sootla2022saute, yu2022reachability, garcia15a, brunke2022safe, gu2024review, ijcai2024p913,gcpo, gao25v}. Split-RL is complementary to these optimization-focused frameworks: rather than enforcing safety through global multipliers or projections, it uses guidance labels to structurally isolate contradictory updates at the state-space level.

\paragraph{Multi-objective RL and gradient conflict.}
Multi-objective RL studies problems with vector-valued returns or learning signals, where policies are optimized under scalarizations, preferences, or Pareto criteria~\citep{roijers2013survey,hayes2022practical}. However, in shared parametric models, disparate objectives can induce incompatible updates. Consequently, multi-task and multi-objective learning research focuses on loss balancing, gradient normalization, and gradient projection techniques, such as PCGrad~\citep{pcgrad}, to mitigate interference~\citep{chen2018gradnorm, LIN2026108317, sener2018multi, navon2022multi, liu2021conflict, he2024multibalance}. In RL specifically, interference under shared policies has been shown to significantly degrade control performance~\citep{liu23a, mao2025a}. Architectural routing methods such as soft modularization and Mixture-of-Experts provide neural alternatives for assigning inputs to specialized subnetworks~\citep{yang2020multi,moe}. Split-RL offers a distinct alternative by using the partitioning structure of decision trees. Rather than applying auxiliary gradient projections or training a separate routing network, Split-RL integrates guidance labels directly into the tree-building process to isolate contradictory updates within disjoint regions of the state-space.

\paragraph{GBT in RL.} GBT is a standard supervised-learning model~\citep{functional_gd,FRIEDMAN2002367}, but its use in RL is limited. Prior work includes boosting function approximators for Q-learning~\citep{abel2016exploratory}, boosting ensemble policies for contextual bandits~\citep{london23a}, and weak-learning reductions for RL~\citep{brukhim2023boosting}. \citet{fuhrer2025gradient} recently introduced GBRL, a GBT framework for actor-critic RL. Split-RL builds on it but changes the role of tree construction: guidance-specific targets are scored separately during split selection and combined only after the routing structure is fixed. This lets the model partition the state-space based on conflicting objectives rather than just fitting an aggregated gradient.

\paragraph{Offline RL from mixed-quality data.}
Offline RL learns from fixed datasets containing trajectories with varying quality. A central challenge is to improve upon the behavior policy while avoiding extrapolation error~\citep{fujimoto2019bcq, kumar2019bear, wu2019brac, kumar2020conservativeqlearningofflinereinforcement, kostrikov2021iql}. In mixed-quality datasets, the optimal balance between behavior cloning (BC) and improvement is often state-dependent~\citep{cai2023td, wolczyk24a}. Minimalist approaches like TD3+BC~\citep{fujimoto2021minimalist} and SPOT~\citep{wu2022supported} weight a BC term using a global coefficient, while DICE-based methods re-weight trajectories based on expert-likeness~\citep{kim2022demodice, jang2023safedice}. Recent work has explored ``imitating the good and avoiding the bad'' using expert and undesirable demonstrations~\citep{hoang2025learning, xudong24a, jiang2025valuealigned}. Split-RL provides a complementary structural mechanism: it uses data provenance as a guidance label to decide locally whether to apply BC or RL updates during tree construction.

\section{Preliminaries}
\label{sec:preliminaries}
This section establishes the formal framework for RL with multiple objectives and defines the additive ensemble structure used for policy parameterization with GBT.
\subsection{RL with Multiple Objectives}
We consider an agent interacting with an environment modeled as a fully observable infinite-horizon discounted MDP, \(\mathcal{M}=(\mathcal{S},\mathcal{A},P,\mathbf{r},\gamma)\), where \(\mathbf{r}:\mathcal{S}\times\mathcal{A}\to\mathbb{R}^M\) is a vector-valued reward function. At each time step \(t\), the agent observes state \(\state_t\), selects an action \(a_t\sim\pi(\cdot\mid s_t)\) according to the policy  \(\pi\), and transitions to \(\state_{t+1}\sim P(\cdot\mid \state_t,\action_t)\) while receiving the reward vector \(\mathbf{r}_t=\mathbf{r}(\state_t,\action_t)\).

We consider \(K\) learning objectives \(J_1(\pi_F),\dots,J_K(\pi_F)\). The number of reward components \(M\) need not equal the number of objectives \(K\): some objectives are driven by the environment, such as the maximization of rewards, while others are defined independently of the reward, such as BC. For any reward component \(m\in\{1,\dots,M\}\), let \(G_t^{(m)}=\sum_{\ell=0}^{\infty}\gamma^\ell r_{t+\ell}^{(m)}\)

be its discounted return. The advantage is \(A_\pi^{(m)}(s,a)=Q_\pi^{(m)}(s,a)-V_\pi^{(m)}(s)\), where
\[
Q_\pi^{(m)}(s,a)=\mathbb{E}_{\pi}\!\left[G_t^{(m)}\mid s_t=s,a_t=a\right],\qquad
V_\pi^{(m)}(s)=\mathbb{E}_{\pi}\!\left[G_t^{(m)}\mid s_t=s\right].
\]
We use GBT policies because tree construction partitions the state space before assigning leaf updates, a natural mechanism for isolating localized conflicts. Although this formulation is stated for general \(K\), in this work we focus on the binary case, where \(K=2\).

\subsection{GBT Policies}
\label{subsec:gbt_prelim}

Split-RL builds on the GBRL framework of \citet{fuhrer2025gradient}.
Following the functional-gradient view of boosting~\citep{functional_gd,FRIEDMAN2002367}, an additive tree ensemble parameterizes the policy:
\begin{equation}
F_T(s)
=
F_0(s)
+
\sum_{t=1}^{T}\eta_t f_t(s),
\end{equation}
where each \(f_t\) is a decision tree with vector-valued leaves, \(\eta_t>0\) is a step size, and \(F_T\) are action logits or distribution parameters of a policy we denote by \(\pi_{F_T}\). At boosting round \(t\), the functional gradient for a single objective \(J\) is expressed as:
\begin{equation}
\nabla_{F_{t-1}}J(\pi_{F_{t-1}})
=
\mathbb{E}_{\pi_{F_{t-1}}}
\left[
\nabla_{F_{t-1}}\log \pi_{F_{t-1}}(a\mid s)\,A(s,a)
\right].
\end{equation}
The next tree is then chosen by projecting this functional gradient onto the tree class, and the ensemble is updated additively:
\begin{equation}
f_t=\arg\min_{f}\left\|\nabla_{F_{t-1}}J(\pi_{F_{t-1}}) - f\right\|_2^2,
\qquad
F_{t}(s)=F_{t-1}(s)+\eta_t f_t(s).
\end{equation}

Crucially, tree construction determines the state-space partition before assigning an update to each leaf. Standard GBT fitting expects a single gradient target per sample. With multiple objectives, their gradients must therefore be combined before candidate splits are scored. This aggregation can reduce the split signal precisely when the objectives disagree, as we show below.

\paragraph{Premature aggregation.}
Let \(L,R\) be the children of a candidate split \(\psi\), and let
\(\mu_k(A)=\frac{1}{|A|}\sum_{i\in A}g_{i,k}\) be the mean gradient of objective \(k\)
over \(A\). Objective \(k\) contributes to \(\psi\) the signal
\(u_k(\psi;\Node)=\bigl(\sqrt{|L|}\,\mu_k(L),\,\sqrt{|R|}\,\mu_k(R)\bigr)\).
Scoring each objective separately with an L2 score gives
\(S_k(\psi;\Node)=\|u_k(\psi;\Node)\|_2^2\), whereas standard GBT fitting combines the
gradients with weights \(\alpha_k\ge0\), \(\sum_k\alpha_k=1\), into a single target
before any split is scored:
\(
S_{\mathrm{agg}}(\psi;\Node)
=
\Bigl\|\sum_k\alpha_k u_k(\psi;\Node)\Bigr\|_2^2 .
\)

\begin{proposition}[Early aggregation loses objective-specific information]
\label{prop:linear_aggregation_loss}
For any such weights,
\[
\textstyle
\sum_k\alpha_k S_k(\psi;\Node)-S_{\mathrm{agg}}(\psi;\Node)
=
\sum_{k<\ell}\alpha_k\alpha_\ell
\|u_k(\psi;\Node)-u_\ell(\psi;\Node)\|_2^2
\ge 0.
\]
\end{proposition}

The gap is exactly the pairwise disagreement between objectives: it vanishes when they agree and grows as they diverge. Split-RL therefore keeps objective-specific gradients separate during tree construction. The identity follows by expanding the aggregated update's squared norm. We give the proof and the cosine split-score analogue in Appendix~\ref{app:split_proofs}.


\section{Split-RL: Localized Signal Isolation via Guided Partitioning}
\label{sec:split}

At each boosting round \(t\), each objective \(J_k\) provides a per-sample gradient vector
\(g_{i,k}\in\mathbb{R}^d\), \(k\in\{1,\dots,K\}\). Guidance is supplied by a label function
\(y:\mathcal{S}\to\{1,\dots,K\}\), assigning each state to an objective.

At a node \(\Node\), the prevalence of label \(k\) is
\(p_k(\Node)=\frac{1}{|\Node|}\sum_{i\in\Node}\mathbb{I}[y(s_i)=k]\).
For a candidate split \(\psi\), Split-RL computes a per-objective split score
\(S_k(\psi;\Node)\) and combines the scores by local label prevalence:
\[
\bar S(\psi;\Node)=\sum_{k=1}^K p_k(\Node)\,S_k(\psi;\Node).
\]
For example, in offline RL, the BC score is computed purely from expert-labeled samples
and the policy-improvement score from exploratory samples.

To resolve localized conflicts, we evaluate how effectively a candidate split \(\psi\)
separates labels when partitioning \(\Node\) into children \(L\) and \(R\).
Letting \(n_{A,k}\) be the number of label-\(k\) samples in subset \(A\), we use the
categorical sum-of-squares error:
\[
\mathrm{SSE}(A)=|A|-\frac{1}{|A|}\sum_{k=1}^K n_{A,k}^2,
\qquad
\mathrm{SSE}(\emptyset)=0.
\]
We define the normalized label impurity after split \(\psi\) as
\[
H(\psi;\Node)
=
\frac{\mathrm{SSE}(L)+\mathrm{SSE}(R)}
{\mathrm{SSE}(\Node)+\epsilon}.
\]
\(H(\psi;\Node)\) measures how effectively a split isolates guidance labels:
\(H=0\) for a label-pure split and \(H\to1\) for a completely mixed split.

However, explicitly forcing purity is potentially harmful when gradients are aligned.
To balance this, we define the node conflict coefficient
\[
\rho(\Node)=
\left[
1-
\frac{
\left\|\sum_{k=1}^{K}\mu_k(\Node)\right\|_2^2
}{
\sum_{k=1}^{K}\|\mu_k(\Node)\|_2^2
}
\right]_+,
\]
where
\(\mu_k(\Node)=\frac{1}{|\Node|}\sum_{i\in\Node}g_{i,k}\)
is the mean gradient vector for objective \(k\).
The coefficient \(\rho(\Node)\) is high when objective gradients cancel and low when
they align, making the purity regularization conflict-dependent.

The Split-RL split score is then:
\begin{equation}
S_{\mathrm{Split}}(\psi;\Node)
=
\bar S(\psi;\Node)
\left(
1-\operatorname{clip}_{[0,1]}
\left(\lambda_0\rho(\Node)H(\psi;\Node)\right)
\right),
\qquad
\lambda_0\ge0.
\label{eq:split_score}
\end{equation}
Here, the single hyperparameter \(\lambda_0\) controls the purity regularization strength.
We now show the threshold at which a label-pure split overtakes an impure one.

\begin{proposition}[Regularization threshold for a purifying split]
\label{prop:purifying_selection}
Let \(\rho(\Node)>0\) and let \(\psi_{\mathrm{pure}}\) satisfy
\(H(\psi_{\mathrm{pure}};\Node)=0\) and
\(\bar S(\psi_{\mathrm{pure}};\Node)>0\).
Then \(\psi_{\mathrm{pure}}\) scores at least as high as every impure candidate at
\(\Node\) whenever
\[
\lambda_0>
\max_{\substack{
\psi:\,H(\psi;\Node)>0,\\
\bar S(\psi;\Node)>\bar S(\psi_{\mathrm{pure}};\Node)
}}
\frac{
1-\bar S(\psi_{\mathrm{pure}};\Node)/\bar S(\psi;\Node)
}{
\rho(\Node)H(\psi;\Node)
}.
\]
\end{proposition}

The maximum is over an empty set when no impure candidate has a higher raw score, so any
\(\lambda_0\ge0\) suffices in that case.

Once the tree topology is fixed, Split-RL computes the final leaf update as a weighted average across labels:
\[
v(\Leaf)=\sum_{k=1}^K p_k(\Leaf)\,\mu_k(\Leaf),
\]
where \(\mu_k(\Leaf)\) is the mean gradient for objective \(k\).
Because aggregation now occurs within a leaf, its label purity determines how closely the applied update follows a particular objective.

\begin{proposition}[Purer leaves track the guided objective]
\label{prop:pure_leaf_update}
For any label \(j\),
\[
\|v(\Leaf)-\mu_j(\Leaf)\|_2
\le
\sum_{k\neq j} p_k(\Leaf)\,
\|\mu_k(\Leaf)-\mu_j(\Leaf)\|_2,
\]
\textup{so a leaf that is pure for \(j\) applies exactly \(\mu_j(\Leaf)\).}
\end{proposition}

Together, these propositions characterize how Split-RL preserves objective-specific signals during tree construction and turns purer partitions into updates that are better aligned with the guided objective. We prove all three propositions in Appendix~\ref{app:split_proofs}.
\section{Experiments}
\label{sec:experiments}
We evaluate Split-RL in settings where objective conflict is localized and observable. In constrained RL, the signals are reward and cost, and guidance labels mark states where cost-sensitive updates should dominate. In offline RL, the signals are BC and policy improvement, and labels indicate expert versus non-expert provenance. All label rules are fixed before training and depend only on the observable state or transition provenance.  Full environment definitions, label rules, costs, and hyperparameters are in Appendices~\ref{app:minigrid_envs}--\ref{app:hyperparameters}.

\subsection{Constrained RL}
\label{sec:exp_constrained_rl}
We implement Split-RL within PPO~\citep{schulman2017proximal}, adding a cost critic for constrained tasks. We compare against neural CPO~\citep{cpo}, CUP~\citep{yang2022constrained}, IPO~\citep{liu2020ipo}, and PPO-Lagrangian~\citep{ppo_lag}, and include a GBT PPO-Lagrangian baseline to isolate the effect of guided partitioning.

\paragraph{MiniGrid.}
We construct three controlled MiniGrid tasks~\citep{MinigridMiniworld23} to isolate localized signal conflict. DynamicCrossing creates a reward--cost routing conflict around a lava barrier with a safe opening, where a moving obstacle near the opening adds timing difficulty. FragileCrossing creates a context-dependent hazard, where ice is dangerous only while the agent carries an object. Corner creates a sparse localized conflict by placing one reward behind a narrow lava entrance.

\paragraph{Traffic signal control.}
We create traffic-signal control scenarios using SUMO-RL~\citep{sumorl}. We use single-intersection, grid, and arterial networks, and introduce bus, premium, and convoy priority events. These scenarios create localized priority conflicts within ordinary traffic: the controller must sometimes sacrifice throughput to serve a specific vehicle class. Special vehicles are injected in half of the episodes, producing infrequent event-based cost signals.

\paragraph{Building energy management.}
We create reserve-control scenarios using CityLearn~\citep{citylearn}. A central agent controls battery storage across three buildings with solar generation over one month of hourly data. The scenarios create temporal conflicts between electricity-cost minimization and reserve maintenance, with labels marking deterministic reserve-critical windows that require sustained anticipatory behavior.

\paragraph{Ablations and interpretability.}
We test whether Split-RL's advantage comes from access to the guidance label by giving the same label to all five constrained-RL baselines, either as an extra observation feature (\emph{Label-in-Obs}) or as a loss mask selecting each sample's objective (\emph{Masked}). We ablate the label-purity regularization strength \(\lambda_0\) in CityLearn Solar Ramp Reserve to isolate the role of the purity penalty. To test robustness to imperfect guidance, we corrupt the label on MiniGrid: each label is independently redrawn uniformly at random with probability \(p\) before it reaches split selection. Finally, we inspect the trained trees on MiniGrid, whose states are easy to visualize as a grid. At tree construction, each leaf stores the proportion of training samples per label. We run evaluation episodes, pass every visited state through the ensemble, and average the stored label proportions across the leaves reached, which we call the \emph{routing density}.

\subsection{Offline RL}
\label{sec:exp_offline}
The offline experiments test whether Split-RL can imitate expert transitions while using non-expert transitions for coverage and policy improvement. Split-RL uses transition provenance as the guidance label to switch between BC targets for expert data and policy-improvement targets for non-expert data. For policy improvement, we use Advantage-Weighted Regression (AWR)~\citep{awr}, which weights supervised regression by exponentiated advantages to provide a stable, off-policy actor-critic target that is a natural fit for decision trees. We evaluate Split-RL across varying dataset sizes of Atari RAM~\citep{bellemare2013arcade}. Baselines include CQL~\citep{kumar2020conservativeqlearningofflinereinforcement}, BCQ~\citep{fujimoto2019bcq}, Decision Transformer (DT)~\citep{chen2021decisiontransformerreinforcementlearning}, and AWR. To examine the role of structural partitioning, we also compare Split-RL against three update-sharing ablations: (i) \textbf{All}, where both BC and AWR targets are computed and averaged for every transition; (ii) \textbf{Masked}, where objectives are strictly partitioned by label, applying BC only to expert transitions and AWR only to non-expert; and (iii) \textbf{Expert-Only BC}, where AWR is applied globally but BC is restricted exclusively to expert transitions. For all experiments, we report the expert-normalized return.
\section{Results}
\label{sec:results}
We report the full results for constrained and offline RL in Appendix~\ref{app:detailed_results}. Constrained RL results are averaged over 15 seeds per configuration and offline results over 50 evaluation episodes (Appendix~\ref{app:hyperparameters}).

\subsection{Constrained RL}
Unlike standard constrained RL, which maximizes reward subject to a fixed episodic cost budget, Split-RL uses a per-state guidance label indicating when the cost signal should dominate the local update. This results in Split-RL being the only method feasible in all scenarios, as shown in Table~\ref{tab:constraint_feasibility_summary}.


\begin{table}[thb!]
\centering
\caption{\textbf{Constrained RL summary.}
Domain columns count feasible scenarios; i.e., mean episodic cost below the limit. The final column counts scenarios where a method has the highest reward among feasible methods; ties are counted for all tied methods.}
\label{tab:constraint_feasibility_summary}
\small
\begin{tabular}{lccccc}
\toprule
\textbf{Method} & \textbf{MiniGrid} & \textbf{SUMO-RL} & \textbf{CityLearn} & \textbf{Feasible} & \textbf{Best feasible reward} \\
\midrule
CPO & 0/3 & 7/9 & 0/3 & 7/15 & 0/15 \\
CUP & 2/3 & 8/9 & 0/3 & 10/15 & 0/15 \\
IPO & 0/3 & 8/9 & 0/3 & 8/15 & 0/15 \\
PPO-Lag & 1/3 & 4/9 & 0/3 & 5/15 & 3/15 \\
PPO-Lag (GBT) & \textbf{3/3} & 3/9 & 0/3 & 6/15 & 1/15 \\
Split-RL (Ours) & \textbf{3/3} & \textbf{9/9} & \textbf{3/3} & \textbf{15/15} & \textbf{11/15} \\
\bottomrule
\end{tabular}
\end{table}

\paragraph{MiniGrid.} Split-RL matches the best feasible operating point on DynamicCrossing and gives the strongest reward--cost profile on FragileCrossing. Corner is the main exception: Split-RL remains safe, but PPO-Lagrangian (GBT) attains a higher feasible reward. These results suggest that Split-RL is most useful when feasibility requires a state- or context-dependent switch between progress-seeking and cost-sensitive updates. When a conservative feasible policy is already easy to represent, as in Corner, the benefit of guided local routing is smaller.

\begin{figure}[thbp!]
\centering
    \includegraphics[width=\linewidth]{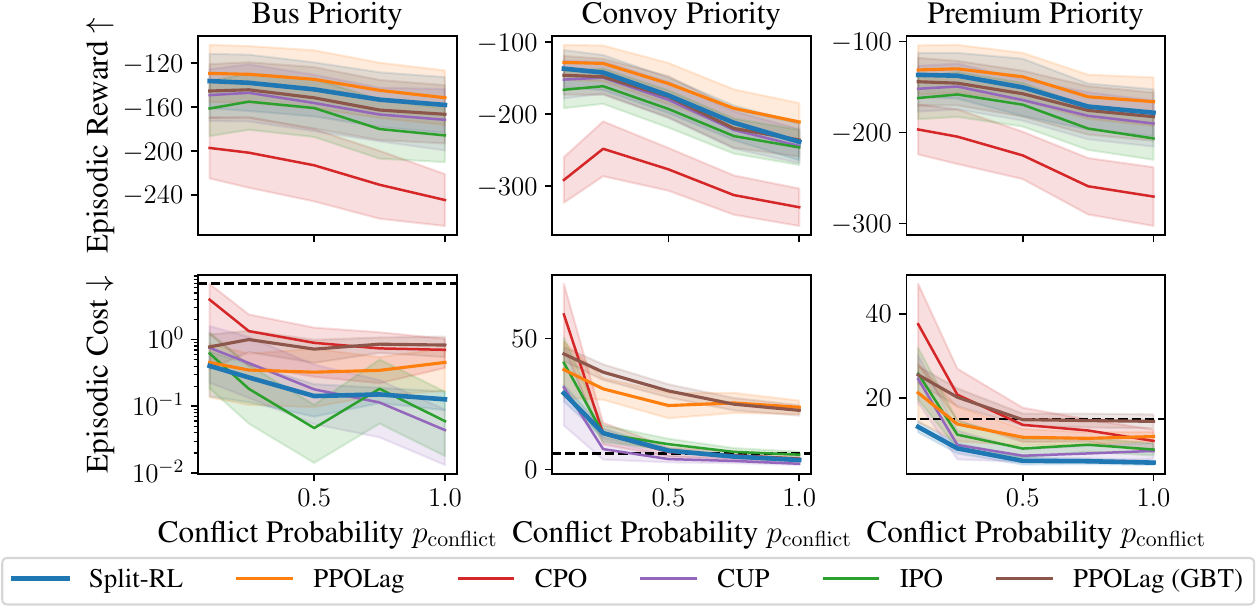}
    \caption{\textbf{Effect of priority-event frequency in SUMO-RL 4x4Grid network.}
Mean episodic reward (top) and episodic cost (bottom) as conflict probability \(p_{\mathrm{conflict}}\) increases. Shading shows \(\pm1\) std. Dashed lines mark cost limits. Split-RL keeps cost low even when conflicts are rare, while maintaining competitive throughput.}
    
    \label{fig:sumo_frequency}
\end{figure}

\paragraph{SUMO-RL.}
Split-RL reliably produces feasible priority-aware policies, achieving feasibility in all nine network--event pairs, the lowest cost in four, and the highest feasible reward in every Arterial scenario (Appendix~\ref{app:detailed_results}). As shown in Figure~\ref{fig:sumo_frequency}, Split-RL maintains low cost even when priority events are rare and remains feasible across all tested conflict probabilities in the premium scenario.

\paragraph{CityLearn.} CityLearn exposes the clearest reward--feasibility tradeoff. Split-RL is feasible in all three scenarios, while no baseline is feasible in any.

Across all three domains, Split-RL substantially outperforms PPO-Lagrangian (GBT), showing that the tree function class alone does not explain the gains.

\paragraph{Does Split-RL have access to privileged information?}
Giving the baselines the label does not produce a systematic gain (Table~\ref{tab:update_sharing_ablation}). Under either condition, every baseline changes by at most one feasible scenario except CPO under Masked, which gains four. These gains are not consistent across scenarios: masking may reduce competing contributions to CPO's constrained update, stabilizing high-variance runs and trading reward for feasibility in some scenarios while degrading others. PPO-Lag (GBT), which shares Split-RL's tree backbone and receives the same labels, loses a feasible scenario under both conditions. Full reward--cost results are in Appendix~\ref{app:ablation_raw}. Overall, label access does not reproduce Split-RL's advantage.

\begin{table}[htb!]
\centering
\caption{\textbf{Does access to the label alone explain Split-RL's gap?} Feasible scenario counts
out of 15 (cost \(\le\) limit, Table~\ref{tab:evaluation_cost_limits}) for each baseline under
three training conditions.}
\label{tab:update_sharing_ablation}
\small
\begin{tabular}{lccc}
\toprule
\textbf{Algorithm} & \textbf{No-label} & \textbf{Label-in-Obs} & \textbf{Masked} \\
\midrule
CPO & 7/15 & 7/15 & 11/15 \\
CUP & 10/15 & 9/15 & 9/15 \\
IPO & 8/15 & 9/15 & 9/15 \\
PPOLag & 5/15 & 5/15 & 6/15 \\
PPOLag (GBT) & 6/15 & 5/15 & 5/15 \\
\bottomrule
\end{tabular}
\end{table}

\begin{figure}[thb!]
\centering
    \includegraphics[width=\textwidth]{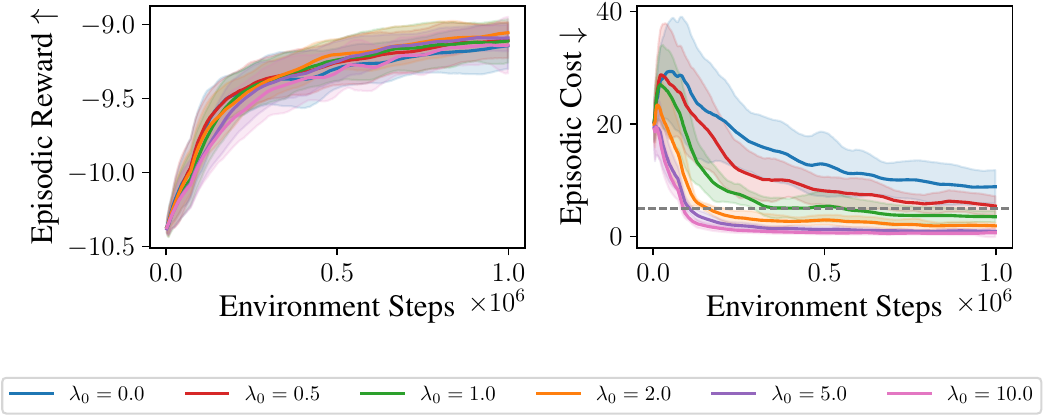}
\caption{\textbf{Label-purity regularization ablation in CityLearn Solar Ramp Reserve.}
Mean episodic reward and cost over training for different \(\lambda_0\). Shading: \(\pm1\) std. Dashed line: cost limit. Nonzero \(\lambda_0\) substantially reduces cost by penalizing label-mixing splits in conflict windows.}
    \label{fig:lambda_penalty_citylearn}
\end{figure}

\paragraph{Does the purity penalty matter?}
Figure~\ref{fig:lambda_penalty_citylearn} shows that it does. Without the penalty (\(\lambda_0=0\)), or with a small \(\lambda_0\), Split-RL remains above the cost limit in Solar Ramp Reserve. Once \(\lambda_0\) is sufficiently large, it becomes feasible without a loss in reward. This mirrors the threshold behavior in Proposition~\ref{prop:purifying_selection}: increasing \(\lambda_0\) can make a label-pure split overtake an impure one when the objectives conflict.

\begin{figure}[htb!]
\centering
    \includegraphics[width=\textwidth]{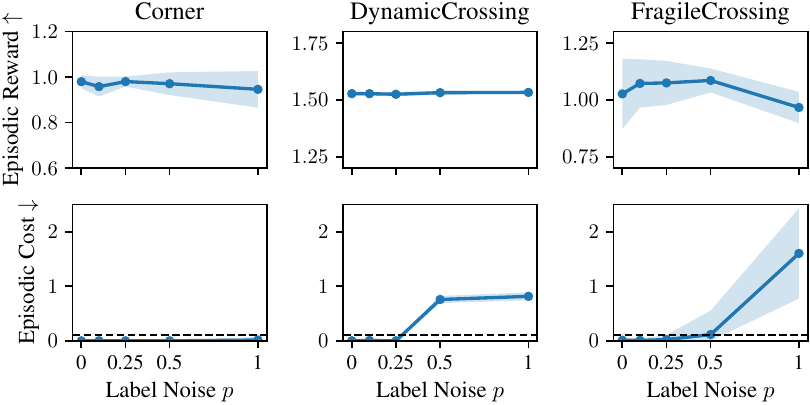}
\caption{\textbf{Robustness to label noise.} Mean episodic reward and cost (\(\pm1\) std)
on MiniGrid. Dashed line: cost limit. Split-RL's performance is stable up to \(p=0.25\).}
    \label{fig:minigrid_noise}
\end{figure}

\paragraph{How robust is Split-RL to label noise?} Performance is stable up to \(p=0.25\): cost stays near its noise-free value, and reward is essentially unchanged across the whole sweep (Figure~\ref{fig:minigrid_noise}). Cost rises sharply once \(p\geq0.5\). This matches the failure mode Split-RL targets: cost is the rare but critical signal that the labels protect from dilution. An uninformative label produces arbitrary routing rather than a fallback to unguided training.

\begin{figure}[htb!]
\centering
    \includegraphics[width=\linewidth]{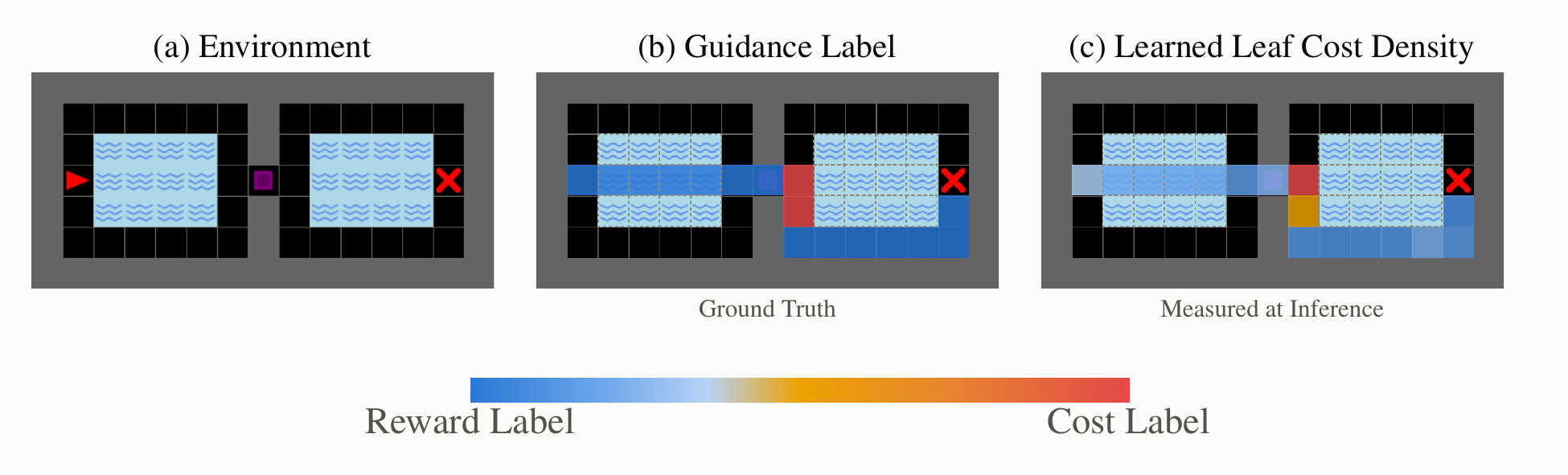}
    \caption{\textbf{Guided routing density under policy rollouts (FragileCrossing).} Fraction of
    cost-labelled training samples in the leaves reached by each visited state, over evaluation
    episodes. Higher density (red) on labelled-cost states shows the ensemble routes by conflict
    structure.}
    \label{fig:interp_main}
\end{figure}

\paragraph{Do the trees route by conflict structure?}
In FragileCrossing, cost-labelled states (Figure~\ref{fig:interp_main}b) reach leaves with high cost-label density (Figure~\ref{fig:interp_main}c). The trees recover the conflict structure from the observation alone, without the label at evaluation. Appendix~\ref{app:interpretability} shows the other two environments.

\subsection{Offline RL}
As shown in Table~\ref{tab:atari_budget_profile}, Split-RL achieves the highest expert-normalized return on average, outperforming DT by \(43\%\) and GBT AWR by \(86\%\). It is near-expert from 50K transitions onward, while standard baselines struggle with the imitation--improvement tradeoff.

\begin{table*}[htb!]
\centering
\caption{\textbf{Atari RAM offline learning: data-budget profile.} Mean $\pm$ std of expert-normalized return across 11 Atari games at each offline dataset size. \textbf{Bold} = best per column. Split-RL produces the best return at every dataset size and exhibits near-expert performance on average.}
\label{tab:atari_budget_profile}
\vspace{0.2em}
\resizebox{\textwidth}{!}{%
\begin{tabular}{llccccccc}
\toprule
 & & \multicolumn{6}{c}{\textbf{Dataset Size}} & \\ 
 \cmidrule(lr){3-8}
 & \textbf{Method} & \textbf{10K} & \textbf{20K} & \textbf{50K} & \textbf{100K} & \textbf{200K} & \textbf{Full} & \textbf{Avg.} \\
\midrule
 \multirow{4}{*}{\rotatebox[origin=c]{90}{\textbf{NN}}} 
   & CQL & 0.27 $\pm$ 0.27 & 0.24 $\pm$ 0.23 & 0.27 $\pm$ 0.30 & 0.35 $\pm$ 0.32 & 0.42 $\pm$ 0.28 & 0.42 $\pm$ 0.29 & 0.33 \\
   & DT  & 0.52 $\pm$ 0.27 & 0.58 $\pm$ 0.32 & 0.71 $\pm$ 0.35 & 0.73 $\pm$ 0.35 & 0.65 $\pm$ 0.35 & 0.69 $\pm$ 0.40 & 0.65 \\
   & AWR & 0.43 $\pm$ 0.37 & 0.53 $\pm$ 0.33 & 0.58 $\pm$ 0.31 & 0.60 $\pm$ 0.37 & 0.59 $\pm$ 0.30 & 0.67 $\pm$ 0.26 & 0.57 \\
   & BCQ & 0.35 $\pm$ 0.36 & 0.34 $\pm$ 0.29 & 0.24 $\pm$ 0.33 & 0.22 $\pm$ 0.27 & 0.18 $\pm$ 0.25 & 0.33 $\pm$ 0.33 & 0.28 \\
\midrule
 \multirow{2}{*}{\rotatebox[origin=c]{90}{\textbf{GBT}}} 
   & AWR & 0.48 $\pm$ 0.30 & 0.57 $\pm$ 0.30 & 0.52 $\pm$ 0.32 & 0.48 $\pm$ 0.32 & 0.50 $\pm$ 0.35 & 0.46 $\pm$ 0.28 & 0.50 \\
   & Split-RL (Ours) & \textbf{0.75 $\pm$ 0.32} & \textbf{0.86 $\pm$ 0.26} & \textbf{0.93 $\pm$ 0.13} & \textbf{1.02 $\pm$ 0.05} & \textbf{1.02 $\pm$ 0.04} & \textbf{0.99 $\pm$ 0.08} & \textbf{0.93} \\
\bottomrule
\end{tabular}%
}
\end{table*}

As shown in Table~\ref{tab:atari_ablation}, where the BC target is applied matters, but it is not sufficient. Applying BC to all samples hurts performance because it clones non-expert behavior. Restricting BC to expert samples gives the strongest AWR+BC baseline, while the fully masked variant is too restrictive. Split-RL still performs best. Masking prevents the two objectives from acting on the same transition, but only Split-RL also keeps their split scores separate during tree construction, preventing premature aggregation.

\begin{table}[hbt!]
\centering
\caption{\textbf{BC target ablation.} Mean \(\pm\) std of expert-normalized return across 11 Atari games. \emph{All}: both signals on all samples; \emph{Expert}: BC on expert samples only; \emph{Masked}: BC on expert, AWR on non-expert exclusively. \textbf{Bold} = best.}
\label{tab:atari_ablation}
\vspace{0.2em}
\resizebox{\textwidth}{!}{%
\begin{tabular}{lccccccc}
\toprule
& \multicolumn{3}{c}{\textbf{NN: AWR + BC}} & \multicolumn{3}{c}{\textbf{GBT: AWR + BC}} & \textbf{Split-RL (Ours)} \\
\cmidrule(lr){2-4} \cmidrule(lr){5-7} \cmidrule(lr){8-8}
& All & Expert & Masked & All & Expert & Masked & \\
\midrule
  Return & 0.68 \(\pm\) 0.28 & 0.81 \(\pm\) 0.29 & 0.59 \(\pm\) 0.42 & 0.68 \(\pm\) 0.29 & 0.88 \(\pm\) 0.25 & 0.52 \(\pm\) 0.35 & \textbf{0.99 \(\pm\) 0.08} \\
\bottomrule
\end{tabular}%
}

\end{table}


\section{Limitations}
Split-RL assumes guidance labels are deterministic and aligned with the objective conflict. A label unrelated to the conflict, or one missing hidden context, can prevent effective learning. It also targets conflicts that are localized and identifiable, so where objectives conflict globally and uniformly, explicit partitioning may offer little advantage over Lagrangian or projection-based methods. The framework is also tied to GBT policies and structured observations: GBT excels on the tabular inputs studied here, but extending Split-RL to pixel-based inputs would likely require a hybrid architecture in which a representation learner extracts features for the tree-based policy. Finally, our experiments use \(K=2\) objectives. Because each split is binary, routing \(K\) label categories into separate leaves requires a tree depth that grows with \(K\).

\section{Conclusion}
We introduced Split-RL, a framework that reframes localized objective conflict as a problem of update sharing. By integrating guidance labels into tree construction, Split-RL routes contradictory signals into disjoint regions of the state-space before aggregation. Our results show that this structural isolation is useful across distinct problem families: Split-RL resolves the imitation--improvement tradeoff in offline RL and achieves consistent feasibility in constrained environments. Crucially, it incorporates rule-based logic without replacing the learned policy: guidance labels define where different objectives control the update, while the learner still optimizes the policy within those regions. When conflicts are localized, deciding which samples share an update can be as important as optimizing the objective itself.

\begin{ack}
This work was funded by NVIDIA, where all authors are employed. The authors received no third-party funding in support of this work during the past 36 months, and have no financial relationships outside the submitted work that could be perceived as influencing it.
\end{ack}

\bibliographystyle{plainnat}
\clearpage 
{\small
\bibliography{references} 
}

\appendix
\clearpage
\section{Theoretical Properties of Split-RL}
\label{app:split_proofs}
In this section, we formalize three properties of Split-RL. The first shows
that early aggregation can lose objective-specific information; we establish the L2
identity and then give the equivalent comparison for the cosine split score used in
our experiments. The second shows how the regularization strength \(\lambda_0\) can
make a purifying split win when conflict is present. The third shows that purer leaves
make the applied update follow the guided objective update more closely.

Let \(L,R\) be the children of a candidate split \(\psi\) at node \(\Node\), and let
\(\mu_k(A)=\frac{1}{|A|}\sum_{i\in A}g_{i,k}\) be the mean gradient of objective \(k\)
over \(A\). Objective \(k\) contributes to \(\psi\) the signal
\begin{equation*}
u_k(\psi;\Node)=\bigl(\sqrt{|L|}\,\mu_k(L),\ \sqrt{|R|}\,\mu_k(R)\bigr),
\end{equation*}
scored separately by \(S_k(\psi;\Node)=\|u_k(\psi;\Node)\|_2^2\), while aggregating the
gradients with weights \(\alpha_k\ge0\), \(\sum_k\alpha_k=1\), before scoring gives
\begin{equation*}
S_{\mathrm{agg}}(\psi;\Node)=\Bigl\|\sum_{k=1}^K\alpha_k u_k(\psi;\Node)\Bigr\|_2^2 .
\end{equation*}
Below, \(\sum_{k<\ell}\) ranges over distinct unordered pairs of objectives, so each
pair is counted once.

\begin{restatedproposition}{prop:linear_aggregation_loss}
For any such weights,
\begin{equation*}
\textstyle\sum_{k}\alpha_k S_k(\psi;\Node) - S_{\mathrm{agg}}(\psi;\Node)
= \sum_{k<\ell}\alpha_k\alpha_\ell\,\|u_k(\psi;\Node)-u_\ell(\psi;\Node)\|_2^2 \ge 0.
\end{equation*}
\end{restatedproposition}

\begin{proof}
Write \(u_k=u_k(\psi;\Node)\). By definition,
\begin{equation*}
\sum_{k=1}^K\alpha_k S_k(\psi;\Node)=\sum_{k=1}^K \alpha_k\|u_k\|_2^2,
\qquad
S_{\mathrm{agg}}(\psi;\Node)=\left\|\sum_{k=1}^K \alpha_k u_k\right\|_2^2.
\end{equation*}
Expanding the aggregated score gives
\begin{equation*}
S_{\mathrm{agg}}(\psi;\Node)=\sum_{k=1}^K \alpha_k^2\|u_k\|_2^2+2\sum_{k<\ell}\alpha_k\alpha_\ell\,u_k^\top u_\ell.
\end{equation*}
Therefore,
\begin{equation*}
\sum_{k=1}^K\alpha_k S_k(\psi;\Node)-S_{\mathrm{agg}}(\psi;\Node)
=
\sum_{k=1}^K \alpha_k(1-\alpha_k)\|u_k\|_2^2
-
2\sum_{k<\ell}\alpha_k\alpha_\ell\,u_k^\top u_\ell.
\end{equation*}

Now expand the right-hand side we want:
\begin{equation*}
\sum_{k<\ell}\alpha_k\alpha_\ell\,\|u_k-u_\ell\|_2^2
=
\sum_{k<\ell}\alpha_k\alpha_\ell\bigl(\|u_k\|_2^2+\|u_\ell\|_2^2-2u_k^\top u_\ell\bigr).
\end{equation*}
For a fixed \(k\), the term \(\|u_k\|_2^2\) appears once for every \(\ell\neq k\), and its total coefficient is
\begin{equation*}
\sum_{\ell\neq k}\alpha_k\alpha_\ell=\alpha_k\sum_{\ell\neq k}\alpha_\ell=\alpha_k(1-\alpha_k),
\end{equation*}
because \(\sum_{\ell=1}^K \alpha_\ell=1\). Hence
\begin{equation*}
\sum_{k<\ell}\alpha_k\alpha_\ell\bigl(\|u_k\|_2^2+\|u_\ell\|_2^2\bigr)
=
\sum_{k=1}^K \alpha_k(1-\alpha_k)\|u_k\|_2^2.
\end{equation*}
Substituting this back yields
\begin{equation*}
\sum_{k<\ell}\alpha_k\alpha_\ell\,\|u_k-u_\ell\|_2^2
=
\sum_{k=1}^K \alpha_k(1-\alpha_k)\|u_k\|_2^2
-
2\sum_{k<\ell}\alpha_k\alpha_\ell\,u_k^\top u_\ell,
\end{equation*}
which is exactly \(\sum_{k=1}^K\alpha_k S_k(\psi;\Node)-S_{\mathrm{agg}}(\psi;\Node)\).
\end{proof}

\paragraph{Cosine split score.}
Let \(d_k=(\sum_{i\in\Node}\|g_{i,k}\|_2^2)^{1/2}\). Scoring \(\psi\) by the cosine
similarity between the gradients and the child means they are fitted with gives
\begin{equation*}
S_k^{\cos}(\psi;\Node)
=\frac{\sum_{A\in\{L,R\}}\sum_{i\in A}g_{i,k}^{\top}\mu_k(A)}
       {d_k\bigl(\sum_{A\in\{L,R\}}|A|\,\|\mu_k(A)\|_2^2\bigr)^{1/2}}
=\frac{\|u_k(\psi;\Node)\|_2}{d_k},
\end{equation*}
using \(\sum_{i\in A}g_{i,k}^{\top}\mu_k(A)=|A|\,\|\mu_k(A)\|_2^2\) in the numerator.
Scoring the aggregated gradients \(\sum_k\alpha_kg_{i,k}\) in the same way
gives \(S^{\cos}_{\mathrm{agg}}(\psi;\Node)=\|\sum_k\alpha_ku_k(\psi;\Node)\|_2/D\), where
\(D=(\sum_{i\in\Node}\|\sum_k\alpha_kg_{i,k}\|_2^2)^{1/2}\). Writing \(u_k=u_k(\psi;\Node)\)
and expanding the square as in Proposition~\ref{prop:linear_aggregation_loss},
\begin{equation*}
D^2\bigl(S^{\cos}_{\mathrm{agg}}\bigr)^2
=\Bigl(\sum_{k=1}^K\alpha_k\|u_k\|_2\Bigr)^{2}-\Delta_\psi,
\qquad
\Delta_\psi=2\sum_{k<\ell}\alpha_k\alpha_\ell\bigl(\|u_k\|_2\|u_\ell\|_2-u_k^{\top}u_\ell\bigr)\ge0 .
\end{equation*}
Angular disagreement reduces the squared aggregated numerator by \(\Delta_\psi\). As in the L2 case, premature aggregation can conceal objective-specific disagreement through cancellation. Scoring objectives separately preserves their split signals during tree construction. Both comparisons concern the score before the label-purity penalty, which we turn to next.

Split-RL penalizes splits that mix guidance labels. Let \(H(\psi;\Node)\) be the normalized
label impurity after \(\psi\) and \(\rho(\Node)\) the node conflict coefficient, both defined
in Section~\ref{sec:split}, and let
\(\bar S(\psi;\Node)=\sum_k p_k(\Node)S_k(\psi;\Node)\) combine the per-objective scores by
label prevalence \(p_k(\Node)\). The selected split maximizes
\begin{equation*}
S_{\mathrm{Split}}(\psi;\Node)=\bar S(\psi;\Node)
\bigl(1-\operatorname{clip}_{[0,1]}(\lambda_0\rho(\Node)H(\psi;\Node))\bigr),
\qquad \lambda_0\ge0 .
\end{equation*}
Call \(\psi\) \emph{label-pure} if \(H(\psi;\Node)=0\) and \emph{impure} otherwise. The following
result gives the regularization strength at which a pure candidate outscores every impure one.

\begin{restatedproposition}{prop:purifying_selection}
Let \(\rho(\Node)>0\) and let \(\psi_{\mathrm{pure}}\) satisfy \(H(\psi_{\mathrm{pure}};\Node)=0\) and \(\bar S(\psi_{\mathrm{pure}};\Node)>0\). Then \(\psi_{\mathrm{pure}}\) scores at least as high as every impure candidate at \(\Node\) whenever
\begin{equation*}
\lambda_0>\max_{\substack{\psi:\,H(\psi;\Node)>0,\\ \bar S(\psi;\Node)>\bar S(\psi_{\mathrm{pure}};\Node)}}\frac{1-\bar S(\psi_{\mathrm{pure}};\Node)/\bar S(\psi;\Node)}{\rho(\Node)H(\psi;\Node)} .
\end{equation*}
\end{restatedproposition}

\begin{proof}
Write \(m(\psi)=1-\operatorname{clip}_{[0,1]}\bigl(\lambda_0\rho(\Node)H(\psi;\Node)\bigr)\in[0,1]\), so that
\begin{equation*}
S_{\mathrm{Split}}(\psi;\Node)=\bar S(\psi;\Node)\,m(\psi).
\end{equation*}
Since \(H(\psi_{\mathrm{pure}};\Node)=0\) we have \(m(\psi_{\mathrm{pure}})=1\) and
\(S_{\mathrm{Split}}(\psi_{\mathrm{pure}};\Node)=\bar S(\psi_{\mathrm{pure}};\Node)\).
Fix an impure candidate \(\psi\). As \(\bar S(\psi;\Node)=\sum_k p_k(\Node)S_k(\psi;\Node)\ge 0\), if
\(\bar S(\psi;\Node)\le \bar S(\psi_{\mathrm{pure}};\Node)\) then
\begin{equation*}
S_{\mathrm{Split}}(\psi;\Node)=\bar S(\psi;\Node)\,m(\psi)\le \bar S(\psi;\Node)\le \bar S(\psi_{\mathrm{pure}};\Node)
=S_{\mathrm{Split}}(\psi_{\mathrm{pure}};\Node).
\end{equation*}
Otherwise \(\bar S(\psi;\Node)>\bar S(\psi_{\mathrm{pure}};\Node)>0\), and
\(S_{\mathrm{Split}}(\psi_{\mathrm{pure}};\Node)>S_{\mathrm{Split}}(\psi;\Node)\) holds exactly when
\begin{equation*}
\operatorname{clip}_{[0,1]}\bigl(\lambda_0\rho(\Node)H(\psi;\Node)\bigr)
>1-\frac{\bar S(\psi_{\mathrm{pure}};\Node)}{\bar S(\psi;\Node)} .
\end{equation*}
The right-hand side lies in \((0,1)\), so
\(\lambda_0>\bigl(1-\bar S(\psi_{\mathrm{pure}};\Node)/\bar S(\psi;\Node)\bigr)/\bigl(\rho(\Node)H(\psi;\Node)\bigr)\)
suffices. Taking the maximum over such \(\psi\) gives the stated condition.
\end{proof}

Proposition~\ref{prop:purifying_selection} isolates the role of the regularization strength \(\lambda_0\). If a pure split is already competitive in raw score, no penalty is needed. Otherwise, increasing \(\lambda_0\) suppresses impure candidates whose raw score advantage is not large enough to overcome the conflict-weighted impurity penalty.

Once the topology is fixed, the leaf update combines the per-objective mean gradients by label
prevalence,
\begin{equation*}
v(\Leaf)=\sum_{k=1}^K p_k(\Leaf)\,\mu_k(\Leaf) ,
\end{equation*}
and \(\Leaf\) is \emph{pure} for label \(j\) when \(p_j(\Leaf)=1\). The following result bounds
how far the applied update can fall from the update of the guided objective.

\begin{restatedproposition}{prop:pure_leaf_update}
For any label \(j\),
\begin{equation*}
\|v(\Leaf)-\mu_j(\Leaf)\|_2
\le
\sum_{k\neq j} p_k(\Leaf)\,
\|\mu_k(\Leaf)-\mu_j(\Leaf)\|_2,
\end{equation*}
\textup{so a leaf that is pure for \(j\) applies exactly \(\mu_j(\Leaf)\).}
\end{restatedproposition}

\begin{proof}
Since \(\sum_{k=1}^K p_k(\Leaf)=1\),
\begin{equation*}
v(\Leaf)-\mu_j(\Leaf)
=\sum_{k=1}^K p_k(\Leaf)\bigl(\mu_k(\Leaf)-\mu_j(\Leaf)\bigr)
=\sum_{k\neq j} p_k(\Leaf)\bigl(\mu_k(\Leaf)-\mu_j(\Leaf)\bigr),
\end{equation*}
the term with \(k=j\) being zero, and the triangle inequality gives the bound. If \(\Leaf\) is pure for \(j\), then \(p_k(\Leaf)=0\) for every \(k\neq j\) and the right-hand side vanishes.
\end{proof}

Proposition~\ref{prop:pure_leaf_update} connects the guidance labels to the actual update applied by the tree. As a leaf becomes purer, its aggregated update moves closer to the update of the guided objective context; in the pure case, the match is exact.

\section{Environment Details}
In this section we describe the details of each environment used in the experiments.
\subsection{MiniGrid Environment Details}
\label{app:minigrid_envs}
\begin{figure}[thbp!]
\centering
    \begin{subfigure}[b]{0.32\linewidth}
        \centering
        \includegraphics[width=\linewidth]{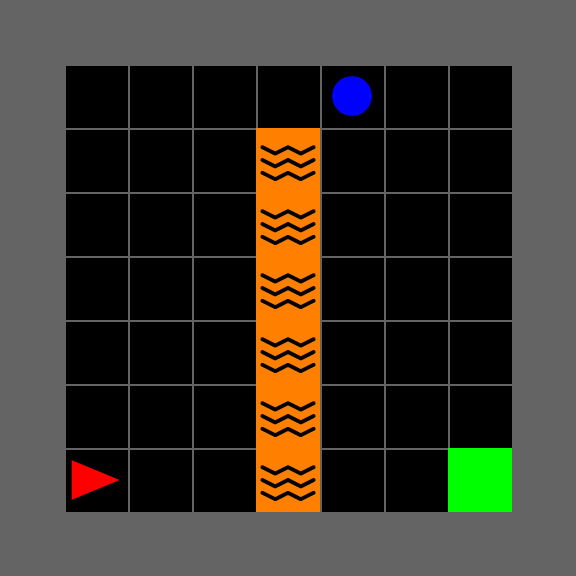}
        \caption{\textbf{DynamicCrossing} - Navigate around a lava river to reach the goal while avoiding a moving obstacle near the crossing.}
        \label{fig:dynamic_crossing_env}
    \end{subfigure}
    \hfill
    \begin{subfigure}[b]{0.32\linewidth}
        \centering
        \includegraphics[width=\linewidth]{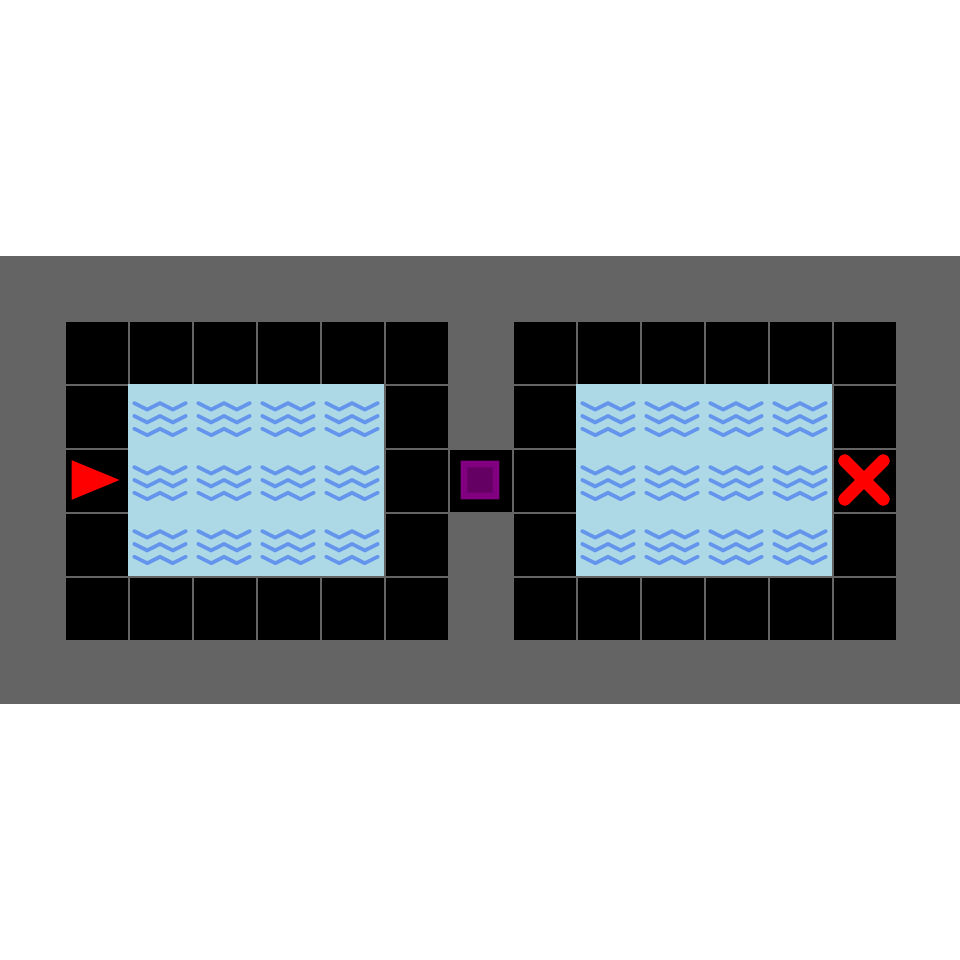}
        \caption{\textbf{FragileCrossing} - Deliver a heavy object to the drop zone; ice (light blue) is hazardous only while carrying.}
        \label{fig:fragile_crossing_env}
    \end{subfigure}
    \hfill
    \begin{subfigure}[b]{0.32\linewidth}
        \centering
        \includegraphics[width=\linewidth]{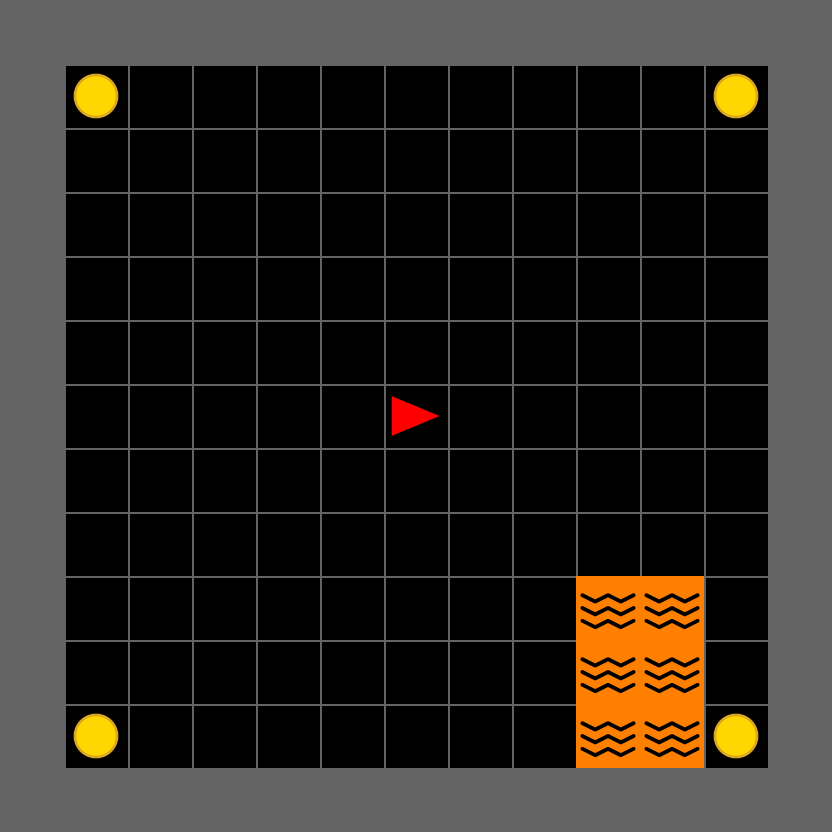}
        \caption{\textbf{Corner} - Collect four coins; the bottom-right corner is surrounded by lava with a single safe entrance.}
        \label{fig:four_corner_env}
    \end{subfigure}
    \caption{\textbf{MiniGrid Environments.} Three custom environments with distinct safety-constraint structures: dynamic hazards, context-dependent constraints, and sparse localized hazards.}
    \label{fig:minigrid}
\end{figure}

The MiniGrid tasks are implemented within the MiniGrid framework~\citep{MinigridMiniworld23} using agent-centric observations that provide a \(7\times 7\) grid field of view and discrete action spaces. Neural network baselines receive a flattened numeric encoding of the grid, while tree-based methods receive categorical observations where each cell is represented as a string token (e.g.\ \texttt{wall}, \texttt{lava\_N}, \texttt{door\_open}), enabling native categorical splits in the tree. Each environment defines a reward for task completion and a binary cost for constraint violations. The evaluation cost limit is \(c \leq 0.1\) for all three environments (see Table~\ref{tab:evaluation_cost_limits}). A visualization of the environments is shown in Figure~\ref{fig:minigrid}.

\subsubsection{DynamicCrossing}

The agent must reach a green goal square in the bottom-right corner of a room. A vertical river of lava (or walls) runs through the center of the grid with a single safe crossing point. Additionally, a moving obstacle (ball) traverses vertically near the crossing, requiring the agent to time its movements.

\paragraph{Action space.} Discrete(4): turn left, turn right, move forward, wait.
\paragraph{Reward.} The reward comprises potential-based shaping and a terminal bonus. Each step, \(r = \phi(s') - \phi(s)\), where \(\phi(s) = \text{clip}(1 - \|s - s_{\text{goal}}\|_2 / d_{\max},\,0,\,1)\). Upon reaching the goal, \(r = 1 - 0.9 \times (\text{step\_count} / \text{max\_steps})\).
\paragraph{Cost.} Binary: \(c=1\) if the agent attempts to move forward into any non-goal obstacle (lava, the moving ball, or a wall).
\paragraph{Label.} \(y(s)=\mathrm{cost}\) when the cell directly in front of the agent is lava or the moving ball; \(y(s)=\mathrm{reward}\) otherwise.
\paragraph{Termination.} The episode ends when the agent reaches the goal, collides with the moving obstacle, or exceeds the step limit. Stepping on lava incurs cost but does not terminate the episode.

\subsubsection{FragileCrossing}

Two rooms are separated by a central wall with a single doorway. The agent must pick up a heavy object in the doorway and deliver it to a drop zone (marked X) in the far room. Both rooms contain horizontal strips of ice that are safe when the agent is empty-handed, but hazardous when carrying the heavy object --- exceeding a configurable number of steps on ice while carrying results in termination.

\paragraph{Action space.} Discrete(5): turn left, turn right, move forward, pickup, drop.
\paragraph{Reward.} Potential-based shaping from Manhattan distance along the mission path from agent to box to goal, plus a terminal bonus for delivery.
\paragraph{Cost.} Binary: \(c=1\) when the agent steps on ice while carrying the heavy object.
\paragraph{Label.} \(y(s)=\mathrm{cost}\) when the agent is carrying the object and is adjacent to or on ice; \(y(s)=\mathrm{reward}\) otherwise.
\paragraph{Termination.} Successful delivery, exceeding the ice-step limit while carrying, or exceeding the total step limit.

\subsubsection{Corner}

The agent starts in the center of a \(13 \times 13\) room and must collect four coins, one in each corner. Three corners are easily accessible, but the bottom-right corner is surrounded by lava with only a single narrow entrance from above.

\paragraph{Action space.} Discrete(4): turn left, turn right, move forward, pickup.
\paragraph{Reward.} \(r = 0.25\) per coin collected; episode success when all four coins are gathered.
\paragraph{Cost.} Binary: \(c=1\) when the agent is standing on a lava cell.
\paragraph{Label.} \(y(s)=\mathrm{cost}\) when the agent is standing on or facing a lava cell; \(y(s)=\mathrm{reward}\) otherwise. The label is more conservative than the cost, firing before the agent steps onto lava.
\paragraph{Termination.} All four coins collected or exceeding the step limit. Stepping on lava incurs cost but does not terminate the episode.

\subsection{SUMO-RL Scenario Details}
\label{app:sumo_scenarios}

\begin{figure}[htbp!]
\centering
    \includegraphics[width=\linewidth]{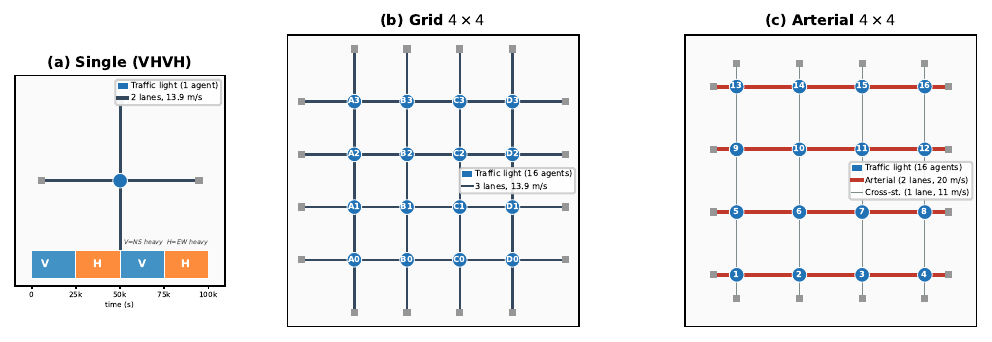}
    \caption{\textbf{SUMO Network Topologies.}
    Blue circles denote traffic-light intersections (agents); grey squares mark boundary entry/exit nodes.
    \textbf{(a)}~Single 4-way intersection (1~agent, \(\text{Discrete}(4)\)) with VHVH non-stationary demand.
    \textbf{(b)}~Grid \(4\times4\): symmetric 16-intersection grid, uniform 3-lane roads, \({\sim}1\,500\)~vehicles, \(\text{Discrete}(8)\).
    \textbf{(c)}~Arterial \(4\times4\): asymmetric 16-intersection grid, EW arterial 2~lanes at 20\,m/s vs.\ NS cross-streets 1~lane at 11\,m/s, \({\sim}2\,500\)~vehicles, \(\text{Discrete}(5)\).}
    \label{fig:sumo_networks}
\end{figure}

SUMO-RL~\citep{sumorl} is a traffic signal control framework built on the SUMO microsimulator~\citep{sumo}.
Control is decentralized; each signalized intersection is controlled by an independent agent that selects the next green phase from a discrete action set.
Observations include per-lane queue lengths, densities, and average speeds for all incoming and outgoing lanes.
All intersections share a single policy (parameter sharing), so the policy must generalize across topological positions.
We evaluate on three network topologies of increasing complexity (Figure~\ref{fig:sumo_networks}): a single 4-way intersection, a symmetric 4\(\times\)4 grid, and an asymmetric 4\(\times\)4 arterial network.

All three SUMO-RL scenarios share the pressure reward \(r_t = \sum_{l \in \mathcal{L}_{\text{out}}} n_l - \sum_{l \in \mathcal{L}_{\text{in}}} n_l\), where \(n_l\) is the number of vehicles currently on lane~\(l\) (fewer vehicles queued on incoming lanes relative to outgoing lanes indicates better throughput), and differ only in cost, label, and injected vehicles.
Evaluation cost limits are: Bus Priority \(c \leq 7\), Convoy Priority \(c \leq 6\), Premium Priority \(c \leq 15\) (see Table~\ref{tab:evaluation_cost_limits}).
We denote the set of lanes served by the current green phase as \(\mathcal{S}\), and \(q_j\) the queue fraction on lane~\(j\).
Injection parameters are listed in Table~\ref{tab:sumo_params}.

\subsubsection{Bus Priority --- Label and Cost}

\paragraph{Observation extension.}
Three features per lane (\(3 \times n_{\text{lanes}}\)): binary bus presence \(b_i \in \{0,1\}\), bus count (normalized), and bus waiting time normalized by a threshold \(T_c\).

\paragraph{Cost.}
Binary:
\begin{equation}
    c_t = \begin{cases}
        1 & \exists\, i : b_i > 0 \;\wedge\; \hat{w}_i^{\text{bus}} \geq 1, \\
        0 & \text{otherwise},
    \end{cases}
\end{equation}
where \(\hat{w}_i^{\text{bus}} = w_i^{\text{bus}} / T_c\) is the normalized bus wait.
Cost is phase-independent: it remains active while any bus is overdue regardless of whether its lane is currently served.

\paragraph{Label.}
\begin{equation}
    y(s) = \begin{cases}
        \mathrm{cost} & \exists\, i :\;
            b_i > 0 \;\wedge\;
            \hat{w}_i^{\text{bus}} \geq T_w / T_c, \\
        \mathrm{reward} & \text{otherwise}.
    \end{cases}
\end{equation}

The label fires when any bus has waited beyond the warning threshold, regardless of which lanes are currently served.

\subsubsection{Convoy Must-Not-Split --- Label and Cost}

\paragraph{Observation extension.}
Three features per lane (\(3 \times n_{\text{lanes}}\)): binary convoy presence \(v_i\), fraction of convoy vehicles observed at this intersection (convoy seen fraction \(\hat{n}_i\)), and convoy progress \(p_i = \text{(vehicles past stop line)} / \text{(total platoon)}\).

\paragraph{Cost.}
Binary active-split detection:
\begin{equation}
    c_t = \begin{cases}
        1 & \exists\, i \notin \mathcal{S} :\; v_i > 0 \;\wedge\; 0 < p_i < 1, \\
        0 & \text{otherwise}.
    \end{cases}
\end{equation}

\paragraph{Label.}
Presence-based, anticipatory:
\begin{equation}
    y(s) = \begin{cases}
        \mathrm{cost} & \exists\, i :\; v_i > 0 \;\wedge\; \hat{n}_i > 0, \\
        \mathrm{reward} & \text{otherwise}.
    \end{cases}
\end{equation}
The label fires as soon as any convoy vehicle is visible at the intersection (\(\hat{n}_i > 0\)), regardless of phase or progress.
This is deliberately anticipatory: the label activates \emph{before} cost (which requires \(0 < p_i < 1\)), providing the cost-sensitive update signal at the moment the phase-hold decision matters most.

The conflict in this scenario is rare and anticipatory. The label fires when a convoy becomes visible, before any cost has been incurred, and the window in which a violation can still be prevented spans only a few steps. A single multiplier shared across all states must therefore be either large enough to influence these rare states, which suppresses throughput across the ordinary traffic filling the rest of the episode, or small enough to preserve throughput, which leaves the cost gradients diluted by the majority.

\subsubsection{Premium Priority --- Label and Cost}

\paragraph{Observation extension.}
Two features per lane (\(2 \times n_{\text{lanes}}\)): binary premium presence \(\rho_i\) and premium wait \(\hat{w}_i^{\text{prem}}\) normalized by \(T_c\).
Premium wait is the \emph{local} delay accumulated at this intersection, computed as route delay minus the delay snapshot at entry.

\paragraph{Cost.}
Binary:
\begin{equation}
    c_t = \begin{cases}
        1 & \exists\, i : \rho_i > 0 \;\wedge\; \hat{w}_i^{\text{prem}} \geq 1, \\
        0 & \text{otherwise}.
    \end{cases}
\end{equation}

\paragraph{Label.}
Two-path decision frontier:
\begin{equation}
    y(s) = \begin{cases}
        \mathrm{cost} & \bigl(\exists\, i :\;
            \rho_i > 0 \;\wedge\;
            \hat{w}_i^{\text{prem}} \geq T_w / T_c\bigr)
            \;\vee\;
            \bigl(\exists\, i \notin \mathcal{S} :\;
            \rho_i > 0 \;\wedge\;
            \text{can\_switch}(s)\bigr), \\
        \mathrm{reward} & \text{otherwise}.
    \end{cases}
\end{equation}
The first condition fires when any premium vehicle's delay approaches the cost threshold (urgency).
The second fires when a premium vehicle is on an unserved lane and the controller is past its minimum green time (actionable frontier).

\begin{table}[hbt!]
\centering
\caption{\textbf{SUMO scenario injection and threshold parameters.} Injection intervals are tuned per network so that each intersection sees \({\sim}\)10--30\% label rate under a random policy.}
\label{tab:sumo_params}
\vspace{0.3em}
\resizebox{\textwidth}{!}{%
\begin{tabular}{llccc}
\toprule
\textbf{Parameter} & \textbf{Description} & \textbf{Bus} & \textbf{Convoy} & \textbf{Premium} \\
\midrule
\multirow{3}{*}{Injection interval (s)} & Single & 90 & 300 & 60 \\
 & Grid \(4\times4\) & 15 & 80 & 8 \\
 & Arterial \(4\times4\) & 20 & 80 & 20 \\
Vehicles per injection & Count per event & 1 & 10--12 & 1 \\
Vehicle length (m) & Injected vehicle size & 12 & 5 & 5 \\
Vehicle speed (m/s) & Max speed & 11.1 & 13.9 & 13.9 \\
Headway (s) & Spacing (convoy only) & --- & 2.0 & --- \\
\(T_c\) (cost threshold) & Normalization for wait & 15\,s & --- & 8\,s \\
\(T_w\) (warn threshold) & Label urgency gate & 6\,s & --- & 3\,s \\
Clean episode prob. & Prob.\ of no injection & \multicolumn{3}{c}{0.5 (all scenarios)} \\
\bottomrule
\end{tabular}%
}
\end{table}

\subsection{CityLearn Scenario Details}
\label{app:citylearn_scenarios}

CityLearn~\citep{citylearn} is a building energy management benchmark where a central agent controls battery storage for a district of three buildings over a one-month episode (720 hourly steps).
At each step the agent observes electricity prices, solar generation, building loads, and battery state-of-charge (SOC), and decides how much to charge or discharge each battery (continuous action space).
The reward is negative electricity cost; the constraint enforces a minimum reserve level whose semantics differ per scenario, creating a temporal conflict: the agent must trade off short-term cost savings against maintaining sufficient reserves for upcoming high-demand or low-generation periods.

All three CityLearn scenarios share the same reward \(r_t = -\frac{1}{N}\sum_i \text{consumption}_i(t) \times \text{price}_i(t)\) and differ only in cost, label, and observation augmentation.
The evaluation cost limit is \(c \leq 5\) for all CityLearn scenarios (see Table~\ref{tab:evaluation_cost_limits}).
Each scenario appends one scalar feature to the observation.
All normalized countdown features use \(T_{\text{ep}}=720\), the episode length.
Event detection and thresholds are computed once from the exogenous schedules at environment initialization.
Parameters are listed in Table~\ref{tab:citylearn_params}.
We denote mean battery SOC as \(\bar{s}\), district import as \(I_t = \max(\sum_i \text{NEC}_i(t),\, 0)\), and per-timestep max electricity price as \(\pi_t\).

\subsubsection{Arbitrage vs.\ Buffer --- Label and Cost}

\paragraph{Event detection.}
Peak-price periods: contiguous runs where \(\pi_t\) exceeds the \(Q(0.92)\) percentile.
A high-price sub-threshold \(\pi_{\text{high}}\) is the \(Q(0.75)\) of \emph{peak-hour} prices (not global).

\paragraph{Obs extension.}
Normalized steps until next peak (\(\text{steps\_until} / T_{\text{ep}}\)).

\paragraph{Cost.}
\begin{equation}
    c_t = \min\!\Big(1,\; \frac{\max(s_{\text{safe}} - \bar{s},\, 0)}{s_{\text{safe}}}\Big), \qquad s_{\text{safe}} = 0.30.
\end{equation}

\paragraph{Label.}
\begin{equation}
    y(s) = \begin{cases}
        \mathrm{cost} & t \in \mathcal{P}_{\text{peak}} \;\wedge\;
            \bigl(\pi_t > \pi_{\text{high}} \;\vee\; \bar{s} < s_{\text{emerg}}\bigr), \\
        \mathrm{reward} & \text{otherwise},
    \end{cases}
    \qquad s_{\text{emerg}} = 0.20 .
\end{equation}

\subsubsection{Demand Response --- Label and Cost}

\paragraph{Event detection.}
Deterministic grid events at fixed timesteps (hours 168, 360, 528), each lasting 12~hours.

\paragraph{Obs extension.}
Normalized steps until next event.

\paragraph{Cost.}
\begin{equation}
    c_t = \begin{cases}
        \min\!\Big(1,\; \dfrac{\max(s_{\text{tgt}} - \bar{s},\, 0)}{s_{\text{tgt}}}\Big) & t \in \mathcal{P}_{\text{event}}, \\[4pt]
        0 & \text{otherwise},
    \end{cases}
    \qquad s_{\text{tgt}} = 0.80.
\end{equation}

\paragraph{Label.}
\begin{equation}
    y(s) = \begin{cases}
        \mathrm{cost} & t \in \mathcal{P}_{\text{event}}, \\
        \mathrm{reward} & \text{otherwise}.
    \end{cases}
\end{equation}
The label is purely event-window membership: it fires throughout each 12-hour grid event, giving cost-sensitive updates control during the entire reserve-critical period.

\subsubsection{Solar Ramp Reserve --- Label and Cost}

\paragraph{Event detection.}
Sunset transitions detected from the solar schedule: hour \(t+1\) where \(\text{solar}(t) > 0\) and \(\text{solar}(t+1) = 0\).
Pre-sunset prep: 2~hours before each sunset.
Post-sunset buffer: 4~hours after each sunset.

\paragraph{Obs extension.}
Normalized steps until next buffer event.

\paragraph{Cost.}
\begin{equation}
    c_t = \begin{cases}
        \min\!\Big(1,\; \dfrac{\max(s_{\text{res}} - \bar{s},\, 0)}{s_{\text{res}}}\Big) & t \in \mathcal{P}_{\text{buffer}}, \\[4pt]
        0 & \text{otherwise},
    \end{cases}
    \qquad s_{\text{res}} = 0.50.
\end{equation}

\paragraph{Label.}
\begin{equation}
    y(s) = \begin{cases}
        \mathrm{cost} & t \in \mathcal{P}_{\text{prep}} \;\vee\; t \in \mathcal{P}_{\text{buffer}}, \\
        \mathrm{reward} & \text{otherwise}.
    \end{cases}
\end{equation}
The label fires during both the 2-hour pre-sunset preparation window and the 4-hour post-sunset buffer, regardless of current electricity prices or battery state-of-charge (SOC).
This gives cost-sensitive updates control over dispatch decisions throughout the entire sunset transition.

\begin{table}[hbt!]
\centering
\caption{\textbf{CityLearn scenario parameters.} All scenarios use the CityLearn Challenge 2023 Phase~2 dataset (3~buildings, 720~hourly steps per episode).}
\label{tab:citylearn_params}
\vspace{0.3em}
\resizebox{\textwidth}{!}{%
\begin{tabular}{lccc}
\toprule
\textbf{Parameter} & \textbf{Arb.\ vs.\ Buffer} & \textbf{Demand Response} & \textbf{Solar Ramp} \\
\midrule
SOC safety / target / reserve & 0.30 & 0.80 & 0.50 \\
Emergency SOC & 0.20 & 0.40 & 0.25 \\
Event quantile & \(Q_{92}\) (price) & fixed hours & solar schedule \\
High price quantile \(\pi_{\text{high}}\) & \(Q_{75}\) (peak-local) & --- & --- \\
Prep / buffer window (h) & --- & --- / 12 & 2 / 4 \\
\bottomrule
\end{tabular}%
}
\end{table}

\subsection{Atari RAM Dataset Details}
\label{app:atari_dataset}

\paragraph{Environment.}
We use the Arcade Learning Environment~\citep{bellemare2013arcade} with RAM observations, which are 128-dimensional byte vectors cast to float32.
All games use the \texttt{NoFrameskip-v4} variant with sticky actions disabled, matching the standard Atari evaluation protocol.
The eleven games evaluated are Qbert, Seaquest, Pong, SpaceInvaders, MsPacman, Frostbite, Alien, Amidar, BankHeist, KungFuMaster, and Gravitar.

\paragraph{Expert policies.}
Expert demonstrations were collected by rolling out pre-trained pixel-based agents while simultaneously recording the corresponding 128-byte RAM vectors.
All expert models were downloaded from the Hugging Face Hub:


\begin{itemize}[leftmargin=1.5em, itemsep=2pt]
  \item \textbf{Qbert, Seaquest, Pong, SpaceInvaders} --- PPO (CnnPolicy) from
  \path{ThomasSimonini/ppo-{Game}NoFrameskip-v4},\footnote{\url{https://huggingface.co/ThomasSimonini}}
  trained with Stable-Baselines3~\citep{stable-baselines3}.

  \item \textbf{MsPacman, Frostbite, Alien, Amidar, BankHeist, KungFuMaster, Gravitar} --- PPO with IMPALA-Net from
  \path{cleanrl/{Game}-v5-ppo_atari_envpool_async_jax_scan_impalanet_machado-seed1},\footnote{\url{https://huggingface.co/cleanrl}}
  trained with CleanRL~\citep{huang2022cleanrl}.
\end{itemize}

All models are publicly available on the Hugging Face Hub without access restrictions.
Stable-Baselines3 is released under the MIT License; CleanRL is released under the MIT License.

\paragraph{Non-expert data.}
Non-expert episodes are collected by sampling actions uniformly at random from the environment's action space in the same RAM-observation setting.

\paragraph{Mixed datasets.}
For each game and dataset size \(N\), we construct a mixed dataset with a 1:1 ratio of expert and non-expert transitions.
Each episode carries a guidance label indicating its provenance, \(y(s)=\mathrm{expert}\) for
expert episodes and \(y(s)=\mathrm{improve}\) for non-expert episodes.
Split-RL uses these labels to route gradient updates; baseline methods ignore them.
Six dataset sizes are evaluated: 10K, 20K, 50K, 100K, 200K, and Full, where Full uses the complete available mixed dataset.
We report normalized values in the main paper calculated as:
 \[
R_{\mathrm{norm}} = \frac{R_\pi - R_{\mathrm{non\text{-}expert}}}{R_{\mathrm{expert}} - R_{\mathrm{non\text{-}expert}}}.
\]

\section{Training Hyperparameters}
\label{app:hyperparameters}
All constrained RL experiments (MiniGrid, SUMO-RL, and CityLearn) are trained for 1M environment steps.
Hyperparameters were selected via Bayesian search on a single seed and then held fixed across all evaluation seeds and scenarios within each domain.
Split-RL and GBT baselines use oblivious trees and the cosine similarity split-score criterion throughout. These were selected as part of the hyperparameter search and we did not run dedicated experiments comparing which tree structure or split criterion is superior.
Neural network baselines use two-layer MLPs (64--64) with tanh activations unless otherwise noted.
We use 15 seeds per configuration for all constrained RL domains (MiniGrid, SUMO-RL, and CityLearn), including ablations. During training, episodic reward and cost are averaged over the last 100 episodes. Tables use the final averages from each seed and report their mean and sample standard deviation. The 95\% confidence intervals use percentile bootstrapping across seeds with 2{,}000 resamples. Cost limits are listed in Table~\ref{tab:evaluation_cost_limits}. Each limit is fixed before training and used as CPO's cost constraint, CUP's projection target, IPO's barrier, and the Lagrangian penalty threshold. The offline-RL evaluation protocol is described below alongside Table~\ref{tab:hp_atari}.

\paragraph{Constrained-RL implementation.}
All constrained-RL methods use the same environment wrappers, rollout collection, evaluation code, and seed protocol. Neural CPO, CUP, IPO, and PPO-Lagrangian are implemented in a shared Stable-Baselines3~\citep{stable-baselines3} codebase, following the corresponding OmniSafe~\citep{omnisafe} implementations. Split-RL and PPO-Lagrangian (GBT) use the same training interface as GBRL tree policies.

\subsection{MiniGrid}

Table~\ref{tab:hp_minigrid} lists the hyperparameters for all MiniGrid methods.
All methods share \(\gamma = 0.99\) and \(\lambda_{\text{GAE}} = 0.95\).

\begin{table}[hbt!]
\centering
\caption{\textbf{MiniGrid hyperparameters.} All methods share \(\gamma{=}0.99\) and \(\lambda_{\text{GAE}}{=}0.95\). NN baselines normalize observations and rewards; GBT methods use no normalization. ``---'' = not applicable.}
\label{tab:hp_minigrid}
\vspace{0.3em}
\resizebox{\textwidth}{!}{%
\begin{tabular}{lcccccc}
\toprule
\textbf{Parameter} & \textbf{CPO} & \textbf{CUP} & \textbf{IPO} & \textbf{PPOLag} & \textbf{PPOLag (GBT)} & \textbf{Split-RL} \\
\midrule
Num.\ envs & 1 & 16 & 8 & 16 & 16 & 16 \\
\(n_{\text{steps}}\) & 4096 & 1024 & 256 & 256 & 256 & 256 \\
Batch size & 256 & 128 & 256 & 128 & 512 & 512 \\
Epochs & 30 & 10 & 10 & 20 & 20 & 20 \\
Learning rate & \(4.3{\times}10^{-4}\) & \(7.4{\times}10^{-5}\) & \(7.6{\times}10^{-4}\) & \(2.1{\times}10^{-4}\) & --- & --- \\
Policy LR & --- & --- & --- & --- & 0.17 & 0.17 \\
Value LR & --- & --- & --- & --- & 0.01 & 0.01 \\
Entropy coef. & --- & --- & 0.02 & --- & 0.0 & 0.0 \\
Clip range & --- & --- & --- & 0.25 & 0.2 & 0.2 \\
Tree depth & --- & --- & --- & --- & 4 & 4 \\
\(\lambda_0\) & --- & --- & --- & --- & --- & 1.0 \\
\bottomrule
\end{tabular}%
}
\end{table}

\subsection{SUMO-RL}

Table~\ref{tab:hp_sumo} lists the hyperparameters for the SUMO-RL traffic signal control experiments.
Training uses 1M environment steps per network--scenario pair with a conflict probability of 0.5 (special vehicles are injected in 50\% of episodes; the remaining episodes are clean).

\begin{table}[htbp!]
\centering
\caption{\textbf{SUMO-RL hyperparameters.} All methods use reward normalization. GBRL methods use oblivious trees.}
\label{tab:hp_sumo}
\vspace{0.3em}
\resizebox{\textwidth}{!}{%
\begin{tabular}{lcccccc}
\toprule
\textbf{Parameter} & \textbf{CPO} & \textbf{CUP} & \textbf{IPO} & \textbf{PPOLag} & \textbf{PPOLag (GBT)} & \textbf{Split-RL} \\
\midrule
\(n_{\text{steps}}\) & 256 & 256 & 512 & 256 & 512 & 512 \\
Batch size & 128 & 256 & 256 & 256 & 128 & 128 \\
Epochs & 10 & 20 & 10 & 20 & 8 & 8 \\
\(\gamma\) & 0.95 & 0.95 & 0.90 & 0.95 & 0.90 & 0.90 \\
\(\lambda_{\text{GAE}}\) & 0.98 & 0.95 & 0.90 & 0.98 & 0.95 & 0.95 \\
Clip range & 0.2 & 0.2 & 0.3 & 0.1 & 0.2 & 0.2 \\
Entropy coef. & 0.01 & \(4.2{\times}10^{-3}\) & \(2.2{\times}10^{-5}\) & \(1.1{\times}10^{-5}\) & \(3.1{\times}10^{-5}\) & \(3.1{\times}10^{-5}\) \\
Learning rate & \(2.6{\times}10^{-3}\) & \(7.7{\times}10^{-5}\) & \(1.1{\times}10^{-4}\) & \(2.3{\times}10^{-4}\) & --- & --- \\
Policy LR & --- & --- & --- & --- & 0.060 & 0.060 \\
Value LR & --- & --- & --- & --- & 0.036 & 0.036 \\
Cost LR & --- & --- & --- & --- & 0.008 & 0.008 \\
Max grad norm & 0.5 & 1.19 & 1.24 & 0.88 & 1.0 & 1.0 \\
Target KL & 0.01 & --- & --- & --- & 0.02 & 0.02 \\
Tree depth & --- & --- & --- & --- & 5 & 5 \\
\(\lambda_0\) & --- & --- & --- & --- & --- & 1.0 \\
\bottomrule
\end{tabular}%
}
\end{table}

\subsection{CityLearn}

Table~\ref{tab:hp_citylearn} lists the hyperparameters for the CityLearn building energy management experiments.
Training uses 1M environment steps per scenario.
Actions are continuous (battery charge/discharge); Split-RL and GBRL use a fixed log-std initialization.

\begin{table}[htbp!]
\centering
\caption{\textbf{CityLearn hyperparameters.} All GBRL methods use oblivious depth-4 trees. NN baselines use reward normalization; GBT methods use no normalization.}
\label{tab:hp_citylearn}
\vspace{0.3em}
\resizebox{\textwidth}{!}{%
\begin{tabular}{lcccccc}
\toprule
\textbf{Parameter} & \textbf{CPO} & \textbf{CUP} & \textbf{IPO} & \textbf{PPOLag} & \textbf{PPOLag (GBT)} & \textbf{Split-RL} \\
\midrule
\(n_{\text{steps}}\) & 512 & 256 & 512 & 256 & 256 & 256 \\
Batch size & 128 & 256 & 64 & 64 & 64 & 128 \\
Epochs & 10 & 20 & 20 & 4 & 10 & 20 \\
\(\gamma\) & 0.90 & 0.95 & 0.90 & 0.90 & 0.95 & 0.95 \\
\(\lambda_{\text{GAE}}\) & 0.90 & 0.95 & 0.95 & 0.95 & 0.95 & 0.95 \\
Clip range & 0.2 & 0.2 & 0.2 & 0.3 & 0.2 & 0.2 \\
Entropy coef. & 0.01 & \(4.0{\times}10^{-3}\) & \(4.2{\times}10^{-3}\) & \(2.5{\times}10^{-5}\) & \(5.9{\times}10^{-4}\) & \(5.9{\times}10^{-4}\) \\
Learning rate & \(5.3{\times}10^{-5}\) & \(4.3{\times}10^{-5}\) & \(2.8{\times}10^{-5}\) & \(1.9{\times}10^{-3}\) & --- & --- \\
Policy LR & --- & --- & --- & --- & \(4.7{\times}10^{-5}\) & \(4.7{\times}10^{-5}\) \\
Value LR & --- & --- & --- & --- & \(4.3{\times}10^{-4}\) & \(4.3{\times}10^{-4}\) \\
Cost LR & --- & --- & --- & --- & --- & \(2.0{\times}10^{-3}\) \\
Target KL & 0.02 & 0.02 & --- & --- & 0.05 & 0.05 \\
Fixed std & --- & --- & --- & --- & yes & yes \\
Log-std init & --- & --- & --- & --- & \(-3\) & \(-3\) \\
Tree depth & --- & --- & --- & --- & 4 & 4 \\
\(\lambda_0\) & --- & --- & --- & --- & --- & 1.0 \\
\bottomrule
\end{tabular}%
}
\end{table}

\subsection{Atari RAM (Offline)}
Table~\ref{tab:hp_atari} lists the hyperparameters for the offline Atari RAM experiments.
Training budgets differ by method: CQL and BCQ train for 500K gradient steps, Decision Transformer for 250K steps, and AWR-based methods (NN and GBRL) including Split-RL train for 50K steps.
All methods use the same fixed mixed dataset per game (expert + random transitions at each reported dataset size) and each method is trained once per game and dataset size. For evaluation, we use deterministic action selection on five environment seeds with 10 episodes each, for a total of 50 evaluation episodes.

\paragraph{Offline-RL implementation.}
CQL, BCQ, and Decision Transformer use d3rlpy implementations~\citep{d3rlpy}. AWR and AWR+BC are implemented in our codebase for both neural and GBT policies.
To keep the BC and AWR gradients on comparable scales, Split-RL divides each objective’s gradients by their standard deviation before aggregation.
\begin{table}[htbp!]
\centering
\caption{\textbf{Atari RAM offline hyperparameters.} The GBT baselines use the shared hyperparameter values selected for Split-RL. The main difference is whether guidance labels (Split-RL) or BC regularization (AWR+BC) is active.}
\label{tab:hp_atari}
\vspace{0.3em}
\resizebox{\textwidth}{!}{%
\begin{tabular}{lcccccccc}
\toprule
\textbf{Parameter} & \textbf{CQL} & \textbf{DT} & \textbf{AWR (NN)} & \textbf{AWR+BC (NN)} & \textbf{BCQ} & \textbf{AWR (GBRL)} & \textbf{AWR+BC (GBRL)} & \textbf{Split-RL} \\
\midrule
Training steps & 500K & 250K & 50K & 50K & 500K & 50K & 50K & 50K \\
Batch size & 128 & 128 & 256 & 256 & 128 & 1024 & 1024 & 1024 \\
\(\gamma\) & 0.99 & --- & 1.0 & 1.0 & 0.99 & 1.0 & 1.0 & 1.0 \\
\(\beta\) (AWR temp) & --- & --- & 1.0 & 1.0 & --- & 1.0 & 1.0 & 1.0 \\
Learning rate & \(2.6{\times}10^{-4}\) & \(2.6{\times}10^{-4}\) & \(1.4{\times}10^{-4}\) & \(1.4{\times}10^{-4}\) & \(2.1{\times}10^{-4}\) & --- & --- & --- \\
Policy LR & --- & --- & --- & --- & --- & 0.025 & 0.025 & 0.025 \\
Value LR & --- & --- & --- & --- & --- & 0.005 & 0.005 & 0.005 \\
Entropy coef. & --- & --- & \(1.5{\times}10^{-4}\) & \(1.5{\times}10^{-4}\) & --- & 0.01 & 0.01 & 0.01 \\
CQL \(\alpha\) & 4.14 & --- & --- & --- & --- & --- & --- & --- \\
BC coef. & --- & --- & --- & 1.0 & --- & --- & 1.0 & --- \\
Context length & --- & 30 & --- & --- & --- & --- & --- & --- \\
Tree depth & --- & --- & --- & --- & --- & 5 & 5 & 5 \\
Eval.\ seeds & 5 & 5 & 5 & 5 & 5 & 5 & 5 & 5 \\
\(\lambda_0\) & --- & --- & --- & --- & --- & --- & --- & 1.0 \\
\bottomrule
\end{tabular}%
}
\end{table}

\section{Detailed Results}
\label{app:detailed_results}
This appendix reports the raw values underlying the main summaries. For constrained RL, bold entries mark the feasible method with the highest reward in each scenario. For Atari RAM, we report raw returns per game and dataset size; the main text reports expert-normalized averages computed from these values.

Feasibility is determined using the cost limits in Table~\ref{tab:evaluation_cost_limits}. The cost limit is part of the problem specification rather than a post-hoc threshold: a single limit per scenario is set before training, and every constrained baseline receives it and optimizes against it, entering CPO's cost constraint, CUP's projection, IPO's barrier, and the Lagrangian penalty. Each method therefore targets exactly the limit it is scored against. Most mean-cost violations are large rather than marginal, while exceptions include PPO-Lagrangian on SUMO-RL Single/Premium (15.06 against 15) and CUP on CityLearn Arbitrage vs. Buffer (5.06 against 5).

\begin{table}[hbt!]
\centering
\caption{\textbf{Evaluation cost limits.} A method is \emph{feasible} if its mean episodic cost does not exceed the limit for its scenario. Bold entries in Tables~\ref{tab:full_results_minigrid}--\ref{tab:full_results_citylearn} mark the feasible method with highest reward.}
\label{tab:evaluation_cost_limits}
\vspace{0.3em}
\small
\begin{tabular}{llc}
\toprule
\textbf{Domain} & \textbf{Scenario} & \textbf{Cost limit} \\
\midrule
MiniGrid & DynamicCrossing & 0.1 \\
MiniGrid & FragileCrossing & 0.1 \\
MiniGrid & Corner & 0.1 \\
\midrule
SUMO-RL & Bus Priority & 7.0 \\
SUMO-RL & Convoy Priority & 6.0 \\
SUMO-RL & Premium Priority & 15.0 \\
\midrule
CityLearn & Arb.\ vs Buffer & 5.0 \\
CityLearn & Demand Response & 5.0 \\
CityLearn & Solar Ramp Reserve & 5.0 \\
\bottomrule
\end{tabular}
\end{table}

\begin{table*}[thbp!]
\centering
\caption{\textbf{MiniGrid full results.}
Mean \(\pm\) std, with 95\% bootstrap CI, over seeds. Episodic reward (\(\uparrow\)) and cost
(\(\downarrow\)). \textbf{Bold} marks the feasible operating point with highest reward;
\underline{underline} marks the feasible operating point with the second-highest reward, using
the limits in Table~\ref{tab:evaluation_cost_limits}.}
\label{tab:full_results_minigrid}
\vspace{0.3em}
\footnotesize
\setlength{\tabcolsep}{2.2pt}
\renewcommand{\arraystretch}{1.6}
\begin{adjustbox}{max width=\textwidth}
\begin{tabular}{@{}llrrrrrr@{}}
\toprule
 & & \multicolumn{4}{c}{\textbf{Neural Network}} & \multicolumn{2}{c}{\textbf{GBT}} \\
\cmidrule(lr){3-6} \cmidrule(lr){7-8}
\textbf{Scenario} & \textbf{Metric}
& \textbf{CPO} & \textbf{CUP} & \textbf{IPO} & \textbf{PPOLag}
& \textbf{PPOLag} & \textbf{Split-RL} \\
\midrule
\multirow{2}{*}{DynamicCrossing}
& Rew. & \shortstack{\(0.85 \pm 0.52\)\\\([0.59,1.11]\)} & \shortstack{\(0.12 \pm 0.13\)\\\([0.05,0.18]\)} & \shortstack{\(0.91 \pm 0.69\)\\\([0.57,1.23]\)} & \shortstack{\(0.53 \pm 0.51\)\\\([0.29,0.79]\)} & \shortstack{\(\underline{1.48 \pm 0.13}\)\\\(\underline{[1.40,1.53]}\)} & \shortstack{\(\mathbf{1.53 \pm 0.01}\)\\\(\mathbf{[1.52,1.53]}\)} \\
& Cost & \shortstack{\(0.34 \pm 0.84\)\\\([0.07,0.80]\)} & \shortstack{\(5.98 \pm 22.74\)\\\([0.05,17.76]\)} & \shortstack{\(2.72 \pm 7.51\)\\\([0.40,6.85]\)} & \shortstack{\(23.38 \pm 61.86\)\\\([2.55,55.79]\)} & \shortstack{\(\underline{0.01 \pm 0.01}\)\\\(\underline{[0.00,0.01]}\)} & \shortstack{\(\mathbf{0.00 \pm 0.00}\)\\\(\mathbf{[0.00,0.00]}\)} \\

\multirow{2}{*}{FragileCrossing}
& Rew. & \shortstack{\(0.48 \pm 0.30\)\\\([0.34,0.62]\)} & \shortstack{\(0.13 \pm 0.20\)\\\([0.03,0.23]\)} & \shortstack{\(0.47 \pm 0.37\)\\\([0.29,0.65]\)} & \shortstack{\(0.36 \pm 0.37\)\\\([0.17,0.55]\)} & \shortstack{\(\underline{0.93 \pm 0.21}\)\\\(\underline{[0.82,1.03]}\)} & \shortstack{\(\mathbf{1.07 \pm 0.11}\)\\\(\mathbf{[1.01,1.10]}\)} \\
& Cost & \shortstack{\(0.34 \pm 0.61\)\\\([0.09,0.66]\)} & \shortstack{\(0.00 \pm 0.00\)\\\([0.00,0.00]\)} & \shortstack{\(0.24 \pm 0.21\)\\\([0.14,0.34]\)} & \shortstack{\(0.02 \pm 0.04\)\\\([0.01,0.04]\)} & \shortstack{\(\underline{0.04 \pm 0.04}\)\\\(\underline{[0.02,0.06]}\)} & \shortstack{\(\mathbf{0.01 \pm 0.01}\)\\\(\mathbf{[0.00,0.01]}\)} \\

\multirow{2}{*}{Corner}
& Rew. & \shortstack{\(0.93 \pm 0.11\)\\\([0.87,0.98]\)} & \shortstack{\(0.00 \pm 0.00\)\\\([0.00,0.00]\)} & \shortstack{\(0.51 \pm 0.48\)\\\([0.27,0.76]\)} & \shortstack{\(0.48 \pm 0.47\)\\\([0.24,0.72]\)} & \shortstack{\(\mathbf{0.97 \pm 0.09}\)\\\(\mathbf{[0.92,1.00]}\)} & \shortstack{\(\underline{0.94 \pm 0.07}\)\\\(\underline{[0.91,0.97]}\)} \\
& Cost & \shortstack{\(0.22 \pm 0.50\)\\\([0.01,0.49]\)} & \shortstack{\(0.02 \pm 0.07\)\\\([0.00,0.06]\)} & \shortstack{\(0.69 \pm 0.84\)\\\([0.36,1.13]\)} & \shortstack{\(0.23 \pm 0.68\)\\\([0.03,0.59]\)} & \shortstack{\(\mathbf{0.03 \pm 0.12}\)\\\(\mathbf{[0.00,0.09]}\)} & \shortstack{\(\underline{0.00 \pm 0.01}\)\\\(\underline{[0.00,0.01]}\)} \\
\bottomrule
\end{tabular}
\end{adjustbox}
\end{table*}

\begin{table*}[hbt!]
\centering
\caption{\textbf{SUMO-RL full results.}
Mean \(\pm\) std, with 95\% bootstrap CI, over seeds. Episodic reward (\(\uparrow\)) and cost
(\(\downarrow\)), evaluated with conflict probability 0.5 (i.e.\ special vehicles are injected in
50\% of episodes). Reward is in units of \(10^{3}\) throughout; cost is unscaled. \textbf{Bold}
marks the feasible operating point with highest reward; \underline{underline} marks the feasible
operating point with the second-highest reward, ties included, using the limits in
Table~\ref{tab:evaluation_cost_limits}.}
\label{tab:full_results_sumo}
\vspace{0.3em}
\footnotesize
\setlength{\tabcolsep}{3.0pt}
\renewcommand{\arraystretch}{1.5}
\begin{adjustbox}{max width=\textwidth}
\begin{tabular}{@{}llrrrrrr@{}}
\toprule
 & & \multicolumn{4}{c}{\textbf{Neural Network}} & \multicolumn{2}{c}{\textbf{GBT}} \\
\cmidrule(lr){3-6} \cmidrule(lr){7-8}
\textbf{Scenario} & \textbf{Metric}
& \textbf{CPO} & \textbf{CUP} & \textbf{IPO} & \textbf{PPOLag}
& \textbf{PPOLag} & \textbf{Split-RL} \\
\midrule
\multirow{2}{*}{Arterial / Bus}
& Rew. & \shortstack{\(-1.407 \pm 0.324\)\\\([-1.570,-1.267]\)} & \shortstack{\(-1.332 \pm 0.228\)\\\([-1.447,-1.229]\)} & \shortstack{\(\underline{-1.267 \pm 0.231}\)\\\(\underline{[-1.389,-1.164]}\)} & \shortstack{\(-1.299 \pm 0.256\)\\\([-1.430,-1.185]\)} & \shortstack{\(-1.210 \pm 0.137\)\\\([-1.278,-1.147]\)} & \shortstack{\(\mathbf{-1.189 \pm 0.181}\)\\\(\mathbf{[-1.279,-1.107]}\)} \\
& Cost & \shortstack{\(3.44 \pm 5.85\)\\\([1.33,6.73]\)} & \shortstack{\(3.44 \pm 3.15\)\\\([2.12,5.14]\)} & \shortstack{\(\underline{1.85 \pm 1.04}\)\\\(\underline{[1.38,2.37]}\)} & \shortstack{\(6.60 \pm 5.01\)\\\([4.21,9.15]\)} & \shortstack{\(20.96 \pm 10.01\)\\\([16.18,26.02]\)} & \shortstack{\(\mathbf{3.92 \pm 2.46}\)\\\(\mathbf{[2.73,5.11]}\)} \\

\multirow{2}{*}{Arterial / Convoy}
& Rew. & \shortstack{\(-1.429 \pm 0.273\)\\\([-1.571,-1.303]\)} & \shortstack{\(-1.470 \pm 0.220\)\\\([-1.589,-1.369]\)} & \shortstack{\(\underline{-1.385 \pm 0.258}\)\\\(\underline{[-1.511,-1.271]}\)} & \shortstack{\(-1.232 \pm 0.250\)\\\([-1.362,-1.118]\)} & \shortstack{\(-1.229 \pm 0.149\)\\\([-1.303,-1.159]\)} & \shortstack{\(\mathbf{-1.296 \pm 0.147}\)\\\(\mathbf{[-1.377,-1.230]}\)} \\
& Cost & \shortstack{\(2.79 \pm 1.73\)\\\([1.98,3.72]\)} & \shortstack{\(1.88 \pm 0.85\)\\\([1.47,2.29]\)} & \shortstack{\(\underline{2.42 \pm 2.65}\)\\\(\underline{[1.30,3.88]}\)} & \shortstack{\(12.35 \pm 6.26\)\\\([9.58,15.60]\)} & \shortstack{\(12.58 \pm 3.12\)\\\([11.17,14.21]\)} & \shortstack{\(\mathbf{2.16 \pm 0.86}\)\\\(\mathbf{[1.80,2.64]}\)} \\

\multirow{2}{*}{Arterial / Premium}
& Rew. & \shortstack{\(-1.378 \pm 0.315\)\\\([-1.530,-1.231]\)} & \shortstack{\(\underline{-1.366 \pm 0.242}\)\\\(\underline{[-1.495,-1.268]}\)} & \shortstack{\(-1.394 \pm 0.250\)\\\([-1.514,-1.275]\)} & \shortstack{\(-1.170 \pm 0.150\)\\\([-1.243,-1.099]\)} & \shortstack{\(-1.259 \pm 0.147\)\\\([-1.335,-1.192]\)} & \shortstack{\(\mathbf{-1.254 \pm 0.140}\)\\\(\mathbf{[-1.322,-1.185]}\)} \\
& Cost & \shortstack{\(7.18 \pm 3.23\)\\\([5.69,8.84]\)} & \shortstack{\(\underline{5.95 \pm 3.53}\)\\\(\underline{[4.38,7.81]}\)} & \shortstack{\(6.43 \pm 2.49\)\\\([5.24,7.61]\)} & \shortstack{\(16.22 \pm 7.27\)\\\([13.37,20.47]\)} & \shortstack{\(37.41 \pm 10.62\)\\\([32.30,42.31]\)} & \shortstack{\(\mathbf{13.91 \pm 6.96}\)\\\(\mathbf{[10.77,17.92]}\)} \\
\midrule

\multirow{2}{*}{Grid / Bus}
& Rew. & \shortstack{\(-0.216 \pm 0.028\)\\\([-0.230,-0.203]\)} & \shortstack{\(-0.156 \pm 0.025\)\\\([-0.169,-0.145]\)} & \shortstack{\(-0.163 \pm 0.024\)\\\([-0.176,-0.152]\)} & \shortstack{\(\mathbf{-0.134 \pm 0.025}\)\\\(\mathbf{[-0.147,-0.123]}\)} & \shortstack{\(-0.153 \pm 0.026\)\\\([-0.167,-0.142]\)} & \shortstack{\(\underline{-0.145 \pm 0.024}\)\\\(\underline{[-0.158,-0.134]}\)} \\
& Cost & \shortstack{\(0.54 \pm 0.43\)\\\([0.35,0.76]\)} & \shortstack{\(0.05 \pm 0.06\)\\\([0.02,0.08]\)} & \shortstack{\(0.02 \pm 0.02\)\\\([0.01,0.03]\)} & \shortstack{\(\mathbf{0.09 \pm 0.07}\)\\\(\mathbf{[0.06,0.13]}\)} & \shortstack{\(0.31 \pm 0.11\)\\\([0.26,0.36]\)} & \shortstack{\(\underline{0.05 \pm 0.03}\)\\\(\underline{[0.04,0.07]}\)} \\

\multirow{2}{*}{Grid / Convoy}
& Rew. & \shortstack{\(-0.267 \pm 0.024\)\\\([-0.279,-0.255]\)} & \shortstack{\(\underline{-0.181 \pm 0.024}\)\\\(\underline{[-0.193,-0.170]}\)} & \shortstack{\(-0.191 \pm 0.023\)\\\([-0.202,-0.180]\)} & \shortstack{\(-0.157 \pm 0.025\)\\\([-0.170,-0.146]\)} & \shortstack{\(-0.178 \pm 0.029\)\\\([-0.193,-0.165]\)} & \shortstack{\(\mathbf{-0.171 \pm 0.027}\)\\\(\mathbf{[-0.185,-0.158]}\)} \\
& Cost & \shortstack{\(2.34 \pm 0.69\)\\\([1.99,2.65]\)} & \shortstack{\(\underline{1.35 \pm 0.42}\)\\\(\underline{[1.15,1.56]}\)} & \shortstack{\(3.39 \pm 0.98\)\\\([2.93,3.89]\)} & \shortstack{\(7.94 \pm 2.37\)\\\([6.79,9.14]\)} & \shortstack{\(8.95 \pm 1.97\)\\\([7.97,9.87]\)} & \shortstack{\(\mathbf{2.41 \pm 0.50}\)\\\(\mathbf{[2.17,2.65]}\)} \\

\multirow{2}{*}{Grid / Premium}
& Rew. & \shortstack{\(-0.225 \pm 0.027\)\\\([-0.238,-0.212]\)} & \shortstack{\(-0.162 \pm 0.025\)\\\([-0.174,-0.150]\)} & \shortstack{\(-0.169 \pm 0.025\)\\\([-0.182,-0.158]\)} & \shortstack{\(\mathbf{-0.138 \pm 0.025}\)\\\(\mathbf{[-0.151,-0.127]}\)}& \shortstack{\(\underline{-0.153 \pm 0.022}\)\\\(\underline{[-0.164,-0.143]}\)} & \shortstack{\(\underline{-0.153 \pm 0.030}\)\\\(\underline{[-0.169,-0.140]}\)} \\
& Cost & \shortstack{\(5.09 \pm 2.28\)\\\([3.99,6.21]\)} & \shortstack{\(2.01 \pm 0.63\)\\\([1.68,2.32]\)} & \shortstack{\(2.80 \pm 0.80\)\\\([2.40,3.18]\)} & \shortstack{\(\mathbf{3.02 \pm 1.18}\)\\\(\mathbf{[2.44,3.59]}\)} & \shortstack{\(\underline{4.22 \pm 1.44}\)\\\(\underline{[3.51,4.86]}\)} & \shortstack{\(\underline{1.84 \pm 0.43}\)\\\(\underline{[1.62,2.04]}\)} \\
\midrule

\multirow{2}{*}{Single / Bus}
& Rew. & \shortstack{\(-8.767 \pm 0.187\)\\\([-8.857,-8.674]\)} & \shortstack{\(-8.839 \pm 0.626\)\\\([-9.183,-8.610]\)} & \shortstack{\(-8.746 \pm 0.182\)\\\([-8.841,-8.663]\)} & \shortstack{\(\mathbf{-8.421 \pm 0.051}\)\\\(\mathbf{[-8.445,-8.395]}\)} & \shortstack{\(-8.519 \pm 0.082\)\\\([-8.558,-8.478]\)} & \shortstack{\(\underline{-8.433 \pm 0.125}\)\\\(\underline{[-8.494,-8.377]}\)} \\
& Cost & \shortstack{\(2.31 \pm 1.15\)\\\([1.81,2.94]\)} & \shortstack{\(2.96 \pm 1.09\)\\\([2.37,3.47]\)} & \shortstack{\(3.76 \pm 0.82\)\\\([3.38,4.19]\)} & \shortstack{\(\mathbf{2.98 \pm 0.64}\)\\\(\mathbf{[2.67,3.27]}\)} & \shortstack{\(4.06 \pm 0.82\)\\\([3.65,4.45]\)} & \shortstack{\(\underline{0.87 \pm 0.14}\)\\\(\underline{[0.79,0.93]}\)} \\

\multirow{2}{*}{Single / Convoy}
& Rew. & \shortstack{\(-75.609 \pm 12.567\)\\\([-80.388,-68.190]\)} & \shortstack{\(-33.296 \pm 8.846\)\\\([-38.061,-29.716]\)} & \shortstack{\(-14.121 \pm 0.720\)\\\([-14.494,-13.774]\)} & \shortstack{\(-17.406 \pm 3.642\)\\\([-19.187,-15.673]\)} & \shortstack{\(-14.949 \pm 1.292\)\\\([-15.578,-14.315]\)} & \shortstack{\(\mathbf{-26.971 \pm 1.655}\)\\\(\mathbf{[-27.818,-26.196]}\)} \\
& Cost & \shortstack{\(19.81 \pm 9.91\)\\\([14.48,24.48]\)} & \shortstack{\(13.81 \pm 7.58\)\\\([10.35,17.54]\)} & \shortstack{\(25.95 \pm 4.56\)\\\([23.80,28.08]\)} & \shortstack{\(23.51 \pm 7.79\)\\\([19.67,27.26]\)} & \shortstack{\(78.83 \pm 13.27\)\\\([72.52,85.53]\)} & \shortstack{\(\mathbf{5.66 \pm 2.50}\)\\\(\mathbf{[4.46,6.96]}\)} \\

\multirow{2}{*}{Single / Premium}
& Rew. & \shortstack{\(-23.398 \pm 25.250\)\\\([-36.364,-11.951]\)} & \shortstack{\(-9.093 \pm 0.283\)\\\([-9.232,-8.954]\)} & \shortstack{\(\underline{-8.955 \pm 0.195}\)\\\(\underline{[-9.049,-8.863]}\)} & \shortstack{\(-8.601 \pm 0.160\)\\\([-8.691,-8.536]\)} & \shortstack{\(-8.541 \pm 0.072\)\\\([-8.576,-8.508]\)} & \shortstack{\(\mathbf{-8.679 \pm 0.183}\)\\\(\mathbf{[-8.775,-8.598]}\)} \\
& Cost & \shortstack{\(38.26 \pm 84.51\)\\\([6.37,87.90]\)} & \shortstack{\(8.63 \pm 3.81\)\\\([6.80,10.63]\)} & \shortstack{\(\underline{13.70 \pm 1.72}\)\\\(\underline{[12.90,14.58]}\)} & \shortstack{\(15.06 \pm 1.75\)\\\([14.22,15.92]\)} & \shortstack{\(24.84 \pm 2.14\)\\\([23.77,25.86]\)} & \shortstack{\(\mathbf{6.95 \pm 1.09}\)\\\(\mathbf{[6.41,7.47]}\)} \\
\bottomrule
\end{tabular}
\end{adjustbox}
\end{table*}

\begin{table*}[hbt!]
\centering
\caption{\textbf{CityLearn full results.}
Mean \(\pm\) std, with 95\% bootstrap CI, over seeds. Episodic reward (\(\uparrow\)) and cost
(\(\downarrow\)). \textbf{Bold} marks the feasible operating point with highest reward; no method
besides Split-RL is feasible in any CityLearn scenario, so no entry is underlined here (using the
limits in Table~\ref{tab:evaluation_cost_limits}).}
\label{tab:full_results_citylearn}
\vspace{0.3em}
\footnotesize
\setlength{\tabcolsep}{2.5pt}
\renewcommand{\arraystretch}{1.6}
\begin{adjustbox}{max width=\textwidth}
\begin{tabular}{@{}llrrrrrr@{}}
\toprule
 & & \multicolumn{4}{c}{\textbf{Neural Network}} & \multicolumn{2}{c}{\textbf{GBT}} \\
\cmidrule(lr){3-6} \cmidrule(lr){7-8}
\textbf{Scenario} & \textbf{Metric}
& \textbf{CPO} & \textbf{CUP} & \textbf{IPO} & \textbf{PPOLag}
& \textbf{PPOLag} & \textbf{Split-RL} \\
\midrule
\multirow{2}{*}{Arb.\ vs Buffer}
& Rew. & \shortstack{\(-7.96 \pm 0.41\)\\\([-8.16,-7.78]\)} & \shortstack{\(-8.48 \pm 0.14\)\\\([-8.54,-8.41]\)} & \shortstack{\(-8.62 \pm 0.05\)\\\([-8.64,-8.59]\)} & \shortstack{\(-8.13 \pm 0.28\)\\\([-8.25,-7.97]\)} & \shortstack{\(-8.81 \pm 0.10\)\\\([-8.86,-8.76]\)} & \shortstack{\(\mathbf{-8.90 \pm 0.13}\)\\\(\mathbf{[-8.97,-8.84]}\)} \\
& Cost & \shortstack{\(5.66 \pm 3.77\)\\\([3.99,7.62]\)} & \shortstack{\(5.06 \pm 11.55\)\\\([1.02,11.54]\)} & \shortstack{\(8.90 \pm 8.48\)\\\([4.92,13.10]\)} & \shortstack{\(15.02 \pm 9.36\)\\\([11.34,20.09]\)} & \shortstack{\(9.97 \pm 2.53\)\\\([8.63,11.17]\)} & \shortstack{\(\mathbf{2.32 \pm 1.47}\)\\\(\mathbf{[1.67,3.11]}\)} \\

\multirow{2}{*}{Demand Response}
& Rew. & \shortstack{\(-7.94 \pm 0.59\)\\\([-8.24,-7.66]\)} & \shortstack{\(-8.07 \pm 0.57\)\\\([-8.33,-7.78]\)} & \shortstack{\(-8.49 \pm 0.16\)\\\([-8.56,-8.41]\)} & \shortstack{\(-8.47 \pm 0.09\)\\\([-8.50,-8.42]\)} & \shortstack{\(-8.90 \pm 0.10\)\\\([-8.95,-8.85]\)} & \shortstack{\(\mathbf{-8.83 \pm 0.05}\)\\\(\mathbf{[-8.85,-8.81]}\)} \\
& Cost & \shortstack{\(8.06 \pm 3.84\)\\\([6.16,9.94]\)} & \shortstack{\(18.30 \pm 10.78\)\\\([12.63,23.42]\)} & \shortstack{\(6.25 \pm 2.54\)\\\([5.10,7.55]\)} & \shortstack{\(28.34 \pm 0.12\)\\\([28.28,28.40]\)} & \shortstack{\(10.46 \pm 2.92\)\\\([9.06,11.90]\)} & \shortstack{\(\mathbf{4.64 \pm 0.72}\)\\\(\mathbf{[4.33,5.00]}\)} \\

\multirow{2}{*}{Solar Ramp}
& Rew. & \shortstack{\(-7.84 \pm 0.51\)\\\([-8.11,-7.59]\)} & \shortstack{\(-8.13 \pm 0.49\)\\\([-8.37,-7.89]\)} & \shortstack{\(-8.55 \pm 0.37\)\\\([-8.72,-8.37]\)} & \shortstack{\(-8.53 \pm 0.09\)\\\([-8.57,-8.48]\)} & \shortstack{\(-8.94 \pm 0.13\)\\\([-9.00,-8.87]\)} & \shortstack{\(\mathbf{-9.11 \pm 0.11}\)\\\(\mathbf{[-9.17,-9.05]}\)} \\
& Cost & \shortstack{\(7.33 \pm 3.82\)\\\([5.43,9.16]\)} & \shortstack{\(6.52 \pm 3.92\)\\\([4.41,8.33]\)} & \shortstack{\(7.32 \pm 3.62\)\\\([5.44,9.01]\)} & \shortstack{\(43.57 \pm 24.01\)\\\([31.78,55.64]\)} & \shortstack{\(10.84 \pm 2.57\)\\\([9.56,11.99]\)} & \shortstack{\(\mathbf{3.10 \pm 0.78}\)\\\(\mathbf{[2.72,3.50]}\)} \\
\bottomrule
\end{tabular}
\end{adjustbox}
\end{table*}

\paragraph{Atari RAM.} Table~\ref{tab:atari_ram_full} reports the raw evaluation return for every game and dataset size, together with the expert and non-expert reference returns used for normalization. The expert-normalized averages in Table~\ref{tab:atari_budget_profile} are computed from these values.

\begin{table*}[hbpt!]
\centering
\caption{\textbf{Atari RAM --- Full Results (All Dataset Sizes).} At the final training step, we average raw return over 10 episodes per evaluation seed and report the mean ± std of the five seed averages. Size column shows total mixed transitions from a 1:1 expert/non-expert dataset (\textit{Full}\,\(\approx\)\,500K). Expert and non-expert columns show the mean return of expert and non-expert policies. \textbf{Bold} = best per row.}
\label{tab:atari_ram_full}
\vspace{0.5em}
\resizebox{\textwidth}{!}{%
\begin{tabular}{ll cc cccc cc}
\toprule
& & \multicolumn{2}{c}{\textbf{Reference}} & \multicolumn{4}{c}{\textbf{NN}} & \multicolumn{2}{c}{\textbf{GBT}} \\
\textbf{Game} & \textbf{Size} & Non-Expert & Expert & CQL & DT & AWR & BCQ & AWR & \textbf{Split-RL} \\
\midrule
  Qbert & 10K & 155.3 & 15704.6 & 282.0 $\pm$ 29.4 & 12930.5 $\pm$ 1393.7 & 10135.0 $\pm$ 1048.2 & 5729.0 $\pm$ 2120.7 & 10351.0 $\pm$ 987.0 & \textbf{15167.0 $\pm$ 421.7} \\
   & 20K &  &  & 5670.0 $\pm$ 1788.5 & 10998.5 $\pm$ 1699.9 & 13983.0 $\pm$ 1308.7 & 6606.0 $\pm$ 1368.7 & 9899.0 $\pm$ 927.6 & \textbf{15723.0 $\pm$ 36.1} \\
   & 50K &  &  & 11366.0 $\pm$ 1167.2 & 7970.5 $\pm$ 1261.7 & 15691.0 $\pm$ 45.1 & 459.0 $\pm$ 28.8 & 7807.0 $\pm$ 635.3 & \textbf{15711.0 $\pm$ 38.0} \\
   & 100K &  &  & 6009.0 $\pm$ 2301.2 & 5911.5 $\pm$ 334.2 & \textbf{15711.0 $\pm$ 38.0} & 688.0 $\pm$ 27.3 & 11105.0 $\pm$ 439.3 & \textbf{15711.0 $\pm$ 38.0} \\
   & 200K &  &  & 10455.0 $\pm$ 1378.7 & 4670.0 $\pm$ 1242.6 & 10446.0 $\pm$ 1372.0 & 440.0 $\pm$ 45.0 & 8140.0 $\pm$ 501.4 & \textbf{15711.0 $\pm$ 38.0} \\
   & Full &  &  & 8158.0 $\pm$ 1312.0 & 4574.0 $\pm$ 28.7 & \textbf{15711.0 $\pm$ 38.0} & 402.0 $\pm$ 14.9 & 10561.0 $\pm$ 536.1 & \textbf{15711.0 $\pm$ 38.0} \\
\midrule
  Seaquest & 10K & 80.9 & 1822.2 & 547.6 $\pm$ 266.7 & 1168.0 $\pm$ 254.8 & 712.0 $\pm$ 248.4 & 510.8 $\pm$ 269.0 & 1175.6 $\pm$ 66.3 & \textbf{1596.4 $\pm$ 137.1} \\
   & 20K &  &  & 128.0 $\pm$ 28.6 & 1022.0 $\pm$ 187.8 & 598.4 $\pm$ 193.3 & 639.2 $\pm$ 259.8 & 1249.6 $\pm$ 59.1 & \textbf{1725.6 $\pm$ 61.5} \\
   & 50K &  &  & 29.2 $\pm$ 4.8 & 1496.4 $\pm$ 14.8 & 656.0 $\pm$ 68.2 & 55.2 $\pm$ 13.2 & 1226.8 $\pm$ 67.5 & \textbf{1821.6 $\pm$ 4.8} \\
   & 100K &  &  & 361.6 $\pm$ 50.4 & 1500.4 $\pm$ 7.6 & 548.8 $\pm$ 107.6 & 212.4 $\pm$ 51.2 & 1164.4 $\pm$ 61.3 & \textbf{1821.6 $\pm$ 4.8} \\
   & 200K &  &  & 854.0 $\pm$ 91.5 & 1476.8 $\pm$ 9.4 & 1354.4 $\pm$ 119.3 & 84.4 $\pm$ 1.5 & 755.2 $\pm$ 62.1 & \textbf{1821.6 $\pm$ 4.8} \\
   & Full &  &  & 364.0 $\pm$ 27.9 & 1573.2 $\pm$ 15.7 & 1491.6 $\pm$ 119.9 & 536.0 $\pm$ 136.6 & 943.2 $\pm$ 72.4 & \textbf{1821.6 $\pm$ 4.8} \\
\midrule
  Pong & 10K & -20.3 & 21.0 & -21.0 $\pm$ 0.0 & \textbf{6.4 $\pm$ 6.5} & -20.1 $\pm$ 0.4 & -1.3 $\pm$ 6.7 & -1.8 $\pm$ 4.2 & 4.9 $\pm$ 3.8 \\
   & 20K &  &  & -5.4 $\pm$ 5.5 & 5.5 $\pm$ 6.1 & 4.0 $\pm$ 7.4 & -8.5 $\pm$ 6.7 & -1.6 $\pm$ 4.2 & \textbf{14.3 $\pm$ 5.0} \\
   & 50K &  &  & -20.5 $\pm$ 0.2 & \textbf{21.0 $\pm$ 0.0} & 14.6 $\pm$ 4.8 & 11.4 $\pm$ 3.2 & -15.1 $\pm$ 1.7 & 14.8 $\pm$ 4.7 \\
   & 100K &  &  & -6.3 $\pm$ 6.2 & 19.4 $\pm$ 1.4 & -5.0 $\pm$ 5.5 & -11.9 $\pm$ 4.7 & -16.9 $\pm$ 1.8 & \textbf{21.0 $\pm$ 0.0} \\
   & 200K &  &  & -19.3 $\pm$ 0.2 & 0.7 $\pm$ 6.7 & -7.1 $\pm$ 3.5 & 11.4 $\pm$ 5.4 & -18.3 $\pm$ 0.6 & \textbf{21.0 $\pm$ 0.0} \\
   & Full &  &  & -19.8 $\pm$ 0.2 & 18.0 $\pm$ 2.9 & -4.5 $\pm$ 5.2 & -13.2 $\pm$ 3.2 & -14.4 $\pm$ 3.9 & \textbf{21.0 $\pm$ 0.0} \\
\midrule
  SpaceInvaders & 10K & 145.2 & 1229.1 & 924.8 $\pm$ 188.1 & 854.6 $\pm$ 139.6 & 1138.4 $\pm$ 98.2 & 385.8 $\pm$ 70.2 & 517.6 $\pm$ 38.3 & \textbf{1263.6 $\pm$ 43.1} \\
   & 20K &  &  & 157.8 $\pm$ 23.9 & 1037.8 $\pm$ 114.7 & 1008.8 $\pm$ 81.2 & 511.6 $\pm$ 93.3 & 469.8 $\pm$ 40.3 & \textbf{1156.4 $\pm$ 93.4} \\
   & 50K &  &  & 417.6 $\pm$ 33.6 & 1229.0 $\pm$ 25.7 & 833.2 $\pm$ 53.6 & 257.8 $\pm$ 46.3 & 688.6 $\pm$ 75.3 & \textbf{1251.6 $\pm$ 44.5} \\
   & 100K &  &  & 262.8 $\pm$ 30.3 & 1229.0 $\pm$ 25.7 & 821.4 $\pm$ 77.9 & 201.0 $\pm$ 11.0 & 441.6 $\pm$ 63.3 & \textbf{1251.6 $\pm$ 44.5} \\
   & 200K &  &  & 486.6 $\pm$ 62.1 & 1229.0 $\pm$ 25.7 & 997.8 $\pm$ 54.3 & 236.2 $\pm$ 49.0 & 451.6 $\pm$ 51.8 & \textbf{1251.6 $\pm$ 44.5} \\
   & Full &  &  & 505.0 $\pm$ 33.5 & 1229.0 $\pm$ 25.7 & 915.8 $\pm$ 78.5 & 370.8 $\pm$ 46.3 & 429.0 $\pm$ 28.5 & \textbf{1251.6 $\pm$ 44.5} \\
\midrule
  Alien & 10K & 206.0 & 5102.5 & 675.8 $\pm$ 202.9 & 898.2 $\pm$ 231.7 & 761.6 $\pm$ 327.9 & 763.2 $\pm$ 321.9 & \textbf{1065.0 $\pm$ 207.6} & 1042.2 $\pm$ 386.6 \\
   & 20K &  &  & 924.6 $\pm$ 322.1 & 1094.6 $\pm$ 129.0 & 1470.2 $\pm$ 609.5 & \textbf{1603.2 $\pm$ 590.8} & 945.8 $\pm$ 226.0 & 996.4 $\pm$ 73.1 \\
   & 50K &  &  & 921.8 $\pm$ 185.1 & 1763.6 $\pm$ 154.5 & 2148.6 $\pm$ 272.9 & 371.2 $\pm$ 30.0 & 1711.8 $\pm$ 248.0 & \textbf{4031.2 $\pm$ 643.8} \\
   & 100K &  &  & 1526.6 $\pm$ 336.9 & 1783.6 $\pm$ 168.5 & 2088.6 $\pm$ 397.8 & 429.6 $\pm$ 38.5 & 1752.2 $\pm$ 280.9 & \textbf{4723.4 $\pm$ 657.4} \\
   & 200K &  &  & 997.4 $\pm$ 147.5 & 1734.4 $\pm$ 61.8 & 3497.6 $\pm$ 486.1 & 559.2 $\pm$ 76.7 & 2125.4 $\pm$ 424.5 & \textbf{5037.8 $\pm$ 733.4} \\
   & Full &  &  & 1257.8 $\pm$ 153.5 & 1769.0 $\pm$ 82.0 & 3094.6 $\pm$ 534.1 & 664.2 $\pm$ 104.0 & 2232.2 $\pm$ 378.0 & \textbf{5156.2 $\pm$ 677.9} \\
\midrule
  MsPacman & 10K & 252.2 & 3597.6 & 2064.0 $\pm$ 615.1 & 1186.8 $\pm$ 280.3 & 3560.8 $\pm$ 839.6 & 3551.2 $\pm$ 841.2 & 3228.8 $\pm$ 266.0 & \textbf{4031.2 $\pm$ 614.9} \\
   & 20K &  &  & 2462.8 $\pm$ 29.0 & 1003.4 $\pm$ 256.5 & 1910.8 $\pm$ 296.6 & 460.8 $\pm$ 43.2 & 3310.4 $\pm$ 297.1 & \textbf{4064.8 $\pm$ 600.1} \\
   & 50K &  &  & 3313.6 $\pm$ 364.7 & 1091.2 $\pm$ 426.6 & 1854.0 $\pm$ 26.1 & 767.6 $\pm$ 82.7 & 3512.0 $\pm$ 370.2 & \textbf{4064.8 $\pm$ 600.1} \\
   & 100K &  &  & 4020.0 $\pm$ 615.6 & 1284.2 $\pm$ 220.3 & 4013.6 $\pm$ 622.1 & 2103.6 $\pm$ 38.3 & 3665.6 $\pm$ 393.4 & \textbf{4064.8 $\pm$ 600.1} \\
   & 200K &  &  & 2187.6 $\pm$ 77.1 & 1252.8 $\pm$ 190.0 & \textbf{4064.8 $\pm$ 600.1} & 1370.0 $\pm$ 279.8 & 3755.6 $\pm$ 410.8 & \textbf{4064.8 $\pm$ 600.1} \\
   & Full &  &  & 2441.2 $\pm$ 24.9 & 448.0 $\pm$ 72.1 & 1854.0 $\pm$ 26.1 & 878.8 $\pm$ 260.6 & 3034.4 $\pm$ 195.8 & \textbf{4064.8 $\pm$ 600.1} \\
\midrule
  Gravitar & 10K & 171.5 & 456.9 & 344.0 $\pm$ 53.0 & \textbf{390.0 $\pm$ 35.4} & 320.0 $\pm$ 43.2 & 276.0 $\pm$ 44.2 & 325.0 $\pm$ 41.8 & 383.0 $\pm$ 35.4 \\
   & 20K &  &  & 230.0 $\pm$ 52.2 & 314.0 $\pm$ 64.5 & 329.0 $\pm$ 32.9 & 241.0 $\pm$ 38.0 & 289.0 $\pm$ 102.0 & \textbf{368.0 $\pm$ 65.2} \\
   & 50K &  &  & 232.0 $\pm$ 33.7 & 370.0 $\pm$ 70.5 & 383.0 $\pm$ 74.4 & 293.0 $\pm$ 46.1 & \textbf{401.0 $\pm$ 59.4} & 371.0 $\pm$ 47.3 \\
   & 100K &  &  & 267.0 $\pm$ 62.3 & 409.0 $\pm$ 81.0 & 378.0 $\pm$ 86.5 & 311.0 $\pm$ 49.6 & 293.0 $\pm$ 48.9 & \textbf{464.0 $\pm$ 57.7} \\
   & 200K &  &  & 357.0 $\pm$ 18.6 & 353.0 $\pm$ 78.0 & 285.0 $\pm$ 57.4 & 332.0 $\pm$ 79.0 & 454.0 $\pm$ 38.4 & \textbf{472.0 $\pm$ 60.5} \\
   & Full &  &  & 350.0 $\pm$ 73.7 & \textbf{431.0 $\pm$ 26.3} & 261.0 $\pm$ 33.4 & 391.0 $\pm$ 88.3 & 307.0 $\pm$ 52.4 & 393.0 $\pm$ 49.5 \\
\midrule
  KungFuMaster & 10K & 331.5 & 34312.9 & 2476.0 $\pm$ 660.6 & 1050.0 $\pm$ 719.1 & 1294.0 $\pm$ 1324.4 & 1098.0 $\pm$ 1149.2 & 1636.0 $\pm$ 179.6 & \textbf{7772.0 $\pm$ 2009.9} \\
   & 20K &  &  & 2194.0 $\pm$ 737.0 & 738.0 $\pm$ 127.5 & 1280.0 $\pm$ 2150.3 & 588.0 $\pm$ 207.3 & 3912.0 $\pm$ 754.3 & \textbf{24080.0 $\pm$ 3484.8} \\
   & 50K &  &  & 914.0 $\pm$ 432.9 & 2570.0 $\pm$ 1713.7 & 740.0 $\pm$ 1180.5 & 932.0 $\pm$ 333.4 & 3656.0 $\pm$ 448.4 & \textbf{27488.0 $\pm$ 2743.8} \\
   & 100K &  &  & 2252.0 $\pm$ 805.2 & 5706.0 $\pm$ 3005.5 & 2926.0 $\pm$ 1527.8 & 1040.0 $\pm$ 57.6 & 5088.0 $\pm$ 1413.4 & \textbf{34992.0 $\pm$ 1815.6} \\
   & 200K &  &  & 17692.0 $\pm$ 1517.0 & 3910.0 $\pm$ 1297.2 & 14236.0 $\pm$ 3561.0 & 1388.0 $\pm$ 588.8 & 5840.0 $\pm$ 1427.3 & \textbf{34992.0 $\pm$ 1815.6} \\
   & Full &  &  & 12720.0 $\pm$ 3149.9 & 2904.0 $\pm$ 1215.4 & 18258.0 $\pm$ 2892.9 & 2716.0 $\pm$ 678.9 & 5312.0 $\pm$ 2157.8 & \textbf{34992.0 $\pm$ 1815.6} \\
\midrule
  Frostbite & 10K & 74.0 & 931.2 & 549.6 $\pm$ 239.0 & 858.8 $\pm$ 324.0 & 875.2 $\pm$ 262.6 & \textbf{1009.0 $\pm$ 264.5} & 144.0 $\pm$ 9.0 & 966.6 $\pm$ 276.9 \\
   & 20K &  &  & 636.2 $\pm$ 205.3 & \textbf{992.0 $\pm$ 253.0} & 986.8 $\pm$ 273.5 & 943.6 $\pm$ 237.5 & 920.0 $\pm$ 257.9 & 971.2 $\pm$ 229.4 \\
   & 50K &  &  & 67.2 $\pm$ 17.0 & \textbf{1060.2 $\pm$ 330.4} & 855.6 $\pm$ 276.4 & 111.6 $\pm$ 5.4 & 163.0 $\pm$ 7.2 & 921.4 $\pm$ 247.7 \\
   & 100K &  &  & 111.2 $\pm$ 9.3 & \textbf{1116.0 $\pm$ 269.2} & 886.0 $\pm$ 257.3 & 107.4 $\pm$ 4.2 & 158.2 $\pm$ 10.3 & 970.4 $\pm$ 247.3 \\
   & 200K &  &  & 134.6 $\pm$ 17.3 & \textbf{1071.8 $\pm$ 282.8} & 473.6 $\pm$ 173.5 & 54.4 $\pm$ 10.4 & 139.8 $\pm$ 7.9 & 924.8 $\pm$ 245.6 \\
   & Full &  &  & 127.2 $\pm$ 9.6 & \textbf{1103.6 $\pm$ 261.8} & 1009.0 $\pm$ 263.0 & 853.8 $\pm$ 367.3 & 150.4 $\pm$ 6.0 & 907.8 $\pm$ 251.2 \\
\midrule
  Amidar & 10K & 1.2 & 1049.8 & 66.2 $\pm$ 19.0 & 381.6 $\pm$ 117.0 & 252.0 $\pm$ 84.0 & 0.0 $\pm$ 0.0 & \textbf{513.2 $\pm$ 59.6} & 472.7 $\pm$ 55.6 \\
   & 20K &  &  & 2.8 $\pm$ 0.4 & 916.5 $\pm$ 103.0 & 882.0 $\pm$ 126.0 & 811.7 $\pm$ 126.1 & 699.8 $\pm$ 45.3 & \textbf{1050.0 $\pm$ 0.0} \\
   & 50K &  &  & 300.2 $\pm$ 21.8 & \textbf{1049.9 $\pm$ 0.1} & 881.8 $\pm$ 55.9 & 1049.8 $\pm$ 0.1 & 629.7 $\pm$ 47.1 & 1049.8 $\pm$ 0.1 \\
   & 100K &  &  & 166.1 $\pm$ 26.3 & \textbf{1049.9 $\pm$ 0.1} & 1049.8 $\pm$ 0.1 & 888.8 $\pm$ 46.6 & 569.6 $\pm$ 54.7 & 1049.8 $\pm$ 0.1 \\
   & 200K &  &  & 265.2 $\pm$ 68.8 & \textbf{1049.9 $\pm$ 0.1} & 943.6 $\pm$ 40.0 & 17.0 $\pm$ 0.9 & 572.8 $\pm$ 37.1 & 1049.8 $\pm$ 0.1 \\
   & Full &  &  & 754.4 $\pm$ 91.0 & \textbf{1049.9 $\pm$ 0.1} & 1049.8 $\pm$ 0.1 & 927.7 $\pm$ 51.6 & 568.4 $\pm$ 54.5 & 1049.8 $\pm$ 0.1 \\
\midrule
  BankHeist & 10K & 14.1 & 1361.2 & 67.8 $\pm$ 47.8 & 647.4 $\pm$ 85.2 & 43.2 $\pm$ 40.6 & 23.2 $\pm$ 4.9 & \textbf{1373.8 $\pm$ 1.6} & 1354.2 $\pm$ 16.1 \\
   & 20K &  &  & 169.6 $\pm$ 161.7 & 1164.2 $\pm$ 47.3 & 98.4 $\pm$ 101.9 & 58.4 $\pm$ 51.4 & \textbf{1377.2 $\pm$ 1.5} & 1368.0 $\pm$ 5.8 \\
   & 50K &  &  & 688.8 $\pm$ 121.8 & 1331.6 $\pm$ 18.0 & 272.0 $\pm$ 160.6 & 64.2 $\pm$ 103.9 & \textbf{1375.4 $\pm$ 1.5} & 1355.4 $\pm$ 16.0 \\
   & 100K &  &  & 1131.4 $\pm$ 16.3 & 1371.6 $\pm$ 10.1 & 127.2 $\pm$ 101.7 & 64.2 $\pm$ 23.7 & \textbf{1372.0 $\pm$ 3.0} & 1369.2 $\pm$ 8.9 \\
   & 200K &  &  & 1362.0 $\pm$ 1.3 & \textbf{1374.4 $\pm$ 4.8} & 14.0 $\pm$ 2.5 & 152.0 $\pm$ 104.9 & 1374.2 $\pm$ 1.8 & 1365.8 $\pm$ 7.1 \\
   & Full &  &  & 1362.0 $\pm$ 6.7 & 1370.0 $\pm$ 10.8 & 641.0 $\pm$ 161.5 & 28.8 $\pm$ 1.8 & \textbf{1372.8 $\pm$ 3.8} & 1366.2 $\pm$ 5.6 \\
\bottomrule
\end{tabular}%
}
\end{table*}

Table~\ref{tab:atari_bc_ablation_raw} gives the corresponding per-game returns for the BC target ablation summarized in Table~\ref{tab:atari_ablation}, at the full dataset size.

\begin{table*}[hbt!]
\centering
\caption{\textbf{BC target ablation --- per-game raw returns.} At the final training step, we average raw return over 10 episodes per evaluation seed and report the mean ± std of the five seed averages. \emph{Non-Expert} and \emph{Expert} columns show reference returns used for normalization. See Table~\ref{tab:atari_ablation} for method descriptions. \textbf{Bold} = best per row.}
\label{tab:atari_bc_ablation_raw}
\vspace{0.5em}
\resizebox{\textwidth}{!}{%
\begin{tabular}{lccccccccc}
\toprule
& & & \multicolumn{3}{c}{\textbf{NN}} & \multicolumn{3}{c}{\textbf{GBT}} & \\
\cmidrule(lr){4-6} \cmidrule(lr){7-9} \cmidrule(lr){10-10}
& & & \multicolumn{3}{c}{AWR + BC} & \multicolumn{3}{c}{AWR + BC} & \textbf{Split-RL} \\
\cmidrule(lr){4-6} \cmidrule(lr){7-9}
\textbf{Environment} & Non-Expert & Expert & all & expert & masked & all & expert & masked & \\
\midrule
  Qbert & 155.3 & 15704.6 & \textbf{15711.0 \(\pm\) 38.0} & \textbf{15711.0 \(\pm\) 38.0} & 15594.0 \(\pm\) 29.8 & 9878.0 \(\pm\) 308.8 & \textbf{15711.0 \(\pm\) 38.0} & 10322.0 \(\pm\) 579.0 & \textbf{15711.0 \(\pm\) 38.0} \\
  Seaquest & 80.9 & 1822.2 & 1238.4 \(\pm\) 114.4 & 1465.2 \(\pm\) 157.9 & 712.8 \(\pm\) 161.7 & 953.2 \(\pm\) 35.6 & 1258.0 \(\pm\) 60.2 & 1005.6 \(\pm\) 105.8 & \textbf{1821.6 \(\pm\) 4.8} \\
  Pong & -20.3 & 21.0 & -5.0 \(\pm\) 5.5 & -2.0 \(\pm\) 4.3 & -10.9 \(\pm\) 3.4 & 0.7 \(\pm\) 5.6 & \textbf{21.0 \(\pm\) 0.0} & 11.4 \(\pm\) 5.4 & \textbf{21.0 \(\pm\) 0.0} \\
  SpaceInvaders & 145.2 & 1229.1 & 551.6 \(\pm\) 46.8 & 1245.6 \(\pm\) 43.5 & 1220.0 \(\pm\) 46.0 & 542.4 \(\pm\) 20.3 & 801.4 \(\pm\) 77.2 & 471.6 \(\pm\) 34.2 & \textbf{1251.6 \(\pm\) 44.5} \\
  Alien & 206.0 & 5102.5 & 4196.0 \(\pm\) 461.9 & 3510.0 \(\pm\) 565.4 & 3276.8 \(\pm\) 232.2 & 2886.8 \(\pm\) 268.1 & 4094.0 \(\pm\) 333.9 & 1254.0 \(\pm\) 76.2 & \textbf{5156.2 \(\pm\) 677.9} \\
  MsPacman & 252.2 & 3597.6 & 4020.8 \(\pm\) 619.0 & 4064.8 \(\pm\) 600.1 & 3775.2 \(\pm\) 482.7 & 4293.2 \(\pm\) 562.4 & \textbf{4612.0 \(\pm\) 668.4} & 4298.8 \(\pm\) 553.6 & 4064.8 \(\pm\) 600.1 \\
  Gravitar & 171.5 & 456.9 & 355.0 \(\pm\) 98.5 & 347.0 \(\pm\) 52.0 & 254.0 \(\pm\) 78.9 & \textbf{448.0 \(\pm\) 44.6} & 436.0 \(\pm\) 13.2 & 266.0 \(\pm\) 67.6 & 393.0 \(\pm\) 49.5 \\
  KungFuMaster & 331.5 & 34312.9 & 4642.0 \(\pm\) 2638.3 & 6480.0 \(\pm\) 3962.9 & 0.0 \(\pm\) 0.0 & 8128.0 \(\pm\) 982.5 & 11524.0 \(\pm\) 2301.4 & 1976.0 \(\pm\) 363.2 & \textbf{34992.0 \(\pm\) 1815.6} \\
  Frostbite & 74.0 & 931.2 & 853.6 \(\pm\) 221.7 & 959.8 \(\pm\) 229.3 & \textbf{1056.6 \(\pm\) 243.6} & 959.0 \(\pm\) 264.2 & 944.8 \(\pm\) 238.1 & 152.0 \(\pm\) 9.8 & 907.8 \(\pm\) 251.2 \\
  Amidar & 1.2 & 1049.8 & 836.2 \(\pm\) 63.4 & \textbf{1049.8 \(\pm\) 0.1} & 896.0 \(\pm\) 45.2 & 649.0 \(\pm\) 72.9 & \textbf{1049.8 \(\pm\) 0.1} & 733.1 \(\pm\) 54.9 & \textbf{1049.8 \(\pm\) 0.1} \\
  BankHeist & 14.1 & 1361.2 & 915.4 \(\pm\) 180.0 & 1365.6 \(\pm\) 12.7 & 16.8 \(\pm\) 2.3 & 1128.0 \(\pm\) 89.9 & \textbf{1376.6 \(\pm\) 1.6} & 1260.0 \(\pm\) 101.5 & 1366.2 \(\pm\) 5.6 \\
\bottomrule
\end{tabular}%
}
\end{table*}

\section{Interpretability: Routing Density Under Policy Rollouts}
\label{app:interpretability}

\paragraph{What is measured.} The guidance label is used only during training, to score candidate splits; at evaluation it is never an input. To ask whether the ensemble has internalized the conflict structure, we read it back out of the leaves. At tree construction each leaf records the fraction of its training samples that carried the cost label. We pass a visited state through the ensemble, read that fraction from the leaf it reaches in each tree, and average over trees. We call the result the \emph{routing density} of that state: the share of cost-labelled training data sharing a region with it. A model that routes by position alone would assign cost- and reward-labelled states at the same location the same density; a model that routes by conflict structure separates them.

\begin{figure}[htbp!]
\centering
    \includegraphics[width=\textwidth]{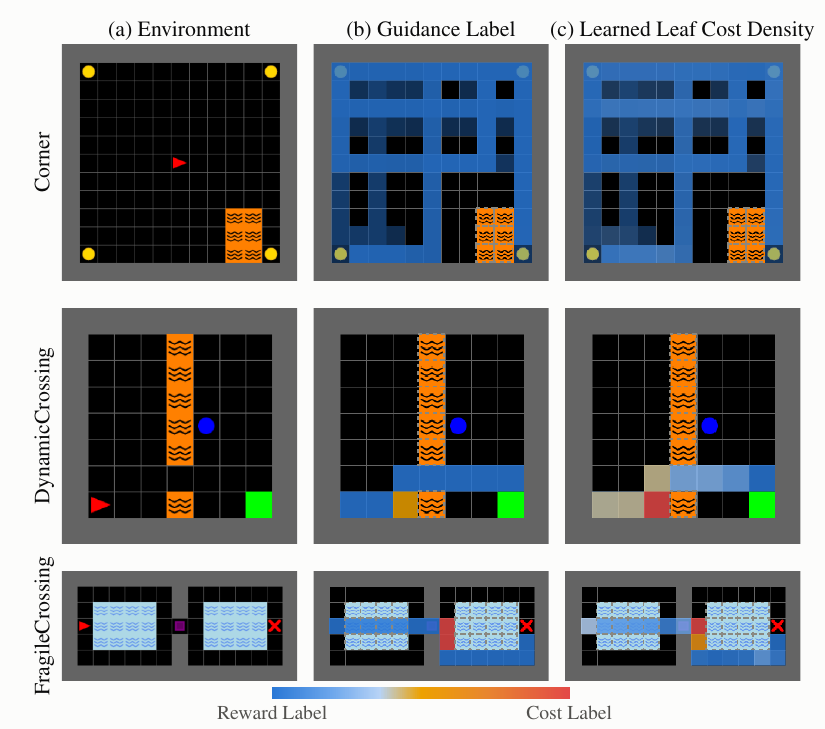}
    \caption{\textbf{Guided routing density under policy rollouts, all environments.} Same
    measurement as Figure~\ref{fig:interp_main}, for Corner, DynamicCrossing, and
    FragileCrossing.}
    \label{fig:interp_traj}
\end{figure}

\paragraph{Protocol.} We run 60 evaluation episodes per environment with actions sampled from the trained policy and record every visited state. Figures~\ref{fig:interp_main} and~\ref{fig:interp_traj} aggregate these visits per cell: panel~(a) shows the environment, panel~(b) the fraction of visits to that cell at which the ground-truth guidance label fired, and panel~(c) the mean routing density of the leaves those same visits reached. Both panels describe the same set of events and are therefore directly comparable. Cells that were never visited are left grey. The recorded rollouts cover 8632 steps over 95 cells in Corner, 686 steps over 11 cells in DynamicCrossing, and 5329 steps over 16 cells in FragileCrossing.

\paragraph{Result.} In FragileCrossing, panels~(b) and~(c) coincide: the cells where the cost label fires during rollouts are the cells whose visits reach cost-heavy leaves. DynamicCrossing shows the same pattern. No label is supplied at evaluation, so the ensemble recovers the conflict structure from the observation alone. The partition is therefore readable after training: high routing density indicates a higher proportion of cost-labelled samples.

\paragraph{Corner.} Corner's guidance label is a property of the cell and heading rather than of the trajectory, and its trained policy avoids the lava entirely, so cost-labelled states are almost never visited under rollouts: only 31 of its 468 reachable state configurations carry the cost label, and the policy reaches few of them. Its panels are therefore close to uniform here. We show it under rollouts anyway, for consistency with the other two environments; enumerating all reachable states instead of sampling trajectories recovers the separation.

\paragraph{Scope.} This probe reports what the leaves contain, not which features the splits use or how much each one contributes. It shows that routing correlates with the conflict structure under the visited state distribution; it does not attribute that behaviour to particular split variables, and it does not measure how much of the chosen action is explained by the routed region. Split-level attribution and feature importance are natural extensions that we leave to future work.

\section{Ablations}
\label{app:ablations}

\subsection{Label-Purity Regularization}

In Figure~\ref{fig:lambda_penalty_citylearn} we ablate \(\lambda_0\) in CityLearn Solar Ramp
Reserve. Removing the purity penalty weakens cost control, while nonzero \(\lambda_0\) lowers
cost by discouraging label-mixing splits in conflict windows.


\subsection{Robustness to Label Noise}
We corrupt the guidance label with probability \(p \in \{0, 0.1, 0.25, 0.5, 1.0\}\): each label is retained with probability \(1-p\) and independently redrawn uniformly at random with probability \(p\) before it reaches the split-selection procedure. Figure~\ref{fig:minigrid_noise}
shows reward and cost across all three MiniGrid environments as \(p\) increases. Performance is
stable up to \(p=0.25\): cost stays near its \(p=0\) value in both DynamicCrossing and
FragileCrossing, and reward is essentially unchanged across the whole sweep, including at
\(p=1.0\), where the label carries no information. Cost rises sharply once \(p \geq 0.5\), which
is expected: Split-RL acts on the label rather than consulting it, so an uninformative label
produces arbitrary routing rather than a graceful fallback to unguided training.
The reward/cost split in this result matches the failure mode the paper is built around: unsafe
states are a minority of the distribution, so cost is the rare-but-critical signal that guidance
protects against dilution; reward, driven by the majority distribution, is far less sensitive to
losing that guidance.


\subsection{Privileged-Information Ablation}
\label{app:ablation_raw}
Section~\ref{sec:results} compares all five constrained-RL baselines under three training
conditions. No-label results are already reported in
Tables~\ref{tab:full_results_minigrid}--\ref{tab:full_results_citylearn}.
Tables~\ref{tab:ablation_raw_sumo_labelinobs}--\ref{tab:ablation_raw_minigrid_masked} below report
the same metrics for Label-in-Obs and Masked, one table per domain per condition, in the same
five-algorithm-column format as Table~\ref{tab:full_results_sumo}. Split-RL is not included in
either condition, since neither is a Split-RL variant. Mean, std, and the 95\% bootstrap CI are shown for each entry, matching Tables~\ref{tab:full_results_minigrid}--\ref{tab:full_results_citylearn}.

Four of the five baselines change by at most one scenario under either condition. CPO under Masked is the exception, gaining four, but the gains are not uniform relative to its own no-label results: in FragileCrossing feasibility comes with reward falling from \(0.48\) to \(0.26\); in Corner its cost of \(0.10 \pm 0.39\) lies on the \(0.1\) limit with a bootstrap CI reaching \(0.31\); in Single/Premium masking stabilizes the run, with cost falling from \(38.26 \pm 84.51\) to \(11.36 \pm 3.49\); and in Solar Ramp Reserve improves on both axes, from reward \(-7.84\) at cost \(7.33\) to \(-7.65\) at \(2.85\). Where CPO was already infeasible it can also degrade: in DynamicCrossing its cost rises from \(0.34\) to \(17.87\), and in Arbitrage vs.\ Buffer its reward falls from \(-7.96\) to \(-29.75\). Across all five baselines, Masked changes \(+4\) of 75 scenario verdicts and Label-in-Obs \(-1\).

\subsubsection{Label-in-Obs}
Tables~\ref{tab:ablation_raw_sumo_labelinobs}--\ref{tab:ablation_raw_minigrid_labelinobs} report SUMO-RL, CityLearn, and MiniGrid with the guidance label appended to the observation as a one-hot feature.

\begin{table*}[hbt!]
\centering
\caption{\textbf{Label-in-Obs, SUMO-RL.} Mean \(\pm\) std, with 95\% bootstrap CI, over seeds.
The guidance label is concatenated into the observation; otherwise standard baselines. Reward is
in units of \(10^{3}\); cost is unscaled. \textbf{Bold} marks entries where the method is feasible (cost within the limit in Table~\ref{tab:evaluation_cost_limits}); unbolded entries exceed the limit.}
\label{tab:ablation_raw_sumo_labelinobs}
\renewcommand{\arraystretch}{1.5}
\resizebox{\textwidth}{!}{%
\begin{tabular}{lllrrrrr}
\toprule
\textbf{Environment} & \textbf{Scenario} & \textbf{Metric} & \textbf{CPO} & \textbf{CUP} & \textbf{IPO} & \textbf{PPOLag} & \textbf{PPOLag (GBT)} \\
\midrule
\multirow{6}{*}{Arterial 4x4}
 & \multirow{2}{*}{Bus}
& Rew. & \shortstack{\(\mathbf{-1.403 \pm 0.341}\)\\\(\mathbf{[-1.576,-1.248]}\)} & \shortstack{\(\mathbf{-1.284 \pm 0.186}\)\\\(\mathbf{[-1.373,-1.197]}\)} & \shortstack{\(\mathbf{-1.295 \pm 0.195}\)\\\(\mathbf{[-1.395,-1.199]}\)} & \shortstack{\(-1.182 \pm 0.185\)\\\([-1.275,-1.091]\)} & \shortstack{\(-1.250 \pm 0.175\)\\\([-1.351,-1.179]\)} \\
& & Cost & \shortstack{\(\mathbf{3.76 \pm 2.49}\)\\\(\mathbf{[2.56,4.96]}\)} & \shortstack{\(\mathbf{1.96 \pm 2.04}\)\\\(\mathbf{[1.08,3.04]}\)} & \shortstack{\(\mathbf{4.63 \pm 4.81}\)\\\(\mathbf{[2.68,7.31]}\)} & \shortstack{\(8.11 \pm 7.48\)\\\([4.94,12.19]\)} & \shortstack{\(24.82 \pm 14.42\)\\\([18.09,31.92]\)} \\
 & \multirow{2}{*}{Convoy}
& Rew. & \shortstack{\(\mathbf{-1.483 \pm 0.354}\)\\\(\mathbf{[-1.648,-1.317]}\)} & \shortstack{\(\mathbf{-1.476 \pm 0.262}\)\\\(\mathbf{[-1.607,-1.352]}\)} & \shortstack{\(\mathbf{-1.264 \pm 0.160}\)\\\(\mathbf{[-1.349,-1.190]}\)} & \shortstack{\(-1.245 \pm 0.193\)\\\([-1.340,-1.154]\)} & \shortstack{\(-1.249 \pm 0.144\)\\\([-1.321,-1.183]\)} \\
& & Cost & \shortstack{\(\mathbf{3.89 \pm 3.22}\)\\\(\mathbf{[2.42,5.62]}\)} & \shortstack{\(\mathbf{1.77 \pm 1.71}\)\\\(\mathbf{[1.05,2.67]}\)} & \shortstack{\(\mathbf{1.57 \pm 1.48}\)\\\(\mathbf{[1.04,2.48]}\)} & \shortstack{\(11.70 \pm 4.04\)\\\([9.77,13.59]\)} & \shortstack{\(14.77 \pm 7.14\)\\\([11.78,18.78]\)} \\
 & \multirow{2}{*}{Premium}
& Rew. & \shortstack{\(\mathbf{-1.397 \pm 0.281}\)\\\(\mathbf{[-1.541,-1.264]}\)} & \shortstack{\(\mathbf{-1.485 \pm 0.327}\)\\\(\mathbf{[-1.653,-1.335]}\)} & \shortstack{\(\mathbf{-1.362 \pm 0.327}\)\\\(\mathbf{[-1.534,-1.212]}\)} & \shortstack{\(-1.194 \pm 0.216\)\\\([-1.302,-1.096]\)} & \shortstack{\(-1.174 \pm 0.136\)\\\([-1.243,-1.114]\)} \\
& & Cost & \shortstack{\(\mathbf{6.75 \pm 4.11}\)\\\(\mathbf{[4.96,8.91]}\)} & \shortstack{\(\mathbf{6.33 \pm 3.83}\)\\\(\mathbf{[4.59,8.10]}\)} & \shortstack{\(\mathbf{6.99 \pm 5.58}\)\\\(\mathbf{[4.75,9.93]}\)} & \shortstack{\(15.35 \pm 8.14\)\\\([12.14,19.89]\)} & \shortstack{\(34.91 \pm 13.59\)\\\([28.49,41.39]\)} \\
\midrule
\multirow{6}{*}{Grid 4x4}
 & \multirow{2}{*}{Bus}
& Rew. & \shortstack{\(\mathbf{-0.198 \pm 0.011}\)\\\(\mathbf{[-0.203,-0.192]}\)} & \shortstack{\(\mathbf{-0.138 \pm 0.005}\)\\\(\mathbf{[-0.140,-0.135]}\)} & \shortstack{\(\mathbf{-0.150 \pm 0.004}\)\\\(\mathbf{[-0.152,-0.148]}\)} & \shortstack{\(\mathbf{-0.119 \pm 0.004}\)\\\(\mathbf{[-0.121,-0.117]}\)} & \shortstack{\(\mathbf{-0.135 \pm 0.004}\)\\\(\mathbf{[-0.138,-0.133]}\)} \\
& & Cost & \shortstack{\(\mathbf{0.24 \pm 0.18}\)\\\(\mathbf{[0.16,0.33]}\)} & \shortstack{\(\mathbf{0.06 \pm 0.07}\)\\\(\mathbf{[0.03,0.10]}\)} & \shortstack{\(\mathbf{0.01 \pm 0.01}\)\\\(\mathbf{[0.01,0.02]}\)} & \shortstack{\(\mathbf{0.13 \pm 0.08}\)\\\(\mathbf{[0.10,0.18]}\)} & \shortstack{\(\mathbf{0.29 \pm 0.12}\)\\\(\mathbf{[0.23,0.34]}\)} \\
 & \multirow{2}{*}{Convoy}
& Rew. & \shortstack{\(\mathbf{-0.341 \pm 0.333}\)\\\(\mathbf{[-0.515,-0.251]}\)} & \shortstack{\(\mathbf{-0.163 \pm 0.012}\)\\\(\mathbf{[-0.168,-0.157]}\)} & \shortstack{\(\mathbf{-0.173 \pm 0.010}\)\\\(\mathbf{[-0.178,-0.168]}\)} & \shortstack{\(-0.140 \pm 0.009\)\\\([-0.144,-0.136]\)} & \shortstack{\(-0.165 \pm 0.005\)\\\([-0.168,-0.163]\)} \\
& & Cost & \shortstack{\(\mathbf{2.73 \pm 1.93}\)\\\(\mathbf{[2.05,3.76]}\)} & \shortstack{\(\mathbf{1.36 \pm 0.55}\)\\\(\mathbf{[1.11,1.63]}\)} & \shortstack{\(\mathbf{2.48 \pm 0.81}\)\\\(\mathbf{[2.09,2.85]}\)} & \shortstack{\(6.87 \pm 2.06\)\\\([5.89,7.88]\)} & \shortstack{\(10.54 \pm 0.62\)\\\([10.25,10.86]\)} \\
 & \multirow{2}{*}{Premium}
& Rew. & \shortstack{\(\mathbf{-0.211 \pm 0.012}\)\\\(\mathbf{[-0.217,-0.205]}\)} & \shortstack{\(\mathbf{-0.144 \pm 0.006}\)\\\(\mathbf{[-0.147,-0.141]}\)} & \shortstack{\(\mathbf{-0.154 \pm 0.005}\)\\\(\mathbf{[-0.157,-0.152]}\)} & \shortstack{\(\mathbf{-0.124 \pm 0.006}\)\\\(\mathbf{[-0.127,-0.121]}\)} & \shortstack{\(\mathbf{-0.141 \pm 0.004}\)\\\(\mathbf{[-0.144,-0.139]}\)} \\
& & Cost & \shortstack{\(\mathbf{4.88 \pm 2.12}\)\\\(\mathbf{[3.87,5.92]}\)} & \shortstack{\(\mathbf{2.04 \pm 0.68}\)\\\(\mathbf{[1.70,2.38]}\)} & \shortstack{\(\mathbf{2.74 \pm 1.04}\)\\\(\mathbf{[2.23,3.22]}\)} & \shortstack{\(\mathbf{3.93 \pm 1.20}\)\\\(\mathbf{[3.36,4.51]}\)} & \shortstack{\(\mathbf{4.83 \pm 1.32}\)\\\(\mathbf{[4.13,5.43]}\)} \\
\midrule
\multirow{6}{*}{Single VHVH}
 & \multirow{2}{*}{Bus}
& Rew. & \shortstack{\(\mathbf{-8.847 \pm 0.255}\)\\\(\mathbf{[-8.980,-8.727]}\)} & \shortstack{\(\mathbf{-8.679 \pm 0.168}\)\\\(\mathbf{[-8.761,-8.594]}\)} & \shortstack{\(\mathbf{-8.717 \pm 0.141}\)\\\(\mathbf{[-8.789,-8.651]}\)} & \shortstack{\(\mathbf{-8.428 \pm 0.071}\)\\\(\mathbf{[-8.462,-8.394]}\)} & \shortstack{\(\mathbf{-8.542 \pm 0.106}\)\\\(\mathbf{[-8.595,-8.488]}\)} \\
& & Cost & \shortstack{\(\mathbf{2.95 \pm 1.48}\)\\\(\mathbf{[2.26,3.70]}\)} & \shortstack{\(\mathbf{3.11 \pm 1.07}\)\\\(\mathbf{[2.58,3.64]}\)} & \shortstack{\(\mathbf{3.85 \pm 0.80}\)\\\(\mathbf{[3.48,4.26]}\)} & \shortstack{\(\mathbf{3.35 \pm 0.67}\)\\\(\mathbf{[3.03,3.67]}\)} & \shortstack{\(\mathbf{3.54 \pm 0.78}\)\\\(\mathbf{[3.17,3.92]}\)} \\
 & \multirow{2}{*}{Convoy}
& Rew. & \shortstack{\(-73.796 \pm 9.963\)\\\([-78.314,-68.208]\)} & \shortstack{\(-30.291 \pm 3.850\)\\\([-32.178,-28.389]\)} & \shortstack{\(-13.762 \pm 0.581\)\\\([-14.038,-13.482]\)} & \shortstack{\(-17.361 \pm 2.338\)\\\([-18.535,-16.265]\)} & \shortstack{\(-15.015 \pm 1.140\)\\\([-15.610,-14.514]\)} \\
& & Cost & \shortstack{\(28.09 \pm 22.24\)\\\([18.14,40.15]\)} & \shortstack{\(8.08 \pm 9.85\)\\\([3.96,13.36]\)} & \shortstack{\(21.92 \pm 4.47\)\\\([19.90,24.09]\)} & \shortstack{\(20.78 \pm 4.49\)\\\([18.65,22.96]\)} & \shortstack{\(72.71 \pm 11.90\)\\\([66.88,78.77]\)} \\
 & \multirow{2}{*}{Premium}
& Rew. & \shortstack{\(-14.990 \pm 17.560\)\\\([-25.189,-9.147]\)} & \shortstack{\(\mathbf{-9.032 \pm 0.200}\)\\\(\mathbf{[-9.127,-8.933]}\)} & \shortstack{\(\mathbf{-8.918 \pm 0.254}\)\\\(\mathbf{[-9.058,-8.810]}\)} & \shortstack{\(\mathbf{-8.613 \pm 0.082}\)\\\(\mathbf{[-8.648,-8.569]}\)} & \shortstack{\(-8.570 \pm 0.073\)\\\([-8.604,-8.532]\)} \\
& & Cost & \shortstack{\(21.71 \pm 44.48\)\\\([8.76,45.25]\)} & \shortstack{\(\mathbf{7.98 \pm 3.29}\)\\\(\mathbf{[6.48,9.70]}\)} & \shortstack{\(\mathbf{14.33 \pm 1.34}\)\\\(\mathbf{[13.62,14.98]}\)} & \shortstack{\(\mathbf{14.51 \pm 2.10}\)\\\(\mathbf{[13.49,15.50]}\)} & \shortstack{\(24.27 \pm 1.71\)\\\([23.42,25.12]\)} \\
\bottomrule
\end{tabular}%
}
\end{table*}

\begin{table}[htb!]
\centering
\caption{\textbf{Label-in-Obs, CityLearn.} Mean \(\pm\) std, with 95\% bootstrap CI, over seeds. \textbf{Bold} marks entries where the method is feasible (cost within the limit in Table~\ref{tab:evaluation_cost_limits}); unbolded entries exceed the limit.}
\label{tab:ablation_raw_citylearn_labelinobs}
\renewcommand{\arraystretch}{1.5}
\resizebox{\textwidth}{!}{%
\begin{tabular}{lrrrrrr}
\toprule
\textbf{Scenario} & \textbf{Metric} & \textbf{CPO} & \textbf{CUP} & \textbf{IPO} & \textbf{PPOLag} & \textbf{PPOLag (GBT)} \\
\midrule
\multirow{2}{*}{Arb.\ vs Buffer} & Rew. & \shortstack{\(-7.75 \pm 0.37\)\\\([-7.93,-7.58]\)} & \shortstack{\(-8.45 \pm 0.33\)\\\([-8.57,-8.29]\)} & \shortstack{\(-8.61 \pm 0.06\)\\\([-8.64,-8.58]\)} & \shortstack{\(-8.11 \pm 0.25\)\\\([-8.22,-7.98]\)} & \shortstack{\(-8.84 \pm 0.13\)\\\([-8.91,-8.78]\)} \\
 & Cost & \shortstack{\(7.33 \pm 3.28\)\\\([5.76,9.00]\)} & \shortstack{\(7.26 \pm 9.64\)\\\([2.94,12.25]\)} & \shortstack{\(6.08 \pm 5.77\)\\\([3.37,9.00]\)} & \shortstack{\(12.91 \pm 5.69\)\\\([10.26,15.85]\)} & \shortstack{\(9.61 \pm 3.60\)\\\([7.75,11.29]\)} \\
\multirow{2}{*}{Demand Response} & Rew. & \shortstack{\(-7.63 \pm 0.38\)\\\([-7.81,-7.45]\)} & \shortstack{\(-7.80 \pm 0.86\)\\\([-8.30,-7.44]\)} & \shortstack{\(-8.33 \pm 0.33\)\\\([-8.48,-8.15]\)} & \shortstack{\(-8.42 \pm 0.25\)\\\([-8.51,-8.28]\)} & \shortstack{\(-8.92 \pm 0.10\)\\\([-8.98,-8.87]\)} \\
 & Cost & \shortstack{\(8.78 \pm 3.63\)\\\([6.88,10.57]\)} & \shortstack{\(9.11 \pm 8.40\)\\\([5.44,13.70]\)} & \shortstack{\(7.03 \pm 5.04\)\\\([4.71,9.76]\)} & \shortstack{\(28.29 \pm 0.15\)\\\([28.22,28.36]\)} & \shortstack{\(10.46 \pm 3.05\)\\\([8.82,12.04]\)} \\
\multirow{2}{*}{Solar Ramp} & Rew. & \shortstack{\(-7.77 \pm 0.47\)\\\([-8.01,-7.55]\)} & \shortstack{\(-8.07 \pm 0.41\)\\\([-8.27,-7.87]\)} & \shortstack{\(-8.55 \pm 0.38\)\\\([-8.72,-8.35]\)} & \shortstack{\(-8.40 \pm 0.19\)\\\([-8.49,-8.30]\)} & \shortstack{\(-8.98 \pm 0.10\)\\\([-9.02,-8.93]\)} \\
 & Cost & \shortstack{\(5.75 \pm 3.22\)\\\([4.23,7.39]\)} & \shortstack{\(7.66 \pm 3.97\)\\\([5.70,9.60]\)} & \shortstack{\(9.04 \pm 3.77\)\\\([7.11,10.69]\)} & \shortstack{\(35.83 \pm 24.31\)\\\([24.68,47.94]\)} & \shortstack{\(10.96 \pm 2.76\)\\\([9.59,12.37]\)} \\
\bottomrule
\end{tabular}%
}
\end{table}
\begin{table*}[hbt!]
\centering
\caption{\textbf{Label-in-Obs, Safety MiniGrid.} Mean \(\pm\) std, with 95\% bootstrap CI, final
training steps. \textbf{Bold} marks entries where the method is feasible (cost within the limit in Table~\ref{tab:evaluation_cost_limits}); unbolded entries exceed the limit.}
\label{tab:ablation_raw_minigrid_labelinobs}
\renewcommand{\arraystretch}{1.5}
\resizebox{\textwidth}{!}{%
\begin{tabular}{llccccc}
\toprule
\textbf{Environment} & \textbf{Metric} & \textbf{CUP} & \textbf{CPO} & \textbf{IPO} & \textbf{PPOLag} & \textbf{PPOLag (GBT)} \\
\midrule
\multirow{2}{*}{DynamicCrossing}
& Reward & \shortstack{\(0.15 \pm 0.10\)\\\([0.10,0.20]\)} & \shortstack{\(0.14 \pm 0.13\)\\\([0.07,0.20]\)} & \shortstack{\(0.29 \pm 0.30\)\\\([0.17,0.46]\)} & \shortstack{\(0.16 \pm 0.35\)\\\([0.04,0.35]\)} & \shortstack{\(\mathbf{1.48 \pm 0.17}\)\\\(\mathbf{[1.39,1.53]}\)} \\
& Cost & \shortstack{\(6.39 \pm 23.96\)\\\([0.11,18.80]\)} & \shortstack{\(0.12 \pm 0.08\)\\\([0.09,0.17]\)} & \shortstack{\(4.61 \pm 10.01\)\\\([1.13,10.16]\)} & \shortstack{\(26.80 \pm 79.85\)\\\([2.03,71.43]\)} & \shortstack{\(\mathbf{0.01 \pm 0.01}\)\\\(\mathbf{[0.00,0.01]}\)} \\
\multirow{2}{*}{FragileCrossing}
& Reward & \shortstack{\(\mathbf{-0.02 \pm 0.13}\)\\\(\mathbf{[-0.08,0.04]}\)} & \shortstack{\(0.27 \pm 0.08\)\\\([0.23,0.31]\)} & \shortstack{\(\mathbf{0.19 \pm 0.22}\)\\\(\mathbf{[0.08,0.30]}\)} & \shortstack{\(\mathbf{0.10 \pm 0.18}\)\\\(\mathbf{[0.02,0.19]}\)} & \shortstack{\(0.90 \pm 0.20\)\\\([0.80,0.99]\)} \\
& Cost & \shortstack{\(\mathbf{0.00 \pm 0.00}\)\\\(\mathbf{[0.00,0.00]}\)} & \shortstack{\(0.15 \pm 0.29\)\\\([0.04,0.32]\)} & \shortstack{\(\mathbf{0.04 \pm 0.08}\)\\\(\mathbf{[0.00,0.08]}\)} & \shortstack{\(\mathbf{0.02 \pm 0.07}\)\\\(\mathbf{[0.00,0.06]}\)} & \shortstack{\(0.20 \pm 0.50\)\\\([0.03,0.46]\)} \\
\multirow{2}{*}{Corner}
& Reward & \shortstack{\(0.06 \pm 0.21\)\\\([0.00,0.17]\)} & \shortstack{\(0.73 \pm 0.30\)\\\([0.57,0.87]\)} & \shortstack{\(0.00 \pm 0.01\)\\\([0.00,0.01]\)} & \shortstack{\(0.00 \pm 0.00\)\\\([0.00,0.00]\)} & \shortstack{\(\mathbf{1.00 \pm 0.00}\)\\\(\mathbf{[1.00,1.00]}\)} \\
& Cost & \shortstack{\(0.37 \pm 1.15\)\\\([0.00,1.04]\)} & \shortstack{\(0.86 \pm 0.98\)\\\([0.41,1.39]\)} & \shortstack{\(25.23 \pm 72.59\)\\\([1.56,64.45]\)} & \shortstack{\(3.91 \pm 10.84\)\\\([0.17,10.42]\)} & \shortstack{\(\mathbf{0.00 \pm 0.00}\)\\\(\mathbf{[0.00,0.00]}\)} \\
\bottomrule
\end{tabular}%
}
\end{table*}

\subsubsection{Masked}
Tables~\ref{tab:ablation_raw_sumo_masked}--\ref{tab:ablation_raw_minigrid_masked} report the same three domains with the label used to mask the losses: reward-labelled samples enter only the reward update, cost-labelled samples only the cost update.

\begin{table*}[hbt!]
\centering
\caption{\textbf{Masked, SUMO-RL.} Mean \(\pm\) std, with 95\% bootstrap CI, over seeds.
Reward-labelled samples get only the reward update, cost-labelled samples get only the cost
update, no conflict-aware score. Reward is in units of \(10^{3}\); cost is unscaled. \textbf{Bold} marks entries where the method is feasible (cost within the limit in Table~\ref{tab:evaluation_cost_limits}); unbolded entries exceed the limit.}
\label{tab:ablation_raw_sumo_masked}
\renewcommand{\arraystretch}{1.5}
\resizebox{\textwidth}{!}{%
\begin{tabular}{lllrrrrr}
\toprule
\textbf{Environment} & \textbf{Scenario} & \textbf{Metric} & \textbf{CPO} & \textbf{CUP} & \textbf{IPO} & \textbf{PPOLag} & \textbf{PPOLag (GBT)} \\
\midrule
\multirow{6}{*}{Arterial 4x4}
 & \multirow{2}{*}{Bus}
& Rew. & \shortstack{\(\mathbf{-1.432 \pm 0.292}\)\\\(\mathbf{[-1.577,-1.291]}\)} & \shortstack{\(\mathbf{-1.264 \pm 0.165}\)\\\(\mathbf{[-1.350,-1.189]}\)} & \shortstack{\(\mathbf{-1.308 \pm 0.286}\)\\\(\mathbf{[-1.472,-1.181]}\)} & \shortstack{\(\mathbf{-1.131 \pm 0.222}\)\\\(\mathbf{[-1.243,-1.038]}\)} & \shortstack{\(-1.196 \pm 0.147\)\\\([-1.272,-1.128]\)} \\
& & Cost & \shortstack{\(\mathbf{3.69 \pm 3.77}\)\\\(\mathbf{[2.10,5.81]}\)} & \shortstack{\(\mathbf{2.34 \pm 1.96}\)\\\(\mathbf{[1.47,3.41]}\)} & \shortstack{\(\mathbf{2.18 \pm 2.07}\)\\\(\mathbf{[1.30,3.34]}\)} & \shortstack{\(\mathbf{6.00 \pm 4.05}\)\\\(\mathbf{[4.16,8.04]}\)} & \shortstack{\(19.07 \pm 8.05\)\\\([15.16,22.86]\)} \\
 & \multirow{2}{*}{Convoy}
& Rew. & \shortstack{\(\mathbf{-1.402 \pm 0.243}\)\\\(\mathbf{[-1.522,-1.286]}\)} & \shortstack{\(-1.411 \pm 0.216\)\\\([-1.515,-1.312]\)} & \shortstack{\(\mathbf{-1.491 \pm 0.298}\)\\\(\mathbf{[-1.663,-1.354]}\)} & \shortstack{\(-1.288 \pm 0.161\)\\\([-1.366,-1.212]\)} & \shortstack{\(-1.219 \pm 0.178\)\\\([-1.312,-1.138]\)} \\
& & Cost & \shortstack{\(\mathbf{4.63 \pm 3.93}\)\\\(\mathbf{[2.93,6.80]}\)} & \shortstack{\(6.71 \pm 5.22\)\\\([4.20,9.38]\)} & \shortstack{\(\mathbf{4.47 \pm 4.50}\)\\\(\mathbf{[2.37,6.86]}\)} & \shortstack{\(8.00 \pm 3.81\)\\\([6.11,9.99]\)} & \shortstack{\(15.72 \pm 3.00\)\\\([14.15,17.14]\)} \\
 & \multirow{2}{*}{Premium}
& Rew. & \shortstack{\(\mathbf{-1.431 \pm 0.237}\)\\\(\mathbf{[-1.550,-1.310]}\)} & \shortstack{\(\mathbf{-1.356 \pm 0.245}\)\\\(\mathbf{[-1.483,-1.245]}\)} & \shortstack{\(\mathbf{-1.223 \pm 0.155}\)\\\(\mathbf{[-1.302,-1.150]}\)} & \shortstack{\(-1.157 \pm 0.140\)\\\([-1.227,-1.091]\)} & \shortstack{\(-1.199 \pm 0.141\)\\\([-1.272,-1.133]\)} \\
& & Cost & \shortstack{\(\mathbf{5.43 \pm 2.24}\)\\\(\mathbf{[4.37,6.53]}\)} & \shortstack{\(\mathbf{7.26 \pm 5.44}\)\\\(\mathbf{[4.80,10.07]}\)} & \shortstack{\(\mathbf{5.34 \pm 2.03}\)\\\(\mathbf{[4.34,6.34]}\)} & \shortstack{\(21.66 \pm 14.76\)\\\([15.16,29.69]\)} & \shortstack{\(42.71 \pm 10.15\)\\\([37.85,47.78]\)} \\
\midrule
\multirow{6}{*}{Grid 4x4}
 & \multirow{2}{*}{Bus}
& Rew. & \shortstack{\(\mathbf{-0.195 \pm 0.008}\)\\\(\mathbf{[-0.199,-0.191]}\)} & \shortstack{\(\mathbf{-0.139 \pm 0.005}\)\\\(\mathbf{[-0.141,-0.136]}\)} & \shortstack{\(\mathbf{-0.147 \pm 0.005}\)\\\(\mathbf{[-0.150,-0.145]}\)} & \shortstack{\(\mathbf{-0.117 \pm 0.003}\)\\\(\mathbf{[-0.118,-0.115]}\)} & \shortstack{\(\mathbf{-0.135 \pm 0.003}\)\\\(\mathbf{[-0.136,-0.133]}\)} \\
& & Cost & \shortstack{\(\mathbf{0.20 \pm 0.11}\)\\\(\mathbf{[0.15,0.26]}\)} & \shortstack{\(\mathbf{0.15 \pm 0.19}\)\\\(\mathbf{[0.07,0.26]}\)} & \shortstack{\(\mathbf{0.24 \pm 0.33}\)\\\(\mathbf{[0.10,0.42]}\)} & \shortstack{\(\mathbf{0.19 \pm 0.19}\)\\\(\mathbf{[0.11,0.31]}\)} & \shortstack{\(\mathbf{0.35 \pm 0.16}\)\\\(\mathbf{[0.27,0.43]}\)} \\
 & \multirow{2}{*}{Convoy}
& Rew. & \shortstack{\(\mathbf{-0.457 \pm 0.529}\)\\\(\mathbf{[-0.756,-0.252]}\)} & \shortstack{\(\mathbf{-0.172 \pm 0.010}\)\\\(\mathbf{[-0.177,-0.167]}\)} & \shortstack{\(\mathbf{-0.179 \pm 0.010}\)\\\(\mathbf{[-0.183,-0.174]}\)} & \shortstack{\(\mathbf{-0.149 \pm 0.007}\)\\\(\mathbf{[-0.152,-0.146]}\)} & \shortstack{\(-0.172 \pm 0.004\)\\\([-0.174,-0.170]\)} \\
& & Cost & \shortstack{\(\mathbf{3.66 \pm 3.92}\)\\\(\mathbf{[2.27,5.91]}\)} & \shortstack{\(\mathbf{1.16 \pm 0.52}\)\\\(\mathbf{[0.91,1.42]}\)} & \shortstack{\(\mathbf{2.14 \pm 0.70}\)\\\(\mathbf{[1.79,2.48]}\)} & \shortstack{\(\mathbf{3.78 \pm 0.77}\)\\\(\mathbf{[3.41,4.19]}\)} & \shortstack{\(13.25 \pm 0.78\)\\\([12.90,13.65]\)} \\
 & \multirow{2}{*}{Premium}
& Rew. & \shortstack{\(\mathbf{-0.206 \pm 0.013}\)\\\(\mathbf{[-0.212,-0.199]}\)} & \shortstack{\(\mathbf{-0.145 \pm 0.009}\)\\\(\mathbf{[-0.150,-0.141]}\)} & \shortstack{\(\mathbf{-0.154 \pm 0.007}\)\\\(\mathbf{[-0.157,-0.151]}\)} & \shortstack{\(\mathbf{-0.119 \pm 0.004}\)\\\(\mathbf{[-0.122,-0.117]}\)} & \shortstack{\(\mathbf{-0.139 \pm 0.003}\)\\\(\mathbf{[-0.140,-0.137]}\)} \\
& & Cost & \shortstack{\(\mathbf{6.01 \pm 1.78}\)\\\(\mathbf{[5.15,6.84]}\)} & \shortstack{\(\mathbf{1.60 \pm 0.52}\)\\\(\mathbf{[1.34,1.84]}\)} & \shortstack{\(\mathbf{3.35 \pm 1.14}\)\\\(\mathbf{[2.82,3.92]}\)} & \shortstack{\(\mathbf{2.84 \pm 1.26}\)\\\(\mathbf{[2.24,3.48]}\)} & \shortstack{\(\mathbf{5.86 \pm 1.56}\)\\\(\mathbf{[5.06,6.56]}\)} \\
\midrule
\multirow{6}{*}{Single VHVH}
 & \multirow{2}{*}{Bus}
& Rew. & \shortstack{\(\mathbf{-8.754 \pm 0.227}\)\\\(\mathbf{[-8.860,-8.638]}\)} & \shortstack{\(\mathbf{-8.704 \pm 0.203}\)\\\(\mathbf{[-8.801,-8.610]}\)} & \shortstack{\(\mathbf{-8.702 \pm 0.140}\)\\\(\mathbf{[-8.769,-8.634]}\)} & \shortstack{\(\mathbf{-8.457 \pm 0.085}\)\\\(\mathbf{[-8.499,-8.415]}\)} & \shortstack{\(\mathbf{-8.516 \pm 0.053}\)\\\(\mathbf{[-8.542,-8.489]}\)} \\
& & Cost & \shortstack{\(\mathbf{3.03 \pm 1.12}\)\\\(\mathbf{[2.55,3.62]}\)} & \shortstack{\(\mathbf{3.00 \pm 1.02}\)\\\(\mathbf{[2.48,3.45]}\)} & \shortstack{\(\mathbf{4.62 \pm 0.96}\)\\\(\mathbf{[4.19,5.12]}\)} & \shortstack{\(\mathbf{2.92 \pm 0.89}\)\\\(\mathbf{[2.50,3.37]}\)} & \shortstack{\(\mathbf{4.20 \pm 0.79}\)\\\(\mathbf{[3.83,4.59]}\)} \\
 & \multirow{2}{*}{Convoy}
& Rew. & \shortstack{\(-72.898 \pm 13.303\)\\\([-78.773,-66.007]\)} & \shortstack{\(-32.999 \pm 3.678\)\\\([-34.795,-31.246]\)} & \shortstack{\(-27.005 \pm 2.706\)\\\([-28.338,-25.603]\)} & \shortstack{\(-28.032 \pm 4.621\)\\\([-30.199,-25.745]\)} & \shortstack{\(-16.880 \pm 2.612\)\\\([-18.187,-15.591]\)} \\
& & Cost & \shortstack{\(21.92 \pm 13.75\)\\\([15.19,28.76]\)} & \shortstack{\(15.95 \pm 16.10\)\\\([8.62,24.34]\)} & \shortstack{\(10.23 \pm 8.09\)\\\([7.44,14.78]\)} & \shortstack{\(12.16 \pm 3.75\)\\\([10.32,14.08]\)} & \shortstack{\(37.11 \pm 7.33\)\\\([33.64,40.56]\)} \\
 & \multirow{2}{*}{Premium}
& Rew. & \shortstack{\(\mathbf{-8.989 \pm 0.167}\)\\\(\mathbf{[-9.068,-8.910]}\)} & \shortstack{\(\mathbf{-9.005 \pm 0.224}\)\\\(\mathbf{[-9.121,-8.905]}\)} & \shortstack{\(\mathbf{-8.989 \pm 0.396}\)\\\(\mathbf{[-9.208,-8.824]}\)} & \shortstack{\(\mathbf{-8.586 \pm 0.084}\)\\\(\mathbf{[-8.626,-8.545]}\)} & \shortstack{\(-8.609 \pm 0.088\)\\\([-8.652,-8.565]\)} \\
& & Cost & \shortstack{\(\mathbf{11.36 \pm 3.49}\)\\\(\mathbf{[9.65,13.13]}\)} & \shortstack{\(\mathbf{9.85 \pm 2.79}\)\\\(\mathbf{[8.59,11.15]}\)} & \shortstack{\(\mathbf{14.28 \pm 1.00}\)\\\(\mathbf{[13.81,14.76]}\)} & \shortstack{\(\mathbf{14.74 \pm 2.25}\)\\\(\mathbf{[13.67,15.92]}\)} & \shortstack{\(25.69 \pm 2.26\)\\\([24.62,26.82]\)} \\
\bottomrule
\end{tabular}%
}
\end{table*}

\begin{table}[htb!]
\centering
\caption{\textbf{Masked, CityLearn.} Mean \(\pm\) std, with 95\% bootstrap CI, over seeds. \textbf{Bold} marks entries where the method is feasible (cost within the limit in Table~\ref{tab:evaluation_cost_limits}); unbolded entries exceed the limit.}
\label{tab:ablation_raw_citylearn_masked}
\renewcommand{\arraystretch}{1.5}
\resizebox{\textwidth}{!}{%
\begin{tabular}{lrrrrrr}
\toprule
\textbf{Scenario} & \textbf{Metric} & \textbf{CPO} & \textbf{CUP} & \textbf{IPO} & \textbf{PPOLag} & \textbf{PPOLag (GBT)} \\
\midrule
\multirow{2}{*}{Arb.\ vs Buffer} & Rew. & \shortstack{\(-29.75 \pm 7.97\)\\\([-33.72,-25.81]\)} & \shortstack{\(-7.23 \pm 0.29\)\\\([-7.39,-7.11]\)} & \shortstack{\(-11.24 \pm 0.13\)\\\([-11.30,-11.18]\)} & \shortstack{\(-8.19 \pm 0.43\)\\\([-8.40,-7.97]\)} & \shortstack{\(-8.79 \pm 0.04\)\\\([-8.80,-8.77]\)} \\
 & Cost & \shortstack{\(9.92 \pm 2.71\)\\\([8.58,11.21]\)} & \shortstack{\(206.65 \pm 16.22\)\\\([198.48,214.25]\)} & \shortstack{\(61.00 \pm 5.57\)\\\([58.47,63.99]\)} & \shortstack{\(238.68 \pm 18.58\)\\\([229.94,248.23]\)} & \shortstack{\(211.04 \pm 9.11\)\\\([206.01,215.09]\)} \\
\multirow{2}{*}{Demand Response} & Rew. & \shortstack{\(-7.55 \pm 0.35\)\\\([-7.73,-7.38]\)} & \shortstack{\(-7.77 \pm 1.00\)\\\([-8.30,-7.33]\)} & \shortstack{\(-10.19 \pm 0.80\)\\\([-10.58,-9.80]\)} & \shortstack{\(-8.48 \pm 0.11\)\\\([-8.51,-8.41]\)} & \shortstack{\(-8.84 \pm 0.09\)\\\([-8.88,-8.80]\)} \\
 & Cost & \shortstack{\(6.07 \pm 4.07\)\\\([4.41,8.16]\)} & \shortstack{\(10.55 \pm 8.60\)\\\([6.78,15.04]\)} & \shortstack{\(5.31 \pm 1.26\)\\\([4.74,5.93]\)} & \shortstack{\(28.24 \pm 0.16\)\\\([28.16,28.32]\)} & \shortstack{\(12.50 \pm 3.70\)\\\([10.84,14.39]\)} \\
\multirow{2}{*}{Solar Ramp} & Rew. & \shortstack{\(\mathbf{-7.65 \pm 0.54}\)\\\(\mathbf{[-7.96,-7.42]}\)} & \shortstack{\(\mathbf{-8.09 \pm 0.67}\)\\\(\mathbf{[-8.43,-7.78]}\)} & \shortstack{\(\mathbf{-8.57 \pm 0.54}\)\\\(\mathbf{[-8.82,-8.29]}\)} & \shortstack{\(-8.77 \pm 0.44\)\\\([-9.02,-8.59]\)} & \shortstack{\(-9.11 \pm 0.07\)\\\([-9.14,-9.08]\)} \\
 & Cost & \shortstack{\(\mathbf{2.85 \pm 1.52}\)\\\(\mathbf{[2.09,3.56]}\)} & \shortstack{\(\mathbf{4.62 \pm 3.13}\)\\\(\mathbf{[3.16,6.09]}\)} & \shortstack{\(\mathbf{3.58 \pm 1.65}\)\\\(\mathbf{[2.77,4.34]}\)} & \shortstack{\(5.20 \pm 3.90\)\\\([3.39,7.20]\)} & \shortstack{\(10.76 \pm 1.01\)\\\([10.25,11.25]\)} \\
\bottomrule
\end{tabular}%
}
\end{table}

\begin{table*}[hbt!]
\centering
\caption{\textbf{Masked, Safety MiniGrid.} Mean \(\pm\) std, with 95\% bootstrap CI, final
training steps. \textbf{Bold} marks entries where the method is feasible (cost within the limit in Table~\ref{tab:evaluation_cost_limits}); unbolded entries exceed the limit.}
\label{tab:ablation_raw_minigrid_masked}
\renewcommand{\arraystretch}{1.5}
\resizebox{\textwidth}{!}{%
\begin{tabular}{llccccc}
\toprule
\textbf{Environment} & \textbf{Metric} & \textbf{CUP} & \textbf{CPO} & \textbf{IPO} & \textbf{PPOLag} & \textbf{PPOLag (GBT)} \\
\midrule
\multirow{2}{*}{DynamicCrossing}
& Reward & \shortstack{\(0.20 \pm 0.20\)\\\([0.11,0.31]\)} & \shortstack{\(0.19 \pm 0.13\)\\\([0.14,0.26]\)} & \shortstack{\(0.39 \pm 0.32\)\\\([0.25,0.56]\)} & \shortstack{\(0.07 \pm 0.14\)\\\([0.01,0.14]\)} & \shortstack{\(1.20 \pm 0.26\)\\\([1.08,1.32]\)} \\
& Cost & \shortstack{\(45.64 \pm 109.96\)\\\([2.19,107.89]\)} & \shortstack{\(17.87 \pm 20.52\)\\\([8.68,29.17]\)} & \shortstack{\(20.82 \pm 35.60\)\\\([6.25,40.96]\)} & \shortstack{\(31.98 \pm 73.13\)\\\([2.70,73.67]\)} & \shortstack{\(0.36 \pm 0.31\)\\\([0.22,0.52]\)} \\
\multirow{2}{*}{FragileCrossing}
& Reward & \shortstack{\(\mathbf{0.12 \pm 0.19}\)\\\(\mathbf{[0.03,0.22]}\)} & \shortstack{\(\mathbf{0.26 \pm 0.13}\)\\\(\mathbf{[0.20,0.32]}\)} & \shortstack{\(0.18 \pm 0.23\)\\\([0.07,0.30]\)} & \shortstack{\(0.22 \pm 0.20\)\\\([0.12,0.32]\)} & \shortstack{\(\mathbf{0.64 \pm 0.01}\)\\\(\mathbf{[0.64,0.65]}\)} \\
& Cost & \shortstack{\(\mathbf{0.05 \pm 0.18}\)\\\(\mathbf{[0.00,0.14]}\)} & \shortstack{\(\mathbf{0.09 \pm 0.09}\)\\\(\mathbf{[0.05,0.14]}\)} & \shortstack{\(0.26 \pm 0.63\)\\\([0.01,0.61]\)} & \shortstack{\(0.13 \pm 0.38\)\\\([0.00,0.35]\)} & \shortstack{\(\mathbf{0.01 \pm 0.02}\)\\\(\mathbf{[0.00,0.02]}\)} \\
\multirow{2}{*}{Corner}
& Reward & \shortstack{\(0.04 \pm 0.14\)\\\([0.00,0.11]\)} & \shortstack{\(\mathbf{0.95 \pm 0.10}\)\\\(\mathbf{[0.89,0.99]}\)} & \shortstack{\(0.73 \pm 0.28\)\\\([0.58,0.85]\)} & \shortstack{\(0.02 \pm 0.08\)\\\([0.00,0.06]\)} & \shortstack{\(\mathbf{1.00 \pm 0.00}\)\\\(\mathbf{[1.00,1.00]}\)} \\
& Cost & \shortstack{\(4.25 \pm 16.48\)\\\([0.00,12.76]\)} & \shortstack{\(\mathbf{0.10 \pm 0.39}\)\\\(\mathbf{[0.00,0.31]}\)} & \shortstack{\(4.05 \pm 4.19\)\\\([2.25,6.23]\)} & \shortstack{\(5.19 \pm 13.63\)\\\([0.00,12.22]\)} & \shortstack{\(\mathbf{0.00 \pm 0.00}\)\\\(\mathbf{[0.00,0.00]}\)} \\
\bottomrule
\end{tabular}%
}
\end{table*}

\clearpage
\input{checklist.tex}

\end{document}

%% file: checklist.tex
\section*{NeurIPS Paper Checklist}

\begin{enumerate}

\item {\bf Claims}
\item[] Question: Do the main claims made in the abstract and introduction accurately reflect the paper's contributions and scope?
\item[] Answer: \answerYes{}
\item[] Justification: The abstract and Section~\ref{sec:introduction} state the scope as guided state splitting for localized signal conflict and summarize the constrained-RL and offline-RL evaluation settings. The claims are matched to the theoretical properties in Appendix~\ref{app:split_proofs}, the empirical results in Section~\ref{sec:results}, and the stated limitations in the Limitations section.

\item {\bf Limitations}
\item[] Question: Does the paper discuss the limitations of the work performed by the authors?
\item[] Answer: \answerYes{}
\item[] Justification: The paper includes a dedicated Limitations section. It discusses the dependence on useful guidance labels, the reduced benefit when conflicts are global rather than localized, and the current restriction to GBT policies and structured observations.

\item {\bf Theory assumptions and proofs}
\item[] Question: For each theoretical result, does the paper provide the full set of assumptions and a complete (and correct) proof?
\item[] Answer: \answerYes{}
\item[] Justification: Appendix~\ref{app:split_proofs} states each proposition with its assumptions and provides complete proofs. The results are algebraic properties of the Split-RL scoring and leaf aggregation rules and are cross-referenced from the main method section.

\item {\bf Experimental result reproducibility}
\item[] Question: Does the paper fully disclose all the information needed to reproduce the main experimental results of the paper to the extent that it affects the main claims and/or conclusions of the paper (regardless of whether the code and data are provided or not)?
\item[] Answer: \answerYes{}
\item[] Justification: Sections~\ref{sec:experiments}--\ref{sec:results} describe the experimental domains, baselines, evaluation metrics, and aggregation over seeds. Appendices~\ref{app:minigrid_envs}--\ref{app:hyperparameters} provide environment definitions, label and cost rules, dataset construction details, training budgets, hyperparameters, and seed counts.
\item[] Guidelines:
\begin{itemize}
\item The answer \answerNA{} means that the paper does not include experiments.
\item If the paper includes experiments, a \answerNo{} answer to this question will not be perceived well by the reviewers: Making the paper reproducible is important, regardless of whether the code and data are provided or not.
\item If the contribution is a dataset and\slash or model, the authors should describe the steps taken to make their results reproducible or verifiable.
\item Depending on the contribution, reproducibility can be accomplished in various ways. For example, if the contribution is a novel architecture, describing the architecture fully might suffice, or if the contribution is a specific model and empirical evaluation, it may be necessary to either make it possible for others to replicate the model with the same dataset, or provide access to the model. In general.
\item Releasing code and data is often one good way to accomplish this, but reproducibility can also be provided via detailed instructions for how to replicate the results, access to a hosted model (e.g., in the case of a large language model), releasing of a model checkpoint, or other means that are appropriate to the research performed.
\item While NeurIPS does not require releasing code, the conference does require all submissions to provide some reasonable avenue for reproducibility, which may depend on the nature of the contribution. For example
\begin{enumerate}
\item If the contribution is primarily a new algorithm, the paper should make it clear how to reproduce that algorithm.
\item If the contribution is primarily a new model architecture, the paper should describe the architecture clearly and fully.
\item If the contribution is a new model (e.g., a large language model), then there should either be a way to access this model for reproducing the results or a way to reproduce the model (e.g., with an open-source dataset or instructions for how to construct the dataset).
\item We recognize that reproducibility may be tricky in some cases, in which case authors are welcome to describe the particular way they provide for reproducibility. In the case of closed-source models, it may be that access to the model is limited in some way (e.g., to registered users), but it should be possible for other researchers to have some path to reproducing or verifying the results.
\end{enumerate}
\end{itemize}

\item {\bf Open access to data and code}
\item[] Question: Does the paper provide open access to the data and code, with sufficient instructions to faithfully reproduce the main experimental results, as described in supplemental material?
\item[] Answer: \answerYes{}
\item[] Justification:  The Split-RL code repository is found at \url{https://github.com/NVlabs/split_rl}. The code release provides the Split-RL implementation, experiment configurations, and instructions for reproducing the main constrained-RL and offline-RL experiments. For Atari offline RL, the release provides scripts that download the public expert-policy checkpoints from their original publicly available repositories and generate the rollout datasets locally; third-party checkpoint weights and generated Atari transition files are not redistributed.

\item {\bf Experimental setting/details}
\item[] Question: Does the paper specify all the training and test details (e.g., data splits, hyperparameters, how they were chosen, type of optimizer) necessary to understand the results?
\item[] Answer: \answerYes{}
\item[] Justification: Section~\ref{sec:experiments} describes the experiment design, baselines, metrics, and evaluation protocol. Appendix~\ref{app:hyperparameters} reports training steps, batch sizes, discount and GAE parameters, learning rates, tree depths, seeds, and how hyperparameters were selected.

\item {\bf Experiment statistical significance}
\item[] Question: Does the paper report error bars suitably and correctly defined or other appropriate information about the statistical significance of the experiments?
\item[] Answer: \answerYes{}
\item[] Justification: Tables report mean \(\pm\) standard deviation across seeds, with the source of variability stated in the captions, and 95\% percentile bootstrap confidence intervals computed over seeds with 2{,}000 resamples. Constrained-RL results are aggregated over 15 seeds. Atari results use one training run per game and dataset size, evaluated over 50 episodes (10 episodes on each of 5 evaluation seeds), and the main normalized summary reports variation across games after normalization.
\item[] Guidelines:
\begin{itemize}
\item The answer \answerNA{} means that the paper does not include experiments.
\item The authors should answer \answerYes{} if the results are accompanied by error bars, confidence intervals, or statistical significance tests, at least for the experiments that support the main claims of the paper.
\item The factors of variability that the error bars are capturing should be clearly stated (for example, train/test split, initialization, random drawing of some parameter, or overall run with given experimental conditions).
\item The method for calculating the error bars should be explained (closed form formula, call to a library function, bootstrap, etc.)
\item The assumptions made should be given (e.g., Normally distributed errors).
\item It should be clear whether the error bar is the standard deviation or the standard error of the mean.
\item It is OK to report 1-sigma error bars, but one should state it.
\item The authors should preferably report a 2-sigma error bar than state that they have a 96\% CI, if the hypothesis of Normality of errors is not verified.
\item For asymmetric distributions, the authors should be careful not to show in tables or plots symmetric error bars that would yield results that are out of range (e.g., negative error rates).
\item If error bars are reported in tables or plots, the authors should explain in the text how they were calculated and reference the corresponding figures or tables in the text.
\end{itemize}

\item {\bf Experiments compute resources}
\item[] Question: For each experiment, does the paper provide sufficient information on the computer resources (type of compute workers, memory, time of execution) needed to reproduce the experiments?
\item[] Answer: \answerYes{}
\item[] Justification: Experiments were run on an H100-class GPU cluster. Individual constrained-RL runs used the training budgets and 15-seed protocol reported in Appendix~\ref{app:hyperparameters}; MiniGrid and SUMO-RL runs took a few hours, while CityLearn runs took approximately 24 hours. Atari RAM offline runs used the same cluster with method-dependent budgets of 50K--500K gradient steps. Preliminary tuning used Bayesian search on a single seed before fixing hyperparameters across evaluation seeds.

\item {\bf Code of ethics}
\item[] Question: Does the research conducted in the paper conform, in every respect, with the NeurIPS Code of Ethics \url{https://neurips.cc/public/EthicsGuidelines}?
\item[] Answer: \answerYes{}
\item[] Justification: We reviewed the NeurIPS Code of Ethics. The work studies RL algorithms using simulated environments and offline benchmark data, and does not involve human subjects, private data, or deployment of high-risk models.
\item[] Guidelines:
\begin{itemize}
\item The answer \answerNA{} means that the authors have not reviewed the NeurIPS Code of Ethics.
\item If the authors answer \answerNo{}, they should explain the special circumstances that require a deviation from the Code of Ethics.
\item The authors should make sure to preserve anonymity (e.g., if there is a special consideration due to laws or regulations in their jurisdiction).
\end{itemize}

\item {\bf Broader impacts}
\item[] Question: Does the paper discuss both potential positive societal impacts and negative societal impacts of the work performed?
\item[] Answer: \answerNA{}
\item[] Justification: Split-RL is a methodological RL contribution evaluated in simulation and offline benchmark settings. The paper does not deploy trained controllers, release a user-facing model, use human-subject or private data, or make application-specific deployment claims; any societal impact would depend on a separate downstream deployment outside the scope of the paper.
\item[] Guidelines:
\begin{itemize}
\item The answer \answerNA{} means that there is no societal impact of the work performed.
\item If the authors answer \answerNA{} or \answerNo{}, they should explain why their work has no societal impact or why the paper does not address societal impact.
\item Examples of negative societal impacts include potential malicious or unintended uses (e.g., disinformation, generating fake profiles, surveillance), fairness considerations (e.g., deployment of technologies that could make decisions that unfairly impact specific groups), privacy considerations, and security considerations.
\item The conference expects that many papers will be foundational research and not tied to particular applications, let alone deployments. However, if there is a direct path to any negative applications, the authors should point it out. For example, it is legitimate to point out that an improvement in the quality of generative models could be used to generate Deepfakes for disinformation.
\item On the other hand, it is not needed to point out that a generic algorithm for optimizing neural networks could enable people to train models that generate Deepfakes faster.
\item The authors should consider possible harms that could arise when the technology is being used as intended and functioning correctly, harms that could arise when the technology is being used as intended but gives incorrect results, and harms following from (intentional or unintentional) misuse of the technology.
\item If there are negative societal impacts, the authors could also discuss possible mitigation strategies (e.g., gated release of models, providing defenses in addition to attacks, mechanisms for monitoring misuse, mechanisms to monitor how a system learns from feedback over time, improving the efficiency and accessibility of ML).
\end{itemize}

\item {\bf Safeguards}
\item[] Question: Does the paper describe safeguards that have been put in place for responsible release of data or models that have a high risk for misuse (e.g., pre-trained language models, image generators, or scraped datasets)?
\item[] Answer: \answerNA{}
\item[] Justification: The paper does not release high-risk pretrained models, image generators, scraped datasets, or other assets with an obvious high risk for misuse. The evaluated environments and offline benchmarks are standard simulation or control benchmarks.

\item {\bf Licenses for existing assets}
\item[] Question: Are the creators or original owners of assets (e.g., code, data, models), used in the paper, properly credited and are the license and terms of use explicitly mentioned and properly respected?
\item[] Answer: \answerYes{}
\item[] Justification: The existing software, environment, and model assets are credited in the main text, appendix, and references.

\item {\bf New assets}
\item[] Question: Are new assets introduced in the paper well documented and is the documentation provided alongside the assets?
\item[] Answer: \answerNA{}
\item[] Justification: The submission does not introduce a standalone new dataset, pretrained model, or benchmark suite as a primary contribution. The custom scenarios and label/cost rules are experimental implementation components documented in Appendices~\ref{app:minigrid_envs}--\ref{app:citylearn_scenarios} and included with the code release; they are not presented as a separate asset release.

\item {\bf Crowdsourcing and research with human subjects}
\item[] Question: For crowdsourcing experiments and research with human subjects, does the paper include the full text of instructions given to participants and screenshots, if applicable, as well as details about compensation (if any)?
\item[] Answer: \answerNA{}
\item[] Justification: The paper does not involve crowdsourcing, user studies, or research with human subjects.

\item {\bf Institutional review board (IRB) approvals or equivalent for research with human subjects}
\item[] Question: Does the paper describe potential risks incurred by study participants, whether such risks were disclosed to the subjects, and whether Institutional Review Board (IRB) approvals (or an equivalent approval/review based on the requirements of your country or institution) were obtained?
\item[] Answer: \answerNA{}
\item[] Justification: The paper does not involve crowdsourcing or research with human subjects, so IRB or equivalent review is not applicable.

\item {\bf Declaration of LLM usage}
\item[] Question: Does the paper describe the usage of LLMs if it is an important, original, or non-standard component of the core methods in this research? Note that if the LLM is used only for writing, editing, or formatting purposes and does \emph{not} impact the core methodology, scientific rigor, or originality of the research, declaration is not required.
\item[] Answer: \answerNA{}
\item[] Justification: LLMs were not used as part of the core methodology or experiments. They were used to review the paper for accuracy and correctness, including checking mathematical derivations, helping formulate and edit paragraphs, checking spelling and grammar, and assisting with the illustration in Figure 1. All content and suggested changes were reviewed and verified by the authors.
\item[] Guidelines:
\begin{itemize}
\item The answer \answerNA{} means that the core method development in this research does not involve LLMs as any important, original, or non-standard components.
\item Please refer to the LLM policy in the NeurIPS handbook for what should or should not be described.
\end{itemize}

\end{enumerate}